%% file: adaptive_weighting_neurips_CR.tex
\documentclass{article}

\usepackage[final]{neurips_2021}

\usepackage{commands}
\usepackage{natbib}
\usepackage{minitoc}

\usepackage[utf8]{inputenc} 
\usepackage[T1]{fontenc}    
\usepackage{hyperref}       
\usepackage{url}            
\usepackage{booktabs}       
\usepackage{amsfonts}       
\usepackage{nicefrac}       
\usepackage{microtype}      
\usepackage{xcolor}         

\title{Statistical Inference with M-Estimators on Adaptively Collected Data}

\author{%
  Kelly W. Zhang \\
  Department of Computer Science\\
  Harvard University \\
  \texttt{kellywzhang@seas.harvard.edu} \\
   \And
   Lucas Janson \\
   Departments of Statistics \\
   Harvard University \\
   \texttt{ljanson@fas.harvard.edu} \\
   \AND
   Susan A. Murphy \\
   Departments of Statistics and Computer Science \\
   Harvard University \\
   \texttt{samurphy@fas.harvard.edu} \\
}

\begin{document}

\maketitle

\begin{abstract}
Bandit algorithms are increasingly used in real-world sequential decision-making problems. Associated with this is an increased desire to be able to use the resulting datasets to answer scientific questions like: Did one type of ad lead to more purchases? In which contexts is a mobile health intervention effective? However, classical statistical approaches fail to provide valid confidence intervals when used with data collected with bandit algorithms. Alternative methods have recently been developed for simple models (e.g., comparison of means). Yet there is a lack of general methods for  conducting statistical inference using more complex models on data collected with (contextual) bandit algorithms; for example, current methods cannot be used for valid inference on parameters in a logistic regression model for a binary reward. In this  work, we develop theory justifying the use of M-estimators---which  includes estimators based on empirical risk minimization as well as maximum likelihood---on data collected with adaptive algorithms, including (contextual) bandit algorithms. Specifically, we show that M-estimators, modified with particular adaptive weights, can be used  to construct asymptotically valid confidence regions for a variety of inferential targets.
\end{abstract}

\doparttoc 
\faketableofcontents 


\section{Introduction}

Due to the need for interventions that are personalized to users, (contextual) bandit algorithms are increasingly used to address sequential decision making problems in health-care \citep{yom2017encouraging,liao2020personalized}, online education \citep{liu2014trading,shaikh2019balancing}, and public policy \citep{kasy2021adaptive,caria2020adaptive}. 
Contextual bandits personalize, that is, minimize regret, by learning to choose the best intervention in each context, i.e., the action that leads to the greatest expected reward. 
Besides the goal of regret minimization,
another critical goal in these real-world problems is to be able to use the resulting data collected by bandit algorithms to advance scientific knowledge \citep{liu2014trading,erraqabi2017trading}. By scientific knowledge, we mean information gained by using the data 
to conduct a variety of statistical analyses, including confidence interval construction and hypothesis testing. 
\bo{While regret minimization is a \textit{within}-experiment learning objective, gaining scientific knowledge from the resulting adaptively collected data is a \textit{between}-experiment learning objective}, which ultimately helps with regret minimization between deployments of bandit algorithms. 
Note that the data collected by bandit algorithms are \textit{adaptively collected} because previously observed contexts, actions, and rewards 
are used to inform what actions to select in future timesteps. 

There are a variety of between-experiment learning questions encountered in real-life applications of bandit algorithms. For example, in real-life sequential decision-making problems there are often a number of additional scientifically interesting outcomes besides the reward that are collected during the experiment. In the online advertising setting, the reward might be whether an ad is clicked on, but one may be interested in the outcome of amount of money spent or the subsequent time spent on the advertiser's website. If it was found that an ad had high click-through rate, but low amounts of money was spent after clicking on the ad, one may redesign the reward used in the next bandit experiment. 
One type of statistical analysis would be to construct confidence intervals for the relative effect of the actions on multiple outcomes (in addition to the reward) conditional on the context. 
Furthermore, due to engineering and practical limitations, some of the variables that might be useful as context are often not accessible to the bandit algorithm online. 
If after-study analyses find some such contextual variables to have sufficiently strong influence on the relative usefulness of an action, this might lead investigators to ensure these variables are accessible to the bandit algorithm in the next experiment. 

As discussed above, we can gain scientific knowledge from data collected with (contextual) bandit algorithms by constructing confidence intervals and performing hypothesis tests for unknown quantities such as the expected outcome for different actions in various contexts. Unfortunately, standard statistical methods developed for i.i.d. data fail to provide valid inference when applied to data collected with common bandit algorithms. For example, assuming the sample mean of rewards for an arm is approximately normal can lead to unreliable confidence intervals and inflated type-1 error; see Section \ref{sec:intuition} for an illustration.
Recently statistical inference methods have been developed for data collected using bandit algorithms \citep{athey, deshpande, zhang2020inference}; however, these methods are limited to inference for parameters of simple models. There is a lack of general statistical inference methods for data collected with (contextual) bandit algorithms in more complex data-analytic settings, including parameters in non-linear models for outcomes; for example, there are currently no methods for constructing valid 
confidence intervals for the parameters of a logistic regression model for binary outcomes or for constructing confidence intervals based on robust estimators like minimizers of the Huber loss function.

In this work we show that a wide variety of estimators which are frequently used both in science and industry on i.i.d. data, namely, M-estimators \citep{van2000asymptotic}, can be used to conduct valid inference on data collected with (contextual) bandit algorithms when adjusted with particular adaptive weights, i.e., weights that are a function of previously collected data. Different forms of adaptive weights are used by existing methods for simple models \citep{deshpande,athey,zhang2020inference}. 
Our work is a step towards developing a general framework for statistical inference on data collected with adaptive algorithms, including (contextual) bandit algorithms.


\section{Problem Formulation}
\label{sec:problemformulation}

We assume that the data we have after running a contextual bandit algorithm is comprised of contexts $\{ X_t \}_{t=1}^T$, actions $\{ A_t \}_{t=1}^T$, and primary outcomes $\{ Y_t \}_{t=1}^T$. $T$ is deterministic and known. 
We assume that rewards are a deterministic function of the primary outcomes, i.e., $R_t = f(Y_t)$ for some known function $f$. 
We are interested in constructing confidence regions for the parameters of the conditional distribution of $Y_t$ given $(X_t, A_t)$. Below we consider $T \to \infty$ in order to derive the asymptotic distributions of estimators and construct asymptotically valid confidence intervals.
We allow the action space $\MC{A}$ to be finite or infinite. We use potential outcome notation \citep{imbens2015causal} and let $\{ Y_t(a) : a \in \MC{A} \}$ denote the potential outcomes of the primary outcome and let $Y_t := Y_t(A_t)$ be the observed outcome. We assume a stochastic contextual bandit environment in which $\left\{  X_t, Y_t(a) : a \in \MC{A} \right\} \iidsim \MC{P} \in \boldP$ for $t \in [1 \colon T]$; 
the contextual bandit environment distribution $\MC{P}$ is in a space of possible environment distributions $\boldP$.
We define the history $\HH_t := \{ X_{t'}, A_{t'}, Y_{t'} \}_{t'=1}^t$ for $t \geq 1$ and $\HH_0 := \emptyset$. Actions $A_t \in \MC{A}$ are selected according to policies $\pi := \{ \pi_t \}_{t \geq 1}$, which define action selection probabilities $\pi_t(A_t, X_t, \HH_{t-1}) := \PP \left( A_t | \HH_{t-1}, X_t \right)$. Even though the potential outcomes are i.i.d., the \textit{observed} data $\{ X_t, A_t, Y_t \}_{t=1}^T$ are \textit{not} because the actions are selected using policies $\pi_t$ which are a function of past data, $\HH_{t-1}$.
Non-independence of observations is a key property of adaptively collected data.


We are interested in constructing confidence regions for some unknown $\theta^*(\MC{P}) \in \Theta \subset \real^d$, which is a parameter of the conditional distribution of $Y_t$ given $(X_t, A_t)$. This work focuses on the setting in which we have a well-specified model for $Y_t$. Specifically, we assume that $\theta^*(\MC{P})$ is a conditionally maximizing value of criterion $m_{\theta}$, i.e., for all $\MC{P} \in \boldP$,
\begin{equation}
	\label{eqn:thetastar}
	\theta^*(\MC{P}) \in \argmax_{\theta \in \Theta} \E_{\MC{P}} \left[ m_{\theta} ( Y_t, X_t, A_t ) | X_t, A_t \right] ~ \TN{ w.p. } 1.
\end{equation}
Note that $\theta^*(\MC{P})$ does not depend on $(X_t, A_t)$ and it is an implicit modelling assumption that such a $\theta^*(\MC{P})$ exists for a given $m_\theta$. Note that this formulation includes semi-parametric models, e.g., the model could constrain the conditional mean of $Y_t$ to be linear in some function of the actions and context, but allow the residuals to follow any mean-zero distribution, including ones that depend on the actions and/or contexts.

To estimate $\theta^*(\MC{P})$, we build on M-estimation \citep{huber1992robust}, which classically selects the estimator $\hat{\theta}$ to be the $\theta \in \Theta$ that maximizes the empirical analogue of Equation~\eqref{eqn:thetastar}:
\begin{equation}
	\label{eqn:unweightedMestimator}
	\hat{\theta}_T := \argmax_{\theta \in \Theta} \frac{1}{T} \sum_{t=1}^T m_{\theta} ( Y_t, X_t, A_t ).
\end{equation}
For example, in a classical linear regression setting with $|\MC{A}| < \infty$ actions, a natural choice for $m_{\theta}$ is the negative of the squared loss function, 
$m_{\theta} ( Y_t, X_t, A_t ) = - ( Y_t - X_t^\top \theta_{A_t} )^2$. 
When $Y_t$ is binary, a natural choice is instead the negative log-likelihood function for a logistic regression model, i.e., $m_{\theta} ( Y_t, X_t, A_t ) = - [ Y_t X_t^\top \theta_{A_t} - \log ( 1 + \exp ( X_t^\top \theta_{A_t} ) ) ]$. 
More generally, $m_{\theta}$ is commonly chosen to be a log-likelihood function or the negative of a robust loss function such as the Huber loss. 
If the data, $\{X_t, A_t, Y_t\}_{t=1}^T$, were independent across time, classical approaches could be used to prove the consistency and asymptotic normality of M-estimators \citep{van2000asymptotic}. However, on data collected with bandit algorithms, standard M-estimators like the ordinary least-squares estimator fail to provide valid confidence intervals \citep{athey,deshpande,zhang2020inference}. In this work, we show that M-estimators can still be used to provide valid statistical inference on adaptively collected data when adjusted with well-chosen adaptive weights.

\section{Adaptively Weighted M-Estimators}
\label{sec:awMestimator}


We consider a weighted M-estimating criteria with adaptive weights $W_t \in \sigma (\HH_{t-1}, X_t, A_t)$ given by $W_t = \sqrt{ \frac{ \pi_t^{\TN{sta}} (A_t, X_t) }{ \pi_t (A_t, X_t, \HH_{t-1}) } }$. Here $\{ \pi_t^{\TN{sta}} \}_{t \geq 1}$ are pre-specified \textit{stabilizing policies} that do not depend on data $\{ Y_t, X_t, A_t \}_{t \geq 1}$.
A default choice for the stabilizing policy when the action space is of size $|\MC{A}| < \infty$ is just $\pi_t^{\TN{sta}}(a, x) = 1/|\MC{A}|$ for all $x$, $a$, and $t$; we discuss considerations for the choice of $\{ \pi_t^{\TN{sta}} \}_{t=1}^T$ in Section \ref{sec:stabilizingPolicy}.
We call these weights \textit{square-root importance weights} because they are the square-root of the standard importance weights \citep{hammersley2013monte,wang2017optimal}. 
Our proposed estimator for $\theta^*(\MC{P})$, $\thetahat_T$, is the maximizer of a \emph{weighted} version of the M-estimation criterion of Equation~\eqref{eqn:unweightedMestimator}: 
\begin{equation*}
	\thetahat_T := \argmax_{\theta \in \Theta} \frac{1}{T} \sum_{t=1}^T W_t m_\theta (Y_t, X_t, A_t)
	=: \argmax_{\theta \in \Theta} M_T( \theta ).
\end{equation*}
Note that $M_T(\theta)$ defined above depends on both the data $\{ X_t, A_t, Y_t \}_{t=1}^T$ and weights $\{ W_t \}_{t=1}^T$. 
We provide asymptotically valid confidence regions for $\theta^*(\MC{P})$ by deriving the asymptotic distribution of $\hat\theta_T$ as $T \to \infty$ and by proving that the convergence in distribution is \textit{uniform} over $\MC{P} \in \boldP$. 
Such convergence allows us to construct a uniformly asymptotically valid $1-\alpha$ level confidence region, $C_T(\alpha)$, for $\theta^*(\MC{P})$, which is a confidence region that satisfies
\begin{equation}
	\label{eqn:uniformCI}
	\liminf_{T \to \infty} \inf_{\MC{P} \in \boldP} \PP_{\MC{P}, \pi} \left( \theta^*(\MC{P}) \in C_T(\alpha) \right) \geq 1-\alpha. 
\end{equation}
If $C_T(\alpha)$ were \emph{not} uniformly valid, then there would exist an $\epsilon > 0$ such that for \emph{every} sample size $T$, $C_T(\alpha)$'s coverage would be below $1-\alpha-\epsilon$ for some worst-case $P_T \in \boldP$. 
Confidence regions which are asymptotically valid, but not \emph{uniformly} asymptotically valid, fail to be reliable in practice
\citep{leeb2005model,romano2012uniform}. 
Note that on i.i.d. data it is generally straightforward to show that estimators that converge in distribution do so uniformly; however, as discussed in \citet{zhang2020inference} and Appendix \ref{app:uniformity}, this is not the case on data collected with bandit algorithms. 

To construct uniformly valid confidence regions for $\theta^*(\MC{P})$ we prove that $\thetahat_T$ is uniformly asymptotically normal in the following sense:
\begin{equation}
	\label{eqn:Mestgoal}
	\Sigma_T(\MC{P})^{-1/2} \ddot{M}_T( \thetahat_T ) \sqrt{T} ( \thetahat_T - \theta^*(\MC{P}) )
	\Dto \N \left( 0, I_d \right) \TN{ uniformly over } \MC{P} \in \boldP,
\end{equation}
where $\ddot{M}_T( \theta ) := \frac{\partial^2}{\partial^2 \theta} M_T( \theta )$ and $\Sigma_T(\MC{P}) := \frac{1}{T} \sum_{t=1}^T \E_{\MC{P}, \pi_t^{\TN{sta}}} \left[ \dot{m}_{\theta^*(\MC{P})}(Y_t, X_t, A_t)^{\otimes 2} \right]$. We define $\dot{m}_{\theta} := \frac{\partial}{\partial \theta} m_{\theta}$. Similarly we define respectively $\ddot{m}_\theta$ and $\dddot{m}_\theta$ as the second and third partial derivatives of $m_\theta$ with respect to $\theta$. For any vector $z$ we define $z^{\otimes2} := z z^\top$. 

\subsection{Intuition for Square-Root Importance Weights}
\label{sec:intuition}


The critical role of the square-root importance weights $W_t = \sqrt{ \frac{\pi_t^{\TN{sta}}(A_t, X_t) }{\pi_t(A_t, X_t, \HH_{t-1}) } }$ is to adjust for instability in the \textit{variance} of M-estimators due to the bandit algorithm. These weights act akin to standard importance weights when squared and adjust a key term in the variance of M-estimators from depending on adaptive policies $\{ \pi_t \}_{t=1}^T$, which can be ill-behaved, to depending on the pre-specified stabilizing policies $\{ \pi_t^{\TN{sta}} \}_{t=1}^T$. See \citet{zhang2020inference} and \citet{deshpande} for more discussion of the ill-behavior of the action selection probabilities for common bandit algorithms, which occurs particularly when there is no unique optimal policy.

As an illustrative example, consider the least-squares estimators in a finite-arm linear contextual bandit setting. Assume that $\E_{\MC{P}} [ Y_t | X_t, A_t=a ] = X_t^\top \theta_a^*(\MC{P}) \TN{ w.p. } 1$. We focus on estimating $\theta_a^*(\MC{P})$ for some $a \in \MC{A}$.
The least-squares estimator corresponds to an M-estimator with $m_{\theta_a}(Y_t, X_t, A_t) = - \1_{A_t = a} ( Y_t - X_t^\top \theta_a )^2$. 
The adaptively weighted least-squares (AW-LS) estimator is $\hat{\theta}_{T,a}^{\TN{AW-LS}} := \argmax_{\theta_a} \{- \sum_{t=1}^T W_t \1_{A_t = a} ( Y_t - X_t^\top \theta_a )^2 \}$. For simplicity, suppose that the stabilizing policy does not change with $t$ and drop the index $t$ to get $\pi^{\TN{sta}}$.
Taking the derivative of this criterion, we get $0 = \sum_{t=1}^T W_t \1_{A_t = a} X_t \big( Y_t - X_t^\top \hat{\theta}_{T,a}^{\TN{AW-LS}} \big)$, and rearranging terms gives
\begin{equation}
	\label{eqn:awlsEquation}
	\frac{1}{ \sqrt{T} } \sum_{t=1}^T W_t \1_{A_t = a} X_t X_t^\top \left( \hat{\theta}_{T,a}^{\TN{AW-LS}} - \theta_a^*(\MC{P}) \right)	 
	= \frac{1}{ \sqrt{T} } \sum_{t=1}^T W_t \1_{A_t = a} X_t \left( Y_t - X_t^\top \theta_a^*(\MC{P}) \right).
\end{equation}
Note that the right hand side of Equation~\eqref{eqn:awlsEquation} is a martingale difference sequence with respect to history $\{ \HH_t \}_{t=0}^T$ because $\E_{\MC{P}, \pi} [ W_t \1_{A_t = a} ( Y_t - X_t^\top \theta_a^*(\MC{P}) ) | \HH_{t-1} ] = 0$ for all $t$; by law of iterated expectations and since $W_t \in \sigma(\HH_{t-1}, X_t, A_t)$, $\E_{\MC{P}, \pi} [ W_t \1_{A_t = a} ( Y_t - X_t^\top \theta_a^*(\MC{P}) ) | \HH_{t-1} ]$ equals
\begin{equation*}
    \E_{\MC{P}} \left[ W_t \pi_t(a, X_t, \HH_{t-1} ) \E_{\MC{P}} \left[ Y_t - X_t^\top \theta_a^*(\MC{P}) | \HH_{t-1}, X_t, A_t = a \right] \big| \HH_{t-1} \right] 
\end{equation*}
\begin{equation*}
    \underset{(i)}{=} \E_{\MC{P}} \left[ W_t \pi_t(a, X_t, \HH_{t-1} ) \E_{\MC{P}} \left[ Y_t - X_t^\top \theta_a^*(\MC{P}) | X_t, A_t = a \right] \big| \HH_{t-1} \right] \underset{(ii)}{=} 0.
\end{equation*}
(i) holds by our i.i.d. potential outcomes assumption. (ii) holds since $\E_{\MC{P}} [ Y_t | X_t, A_t=a ] = X_t^\top \theta_a^*(\MC{P})$. 
We prove that \eqref{eqn:awlsEquation} is uniformly asymptotically normal by applying a martingale central limit theorem (Appendix \ref{app:normalityLemmas}). The key condition in this theorem is that the conditional variance converges uniformly, for which
it is sufficient to show that the conditional covariance of $W_t \1_{A_t = a} \left( Y_t - X_t^\top \theta_a^*(\MC{P}) \right)$ given $\HH_{t-1}$ equals some positive-definite matrix $\Sigma(\MC{P})$ for every $t$, i.e.,
\begin{equation}
	\label{eqn:condvar}
	 \E_{\MC{P}, \pi} \left[ W_t^2 \1_{A_t = a} X_t X_t^\top \left( Y_t - X_t^\top \theta_a^*(\MC{P}) \right)^2 \big| \HH_{t-1} \right] = \Sigma(\MC{P}).
\end{equation}

By law of iterated expectations, $\E_{\MC{P}, \pi} [ W_t^2 \1_{A_t = a} X_t X_t^\top ( Y_t - X_t^\top \theta_a^*(\MC{P}) )^2 \big| \HH_{t-1} ]$ equals
\begin{equation}
    \label{eqn:variancestbilizingintuition}
	 \E_{\MC{P}} \left[ \E_{\MC{P}, \pi} \left[ \frac{\pi^{\TN{sta}}(A_t, X_t) }{\pi_t(A_t, X_t, \HH_{t-1}) } \1_{A_t = a} X_t X_t^\top \left( Y_t - X_t^\top \theta_a^*(\MC{P}) \right)^2 \bigg| \HH_{t-1}, X_t \right] \bigg| \HH_{t-1} \right] \\
\end{equation}
\begin{equation*}
	 \underset{(a)}{=} \E_{\MC{P}} \left[ \E_{\MC{P}, \pi^{\TN{sta}}} \left[ \1_{A_t = a} X_t X_t^\top \left( Y_t - X_t^\top \theta_a^*(\MC{P}) \right)^2 \bigg| \HH_{t-1}, X_t \right] \bigg| \HH_{t-1} \right]
\end{equation*}
\begin{equation*}
	 \underset{(b)}{=} \E_{\MC{P}} \left[ \E_{\MC{P}, \pi^{\TN{sta}}} \left[ \1_{A_t = a} X_t X_t^\top \left( Y_t - X_t^\top \theta_a^*(\MC{P}) \right)^2 \bigg| X_t \right] \bigg| \HH_{t-1} \right]
\end{equation*}
\begin{equation*}
	 \underset{(c)}{=} \E_{\MC{P}} \left[ \E_{\MC{P}, \pi^{\TN{sta}}} \left[ \1_{A_t = a} X_t X_t^\top \left( Y_t - X_t^\top \theta_a^*(\MC{P}) \right)^2 \bigg| X_t \right]  \right]
\end{equation*}
\begin{equation*}
	 \underset{(d)}{=} \E_{\MC{P}, \pi^{\TN{sta}}} [ \1_{A_t = a} X_t X_t^\top ( Y_t - X_t^\top \theta_a^*(\MC{P}) )^2 ] =: \Sigma(\MC{P}).
\end{equation*}
Above, (a) holds because the importance weights change the sampling measure from the adaptive policy $\pi_t$ to the pre-specified stabilizing policy $\pi^{\TN{sta}}$. 
(b) holds by our i.i.d. potential outcomes assumption and because $\pi^{\TN{sta}}$ is a pre-specified policy.
(c) holds because $X_t$ does not depend on $\HH_{t-1}$ by our i.i.d. potential outcomes assumption.
(d) holds by the law of iterated expectations.
Note that $\Sigma(\MC{P})$ does not depend on $t$ because $\pi^{\text{sta}}$ is not time-varying. 
In contrast, without the adaptive weighting, i.e., when $W_t = 1$, the conditional covariance of $\1_{A_t = a} \left( Y_t - X_t^\top \theta_a^*(\MC{P}) \right)$ on $\HH_{t-1}$ is a random variable, due to the adaptive policy $\pi_t$.

\begin{figure}[H] 
	\centerline{
	\includegraphics[width=0.42\linewidth]{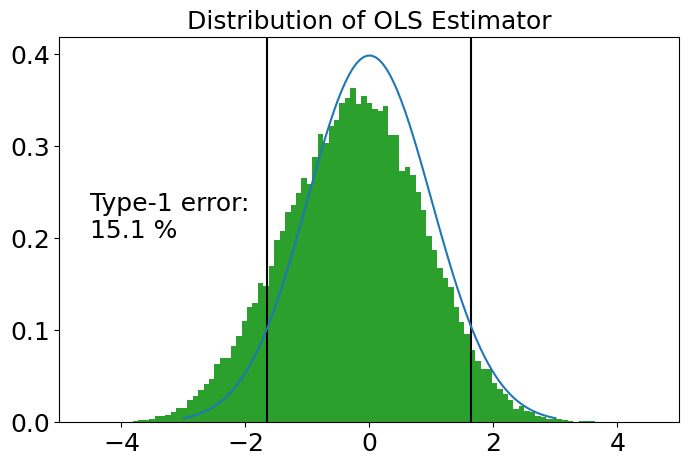}
	~~~~
	\includegraphics[width=0.5\linewidth]{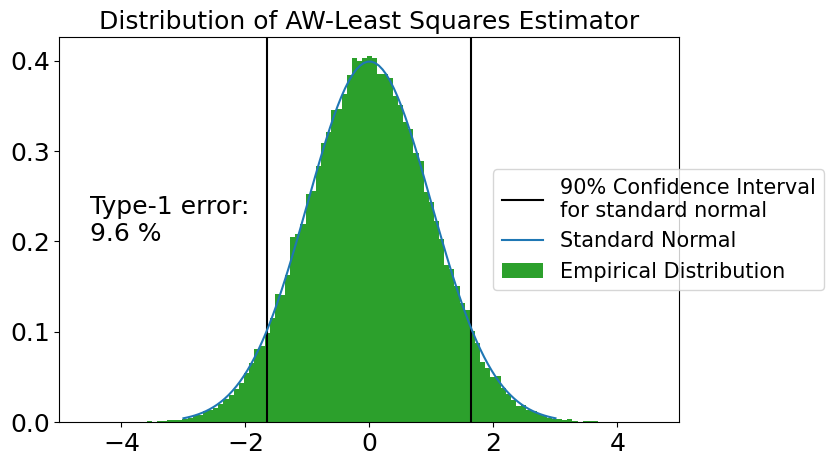}
	}
  \caption{The empirical distributions of the weighted and unweighted least-squares estimators for $\theta_1^*(\MC{P}) := \E_{\MC{P}} [ Y_t(1) ]$ in a two arm bandit setting where $\E_{\MC{P}} [ Y_t(1) ] = \E_{\MC{P}} [ Y_t(0) ] = 0$. We perform Thompson Sampling with $\N(0,1)$ priors, $\N(0,1)$ errors, and $T=1000$. Specifically, we plot $\sqrt{\sum_{t=1}^T A_t } ( \hat{\theta}_{T,1}^{\OLS} - \theta_1^*(\MC{P}) )$ on the left and $\left( \frac{1}{\sqrt{T}} \sum_{t=1}^T \sqrt{ \frac{ 0.5 }{\pi_t(1)}} A_t \right) ( \hat{\theta}_{T,1}^{\TN{AW-LS}} - \theta_1^*(\MC{P}) )$ on the right.}
  \label{fig:illustrating_weighting}
\end{figure}

In Figure~\ref{fig:illustrating_weighting} we plot the empirical distributions of the z-statistic for the least-squares estimator both with and without adaptive weighting. We consider a two-armed bandit with $A_t \in \{0,1\}$. Let $\theta_1^*(\MC{P}) := \E_{\MC{P}}[ Y_t(1) ]$ and $m_{\theta_1}(Y_t, A_t) := - A_t ( Y_t - \theta_1 )^2$. The unweighted version, i.e., the ordinary least-squares (OLS) estimator, is $\thetahat_{T,1}^{\OLS} := \argmax_{\theta_1} \frac{1}{T} \sum_{t=1}^T m_{\theta_1}(Y_t, A_t)$. The adaptively weighted version is $\thetahat_{T,1}^{\TN{AW-LS}} := \argmax_{\theta_1} \frac{1}{T} \sum_{t=1}^T W_t m_{\theta_1}(Y_t, A_t)$. 
We collect data using Thompson Sampling and use a uniform stabilizing policy where $\pi^{\TN{sta}}(1) = \pi^{\TN{sta}}(0) = 0.5$. It is clear that the least-squares estimator with adaptive weighting has a z-statistic that is much closer to a normal distribution. 

The square-root importance weights are a form of variance stabilizing weights, akin to those introduced in \citet{athey} for estimating means and differences in means on data collected with multi-armed bandits. In fact, in the special case that $|\MC{A}| < \infty$ and $\phi(X_t, A_t) = [\1_{A_t=1}, \1_{A_t=2}, ..., \1_{A_t=|\MC{A}|}]^\top$, the adaptively weighted least-squares estimator is equivalent to the weighted average estimator of \citet{athey}. See Section \ref{sec:relatedwork} for more on \citet{athey}.
 


\subsection{Asymptotic Normality and Confidence Regions}
\label{sec:normality}

We now discuss conditions under which 
the adaptively weighted M-estimators are asymptotically normal in the sense of Equation~\eqref{eqn:Mestgoal}. In general, our conditions differ from those made for standard M-estimators on i.i.d. data because (i) the data is adaptively collected, i.e., $\pi_t$ can depend on $\HH_{t-1}$ and (ii) we ensure uniform convergence over $\MC{P} \in \boldP$, which is stronger than guaranteeing convergence pointwise for each $\MC{P} \in \boldP$.

\begin{condition}[Stochastic Bandit Environment]
	\label{cond:iidpo}
	Potential outcomes $\left\{ X_t, Y_t(a) : a \in \MC{A} \right\} \iidsim \MC{P} \in \boldP$ over $t \in [1 \colon T]$.
\end{condition}
Condition \ref{cond:iidpo} implies that $Y_t$ is independent of $\mathcal{H}_{t-1}$ given $X_t$ and $A_t$, and the conditional distribution $Y_t \mid X_t, A_t$ is invariant over time. Also note that action space $\MC{A}$ can be finite or infinite.

\begin{condition}[Differentiable]
	\label{cond:differentiable}
	The first three derivatives of $m_\theta(y, x, a)$ with respect to $\theta$ exist for every $\theta \in \Theta$, every $a \in \mathcal{A}$, and every $(x, y)$ in the joint support of $\{ \MC{P} : \MC{P} \in \boldP \}$.
\end{condition}

\begin{condition}[Bounded Parameter Space]
	\label{cond:compact}
	For all $\MC{P} \in \boldP$, $\theta^*(\MC{P}) \in \Theta$, a bounded open subset of $\real^d$.
\end{condition}

\begin{condition}[Lipschitz]
	\label{cond:lipschitz}
	There exists some real-valued function $g$ such that (i) $\sup_{\MC{P} \in \boldP, t \geq 1} \E_{\MC{P}, \pi_t^{\TN{sta}}}[ g (Y_t, X_t, A_t)^2 ]$ is bounded and (ii) for all $\theta, \theta' \in \Theta$,
	\begin{equation*}
		\left| m_{\theta}  (Y_t, X_t, A_t)  - m_{\theta'}  (Y_t, X_t, A_t) \right| \leq g  (Y_t, X_t, A_t) \| \theta - \theta' \|_2.
	\end{equation*}
\end{condition}
Conditions \ref{cond:compact} and \ref{cond:lipschitz} together restrict the complexity of the function $m$ in order to ensure a martingale law of large numbers result holds uniformly over functions $\{ m_\theta : \theta \in \Theta \}$; this is used to prove the consistency of $\thetahat_T$. Similar conditions are commonly used to prove consistency of M-estimators based on i.i.d. data, although the boundedness of the parameter space can be dropped when $m_\theta$ is a concave function of $\theta$ for all $Y_t, A_t, X_t$ (as it is in many canonical examples such as least squares) \citep{van2000asymptotic,engle1994handbook,bura2018asymptotic}; we expect that a similar result would hold for adaptively weighted M-estimators.


\begin{condition}[Moments]
	\label{cond:moments}
	The fourth moments of $m_{\theta^*(\MC{P})} (Y_t, X_t, A_t)$, $\dot{m}_{\theta^*(\MC{P})} (Y_t, X_t, A_t)$, and $\ddot{m}_{\theta^*(\MC{P})} (Y_t, X_t, A_t)$ with respect to $\MC{P}$ and policy $\pi_t^{\TN{sta}}$ are bounded uniformly over $\MC{P} \in \boldP$ and $t \geq 1$. 
	For all sufficiently large $T$, the minimum eigenvalue of $\Sigma_{T, P} := \frac{1}{T} \sum_{t=1}^T \E_{\MC{P}, \pi_t^{\TN{sta}} } \left[ \dot{m}_{\theta^*(\MC{P})} (Y_t, X_t, A_t)^{\otimes 2} \right]$ is bounded above $\delta_{\dot{m}^2} > 0$ for all $\MC{P} \in \boldP$.
\end{condition}
Condition \ref{cond:moments} is similar to those of \citet[Theorem 5.41]{van2000asymptotic}. However, to guarantee uniform convergence we assume that moment bounds hold uniformly over $\MC{P} \in \boldP$ and $t \geq 1$.

\begin{condition}[Third Derivative Domination]
	\label{cond:thirdderivdomination}
	For $B \in \real^{d \by d \by d}$, we define $\| B \|_1 := \sum_{i=1}^d \sum_{j=1}^d \sum_{k=1}^d |B_{i,j,k}|$. There exists a function $\dddot{m} (Y_t, X_t, A_t) \in \real^{d \by d \by d}$ such that (i) $\sup_{\MC{P} \in \boldP, t \geq 1} \E_{\MC{P}, \pi_t^{\TN{sta}}} \left[ \| \dddot{m} (Y_t, X_t, A_t) \|_1^2 \right]$ is bounded and (ii) for all $\MC{P} \in \boldP$ there exists some $\epsilon_{ \dddot{m} } > 0$ such that the following holds with probability $1$,
	\begin{equation*}
		\label{eqn:thirdderiv}
		\sup_{\theta \in \Theta : \| \theta - \theta^*(\MC{P}) \| \leq \epsilon_{\dddot{m}} } \| \dddot{m}_{\theta} (Y_t, X_t, A_t) \|_1
		\leq \| \dddot{m} (Y_t, X_t, A_t) \|_1.
	\end{equation*}
\end{condition}
Condition \ref{cond:thirdderivdomination} is again similar to those in classical M-estimator asymptotic normality proofs \citep[Theorem 5.41]{van2000asymptotic}.



\begin{condition}[Maximizing Solution] 
	\label{cond:optimalsolution}
	~ \\
	(i) For all $\MC{P} \in \boldP$, there exists a $\theta^*(\MC{P}) \in \Theta$ such that (a) $\theta^*(\MC{P}) \in \argmax_{\theta \in \Theta} \E_{\MC{P}} \left[ m_{\theta}(Y_t, X_t, A_t) \big| X_t, A_t \right]$ w.p. $1$, 
	(b) $\E_{\MC{P}} \left[ \dot{m}_{\theta^*(\MC{P})}(Y_t, X_t, A_t) \big| X_t, A_t \right] = 0$  w.p. $1$, and (c) $\E_{\MC{P}} \left[ \ddot{m}_{\theta^*(\MC{P})}(Y_t, X_t, A_t) \big| X_t, A_t \right] \preceq 0$ w.p. $1$. \\
	(ii) There exists some positive definite matrix $H$ such that $-\frac{1}{T} \sum_{t=1}^T \E_{\MC{P}, \pi_t^{\TN{sta}} } \left[ \ddot{m}_{\theta^*(\MC{P})} (Y_t, X_t, A_t) \right] \succeq H$ for all $\MC{P} \in \boldP$ and all sufficiently large $T$.
\end{condition}
For matrices $A, B$, we define $A \succeq B$ to mean that $A - B$ is positive semi-definite, as used above.
Condition \ref{cond:optimalsolution} (i) ensures that $\theta^*(\MC{P})$ is a conditionally maximizing solution for all contexts $X_t$ and actions $A_t$; this ensures that $\{ \dot{m}_{\theta^*(\MC{P})} (Y_t, X_t, A_t) \}_{t=1}^T$ is a martingale difference sequence with respect to $\{ \HH_t \}_{t=1}^T$. Note it does not require $\theta^*(\MC{P})$ to always be a conditionally \emph{unique} optimal solution. 
Condition \ref{cond:optimalsolution} (ii) is related to the local curvature at the maximizing solution and the analogous condition in the i.i.d. setting is trivially satisfied; we specifically use this condition to ensure we can replace $\ddot{M}( \theta^*(\MC{P}) )$ with $\ddot{M}( \thetahat_T )$ in our asymptotic normality result, i.e., that $\ddot{M}( \theta^*(\MC{P}) )^{-1} \ddot{M}( \thetahat_T ) \Pto I_d$ uniformly over $\MC{P} \in \boldP$.

\begin{condition}[Well-Separated Solution] 
	\label{cond:wellseparated}
	For all sufficiently large $T$, for any $\epsilon > 0$, there exists some $\delta > 0$ such that for all $\MC{P} \in \boldP$,
	\begin{equation*}
		\inf_{\theta \in \Theta:\| \theta-\theta^*(\MC{P}) \|_2 > \epsilon} \bigg\{ \frac{1}{T} \sum_{t=1}^T \E_{\MC{P}, \pi_t^{\TN{sta}}} \left[ m_{\theta^*(\MC{P})}(Y_t, X_t, A_t) - m_{\theta}(Y_t, X_t, A_t) \right] \bigg\}
	\geq \delta.
	\end{equation*}
\end{condition}
A well-separated solution condition akin to Condition \ref{cond:wellseparated} is commonly assumed in order to prove consistency of M-estimators, e.g., see \citet[Theorem 5.7]{van2000asymptotic}. 
Note that the difference between Condition \ref{cond:optimalsolution} (i) and Condition \ref{cond:wellseparated} is that the former is a conditional statement (conditional on $X_t, A_t$) and the latter is a marginal statement (marginal over $X_t, A_t$, where $A_t$ is chosen according to stabilizing policies $\pi_t^{\TN{sta}}$). Condition \ref{cond:optimalsolution} (i) means there is a $\theta^*(\MC{P})$ solution for all contexts $X_t$ and actions $A_t$ that does not need to be unique, however Condition \ref{cond:wellseparated} assumes that marginally over $X_t, A_t$ there is a well-separated solution.

\begin{condition}[Bounded Importance Ratios]
	\label{cond:sqrtweights}
	$\{ \pi_t^{\TN{sta}} \}_{t=1}^T$ do not depend on data $\{ Y_t, X_t, A_t \}_{t=1}^T$.
	For all $t \geq 1$, $\rho_{\min} \leq \frac{ \pi_t^{\TN{sta}} (A_t, X_t) }{ \pi_t (A_t, X_t, \HH_{t-1}) } \leq \rho_{\max}$ w.p. $1$ for some constants $0 < \rho_{\min} \leq \rho_{\max} < \infty$.
\end{condition}
Note that Condition \ref{cond:sqrtweights} implies that for a stabilizing policy that is not time-varying, the action selection probabilities of the bandit algorithm $\pi_t (A_t, X_t, \HH_{t-1})$ must be bounded away from zero w.p. $1$. Similar boundedness assumptions are also made in the off-policy evaluation literature \citep{thomas2016data,kallus2020double}. We discuss this condition further in Sections \ref{sec:stabilizingPolicy} and \ref{sec:discussion}. 


\begin{theorem}[Uniform Asymptotic Normality of Adaptively Weighted M-Estimators]
	\label{thm:normality}
	Under Conditions \ref{cond:iidpo}-\ref{cond:sqrtweights} we have that $\hat{\theta}_T \Pto \theta^*(\MC{P})$ uniformly over $\MC{P} \in \boldP$. Additionally,
	\begin{equation}
    		\label{eqn:Mestgoal2}
		\Sigma_T(\MC{P})^{-1/2} \ddot{M}_T( \thetahat_T ) \sqrt{T} ( \thetahat_T - \theta^*(\MC{P}) )
		\Dto \N \left( 0, I_d \right) \TN{ uniformly over } \MC{P} \in \boldP.
	\end{equation}
\end{theorem}
The asymptotic normality result of equation \eqref{eqn:Mestgoal2} guarantees that for $d$-dimensional $\theta^*(\MC{P})$,
\begin{equation*}
	\liminf_{T \to \infty} \inf_{\MC{P} \in \boldP} \PP_{\MC{P}, \pi} \left( \left[ \Sigma_T(\MC{P})^{-1/2} \ddot{M}_T( \thetahat_T ) \sqrt{T} ( \thetahat_T - \theta^*(\MC{P}) ) \right]^{\otimes 2}
	\leq \chi^2_{d,(1-\alpha)} \right) = 1-\alpha.
\end{equation*}
Above $\chi^2_{d,(1-\alpha)}$ is the $1-\alpha$ quantile of the $\chi^2$ distribution with $d$ degrees of freedom.
Note that the region $C_T(\alpha) := \big\{ \theta \in \Theta : [ \Sigma_T(\MC{P})^{-1/2} \ddot{M}_T( \thetahat_T ) \sqrt{T} ( \thetahat_T - \theta^*(\MC{P}) ) ]^{\otimes 2} \leq \chi^2_{d,(1-\alpha)} \big\}$ defines a $d$-dimensional hyper-ellipsoid confidence region for $\theta^*(\MC{P})$. Also note that since $\ddot{M}_T(\thetahat_T)$ does not concentrate under standard bandit algorithms, we cannot use standard arguments to justify treating $\thetahat_T$ as multivariate normal with covariance $\ddot{M}_T(\thetahat_T)^{-1} \Sigma_T(\MC{P}) \ddot{M}_T(\thetahat_T)^{-1}$. Nevertheless, Theorem \ref{thm:normality} can be used to guarantee valid confidence regions for subset of entries in $\theta^*(\MC{P})$ by using projected confidence regions \citep{nickerson1994construction}. Projected confidence regions take a confidence region for all parameters $\theta^*(\MC{P})$ and project it onto the lower dimensional space on which the subset of target parameters lie (Appendix \ref{app:simCR}). 
\subsection{Choice of Stabilizing Policy}
\label{sec:stabilizingPolicy}


When the action space is bounded, using weights $W_t = 1/ \sqrt{\pi_t(A_t, X_t, \HH_{t-1}) }$ is equivalent to using square-root importance weights with a stabilizing policy that selects actions uniformly over $\MC{A}$; this is because weighted M-estimators are invariant to all weights being scaled by the same constant. 
It can make sense to choose a non-uniform stabilizing policy in order to prevent the square-root importance weights from growing too large and to ensure Condition \ref{cond:sqrtweights} holds; disproportionately up-weighting a few observations can lead to unstable estimators. 
Note that an analogue of our stabilizing policy exists in the causal inference literature, 
namely, ``stabilized weights" use a probability density in the numerator of the weights to prevent them from becoming too large 
\citep{robins2000marginal}. 


We now discuss how to choose stabilizing policies $\{ \pi_t^{\TN{sta}} \}_{t \geq 1}$ in order to minimize the asymptotic variance of adaptively weighted M-estimators. We focus on the adaptively weighted least-squares estimator when we have a linear outcome model $\E_{\MC{P}}[ Y_t | X_t, A_t ] = X_t^\top \theta_{A_t}$:
\begin{equation} 
    \label{eqn:awlscriterionPieval}
	\hat{\theta}^{\TN{AW-LS}} := \argmax_{\theta \in \Theta} \bigg\{ \frac{1}{T} \sum_{t=1}^T W_t \left( Y_t - X_t^\top \theta_{A_t} \right)^2 \bigg\}.
\end{equation}
Recall that our use of adaptive weights is to adjust for instability in the variance of M-estimators induced by the bandit algorithm in order to construct valid confidence regions; note that weighted estimators are not typically used for this reason. On i.i.d. data, the least-squares criterion is weighted like in Equation~\eqref{eqn:awlscriterionPieval} in order to minimize the variance of estimators under noise heteroskedasticity; in this setting, the best linear unbiased estimator has weights $W_t = 1/\sigma^2 (A_t, X_t)$ where $\sigma^2(A_t, X_t) := \E_{\MC{P}} [ ( Y_t - X_t^\top \theta_{A_t}^*(\MC{P}) )^2 | X_t, A_t ]$;
this up-weights the importance of observations with low noise variance. Intuitively, if we do not need to variance stabilize, $\{ W_t \}_{t\geq 1}$ should be determined by the relative importance of minimizing the errors for different observations, i.e., their noise variance.

In light of this observation, we expect that under homoskedastic noise there is no reason to up-weight some observations over others. This would recommend choosing the stabilizing policy to make $W_t = \sqrt{  \pi^{\TN{sta}}_t(A_t, X_t) / \pi_t(A_t, X_t, \HH_{t-1}) }$ as close to $1$ as possible, subject to the constraint that the stabilizing policies are pre-specified, i.e., $\{ \pi^{\TN{sta}}_t \}_{t \geq 1}$ do not depend on data $\{ Y_t, X_t, A_t \}_{t \geq 1}$ (see Appendix \ref{app:stabilizingpolicy} for details). 
Since adjusting for heteroskedasticity and variance stabilization are distinct uses of weights, under heteroskedasticity, we recommend that the weights are combined in the following sense: $W_t =\left( 1/\sigma^2 (A_t, X_t)\right) \sqrt{  \pi^{\TN{sta}}_t(A_t, X_t) / \pi_t(A_t, X_t, \HH_{t-1}) }$. 
This would mean that to minimize variance, we still want to choose the stabilizing policies to make $\pi^{\TN{sta}}_t(A_t, X_t) / \pi_t(A_t, X_t, \HH_{t-1})$ as close to $1$ possible, subject to the pre-specified constraint.

\section{Related Work}
\label{sec:relatedwork}

\citet{villar2015multi} and \citet{rafferty2019statistical} empirically illustrate that classical ordinary least squares (OLS) inference methods have inflated Type-1 error when used on data collected with  a variety of regret-minimizing multi-armed bandit algorithms. \citet{chen2020statistical} prove that the OLS estimator is asymptotically normal on data collected with an $\epsilon$-greedy algorithm, but their results do not cover settings in which there is no unique optimal policy, e.g., a multi-arm bandit with two identical arms (Appendix \ref{app:chen_paper}). 
Recent work has discussed the non-normality of OLS on data collected with bandit algorithms when there is no unique optimal policy and proposed alternative methods for statistical inference. A common thread between these methods is that they all utilize a form of \textit{adaptive weighting}. 
\citet{deshpande}  introduced the W-decorrelated estimator, which adjusts the OLS estimator with a sum of adaptively weighted residuals. In the multi-armed bandit setting, the W-decorrelated estimator up-weights observations from early in the study and down-weights observations from later in the study \citep{zhang2020inference}. In the batched bandit setting, \citet{zhang2020inference} show that the Z-statistics for the OLS estimators computed separately on each batch are jointly asymptotically normal. Standardizing the OLS statistic for each batch effectively adaptively re-weights the observations in each batch. 


\citet{athey} introduce adaptively weighted versions of both the standard augmented-inverse propensity weighted estimator (AW-AIPW) and the sample mean (AWA) for estimating parameters of simple models on data collected with bandit algorithms. They introduce a class  of adaptive ``variance stabilizing'' weights, for which 
the variance of a normalized version of their estimators converges in probability to a constant. 
In their discussion section they note 
open questions, two of which this work addresses:
1) ``What additional estimators can be used for normal inference with adaptively collected data?'' and 2) How do their results generalize to more complex sampling designs, like data collected with contextual bandit algorithms? We demonstrate that variance stabilizing adaptive weights can be used to modify a large class of M-estimators to guarantee valid inference. This generalization allows us to perform valid inference for a large class of important inferential targets: parameters of models for expected outcomes that are context dependent.

Recently, adaptive weighting has also been used in off-policy evaluation methods for when the behavior policy (policy used to collect the data) is a contextual bandit algorithm \citep{bibaut2021post,zhan2021off}. In this literature the estimand is the value, or average expected reward, of a pre-specified policy (note this is a scalar value). In contrast, in our work we are interested in constructing confidence regions for parameters of a model for an outcome (that could be the reward)---for example, this could be parameters of a logistic regression model for a binary outcome. We believe in the future there could be theory that could unify these adaptive weighting methods for these different estimands.

An alternative to using asymptotic approximations to construct confidence intervals is to use high-probability confidence bounds. These bounds provide stronger guarantees than those based on asymptotic approximations, as they are guaranteed to hold for finite samples. The downside is that these bounds are typically much wider, which is why much of classical statistics uses asymptotic approximations. Here we do the same. In Section \ref{sec:examples}, we empirically compare our to the self-normalized martingale bound \citep{abbasi2011improved}, a high-probability bound commonly used in the bandit literature.

\section{Simulation Results}
\label{sec:examples}


In this section, $R_t = Y_t$. 
We consider two settings: a continuous reward setting and a binary reward setting. 
In the continuous reward setting, the rewards are generated with mean $\E_{\MC{P}}[ R_t | X_t, A_t ] = \tilde{X}_t^\top \theta_0^*(\MC{P}) + A_t \tilde{X}_t^\top \theta_1^*(\MC{P})$ and noise drawn from a student's $t$ distribution with five degrees of freedom; here $\tilde{X}_t = [1, X_t] \in \real^3$ ($X_t$ with intercept term), actions $A_t \in \{ 0, 1 \}$, and parameters $\theta_0^*(\MC{P}), \theta_1^*(\MC{P}) \in \real^3$. 
In the binary reward setting, the reward $R_t$ is generated as a Bernoulli with success probability $\E_{\MC{P}}[ R_t | X_t, A_t ] = [ 1 + \exp ( - \tilde{X}_t^\top \theta_0^*(\MC{P}) - A_t \tilde{X}_t^\top \theta_1^*(\MC{P}) ) ]^{-1}$. 
Furthermore, in both simulation settings 
we set
 $\theta_0^*(\MC{P}) = [0.1,0.1,0.1]$ and $\theta_1^*(\MC{P}) = [0,0,0]$, so there is no unique optimal arm; we call vector parameter $\theta_1^*(\MC{P})$ the \textit{advantage} of selecting $A_t=1$ over $A_t=0$.
Also in both  settings, the contexts $X_t$ are drawn i.i.d. from a uniform distribution.

In both simulation settings we collect data using Thompson Sampling with a linear model for the expected reward and normal priors \citep{agrawal2013thompson} (so even when the reward is binary). 
We constrain the action selection probabilities with \textit{clipping} at a rate of $0.05$; this means that while typical Thompson Sampling produces action selection probabilities $\pi_t^{\TN{TS}}(A_t, X_t, \HH_{t-1})$, we instead use action selection probabilities $\pi_t(A_t, X_t, \HH_{t-1}) = 0.05 \vee \left( 0.95 \wedge \pi_t^{\TN{TS}}(A_t, X_t, \HH_{t-1}) \right)$ to select actions. We constrain the action selection probabilities in order to ensure weights $W_t$ are bounded when using a uniform stabilizing policy; see Sections \ref{sec:normality} and \ref{sec:discussion} for more discussion on this boundedness assumption. Also note that increasing the amount the algorithm explores (clipping) decreases the expected width of confidence intervals constructed on the resulting data (see Section \ref{sec:discussion}).

To analyze the data, in the continuous reward setting, we use least-squares estimators with a correctly specified model for the expected reward, i.e., M-estimators with $m_\theta (R_t, X_t, A_t) = - (R_t - \tilde{X}_t^\top \theta_0 - A_t \tilde{X}_t^\top \theta_1 )^2$. We consider both the unweighted and adaptively weighted versions. We also compare to the self-normalized martingale bound \citep{abbasi2011improved} and the W-decorrelated estimator \citep{deshpande}, as they were both developed for the linear expected reward setting. 
For the self-normalized martingale bound, which requires explicit bounds on the parameter space, we set $\Theta = \{ \theta \in \real^6 : \| \theta \|_2 \leq 6 \}$. 
In the binary reward setting, we also assume a correctly specified model for the expected reward.
We use both unweighted and adaptively weighted maximum likelihood estimators (MLEs), which correspond to an M-estimators with $m_\theta(R_t, X_t, A_t)$ set to the negative log-likelihood of $R_t$ given $X_t, A_t$. 
We solve for these estimators using Newton--Raphson optimization and do not put explicit bounds on the parameter space $\Theta$ (note in this case $m_\theta$ is concave in $\theta$ \citep[Chapter 5.4.2]{agresti2015foundations}). See Appendix \ref{app:simulationsall} for additional details and simulation results.

\begin{figure}[t]
	\centerline{ 
	\includegraphics[width=\linewidth]{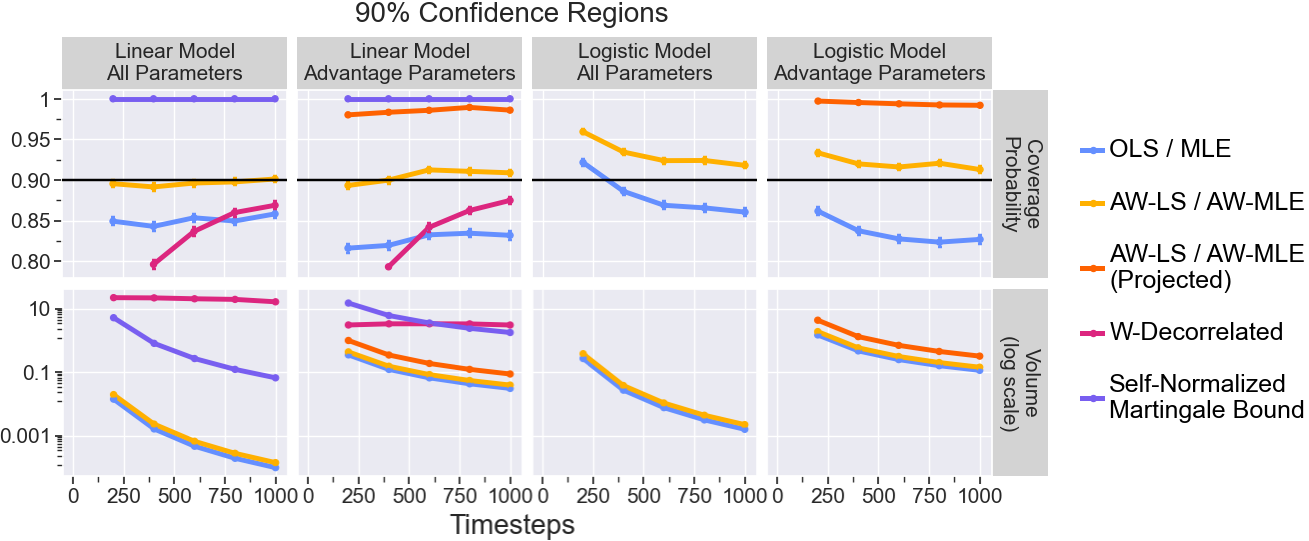}
	}
	\medskip
  \caption{Empirical coverage probabilities (upper row) and volume (lower row) of 90\% confidence ellipsoids. The left two columns are for the linear reward model setting (t-distributed rewards) and the right two columns are for the logistic regression model setting (Bernoulli rewards). We consider confidence ellipsoids for all parameters $\theta^*(\MC{P})$ and for advantage parameters $\theta_1^*(\MC{P})$ for both settings.}
  \label{fig:simulations}
\end{figure}

In Figure \ref{fig:simulations} we plot the empirical coverage probabilities and volumes of 90\% confidence regions for $\theta^*(\MC{P}) := [\theta_0^*(\MC{P}), \theta_1^*(\MC{P})]$ 
and $\theta_1^*(\MC{P})$ in both the continuous and binary reward settings. While the confidence regions based on the unweighted least-squares estimator (OLS) and the unweighted MLE have significant undercoverage that does not improve as $T$ increases, the confidence regions based on the adaptively weighted versions, AW-LS and AW-MLE, have very reliable coverage. 
For the confidence regions for $\theta_1^*(\MC{P})$ based on the AW-LS and AW-MLE, we include both projected confidence regions (for which we have theoretical guarantees) and non-projected confidence regions. The confidence regions based on projections are conservative but nevertheless have comparable volume to those based on OLS and MLE respectively. We do not prove theoretical guarantees for the non-projection confidence regions for AW-LS and AW-MLE, however they perform well across in our simulations.
Both types of confidence regions based on AW-LS have significantly smaller volumes than those constructed using the self-normalized martingale bound and W-decorrelated estimator. Note that the W-decorrelated estimator and self-normalized martingale bounds are designed for linear contextual bandits and are thus not applicable for the logistic regression model setting. The confidence regions constructed using the self-normalized martingale bound have reliable coverage as well, but are very conservative. Empirically, we found that the coverage probabilities of the confidence regions based on the W-decorrelated estimator were very sensitive to the choice of tuning parameters.
We use $5,000$ Monte-Carlo repetitions and the error bars plotted are standard errors.  


\section{Discussion}
\label{sec:discussion}


\bo{Immediate questions}
We assume that ratios $\pi_t^{\TN{sta}}(A_t, X_t) / \pi_t(A_t, X_t, \HH_{t-1})$ are bounded for our theoretical results; this precludes $\pi_t(A_t, X_t, \HH_{t-1})$ from going to zero for a fixed stabilizing policy. For simple models, e.g., the AW-LS estimator, we can let these ratios grow at a certain rate and still guarantee asymptotic normality (Appendix \ref{app:AWLS}); we conjecture similar results hold more generally. 


\bo{Generality and robustness}
This work
assumes that we have a well-specified model for the outcome $Y_t$, i.e., that $\theta^*(\MC{P}) \in \argmax_{\theta \in \Theta} \E_{\MC{P}} [ m_{\theta} ( Y_t, X_t, A_t ) | X_t, A_t ]$ w.p. $1$. 
Our theorems use this assumption to ensure that $\{ W_t \dot{m}_{\theta} ( Y_t, X_t, A_t ) \}_{t \geq 1}$ is a martingale difference sequence with respect to $\{ \HH_{t} \}_{t \geq 0}$. On i.i.d. data it is common to define $\theta^*(\MC{P})$ to be the best \emph{projected} solution, i.e., $\theta_0(\MC{P}) \in \argmax_{\theta \in \Theta} \E_{\MC{P}, \pi} \left[ m_{\theta} ( Y_t, X_t, A_t ) \right]$. Note that the best projected solution, $\theta^*(\MC{P})$, depends on the distribution of the action selection policy $\pi$. It would be ideal to also be able to perform inference for a projected solution on adaptively collected data.

Another natural question is whether adaptive weighting methods work in Markov Decision Processes (MDP) environments. Taking the AW-LS estimator introduced in Section \ref{sec:intuition} as an example, our conditional variance derivation in Equation~\eqref{eqn:variancestbilizingintuition} fails to hold in an MDP setting, specifically equality (c).
However, the conditional variance condition can be satisfied if we instead use weights $W_t = \{ [ \pi_t^{\TN{sta}}(A_t, X_t) p^{\TN{sta}}(X_t) ]/[ \pi_t(A_t, X_t, \HH_{t-1}) \PP_{\MC{P}}(X_t| X_{t-1}, A_{t-1}) ] \}^{1/2}$ where $\PP_{\MC{P}}$ are the state transition probabilities and $p^{\TN{sta}}$ is a pre-specified distribution over states.
In general though we do not expect to know the transition probabilities $\PP_{\MC{P}}$ and if we tried to estimate them, our theory would require the estimator to have error $o_p( 1 / \sqrt{T})$, \emph{below} the parametric rate.

\bo{Trading-off regret minimization and statistical inference objectives}
In sequential decision-making problems there is a fundamental trade-off between minimizing regret and minimizing estimation error for parameters of the environment using the resulting data \citep{bubeck2009pure,dean2018regret}. Given this trade-off there are many open problems regarding how to minimize regret while still guaranteeing a certain amount of power or expected confidence interval width, e.g., developing sample size calculators for use in justifying the number of users in a mobile health trial, and developing new adaptive algorithms  \citep{liu2014trading,erraqabi2017trading,yao2020power}.


\section*{Acknowledgements and Disclosure of Funding}
We thank Yash Nair for feedback on early drafts of this work.

Research reported in this paper was supported by National Institute on Alcohol Abuse and Al-coholism (NIAAA) of the National Institutes of Health under award number R01AA23187, National Institute on Drug Abuse (NIDA) of the National Institutes of Health under award number P50DA039838, National Cancer Institute (NCI) of the National Institutes of Health under award number U01CA229437, and by NIH/NIBIB and OD award number P41EB028242. The content is solely the responsibility of the authors and does not necessarily represent the official views of the National Institutes of Health.

This material is based upon work supported by the National Science Foundation Graduate Research Fellowship Program under Grant No.  DGE1745303.  Any opinions, findings, and conclusions or recommendations expressed in this material are those of the author(s) and do not necessarily reflect the views of the National Science Foundation.

\bibliography{adaptive_weighting_neurips_CR}

\clearpage
\appendix

\addcontentsline{toc}{section}{Appendix} 
\part{Appendix} 
\parttoc 


\input{appendix/simulations}

\clearpage

\input{appendix/m_estimator}

\clearpage

\input{appendix/stabilizing_policy}

\clearpage

\input{appendix/uniformity_stuff}

\clearpage

\input{appendix/chen_paper}

\end{document}

%% file: appendix/simulations.tex
\section{Simulations}
\label{app:simulationsall}

\subsection{Simulation Details}
\label{app:simulationdetails}

\paragraph{Simulation Environment}
\begin{itemize}
	\item Each dimension of $X_t$ is sampled independently from $\TN{Uniform}(0,5)$.
	\item $\theta^*(\MC{P}) = [\theta_0^*(\MC{P}), \theta_1^*(\MC{P})] = [0.1, 0.1, 0.1, 0, 0, 0]$, where $\theta_0^*(\MC{P}), \theta_1^*(\MC{P}) \in \real^3$. \\
	Below also include simulations where $[\theta_0^*(\MC{P}), \theta_1^*(\MC{P})] = [0.1, 0.1, 0.1, 0.2, 0.1, 0]$.
	\item t-Distributed rewards: $R_t | X_t, A_t \sim t_5 + \tilde{X}_t^\top \theta_0^*(\MC{P}) + A_t \tilde{X}_t^\top \theta_1^*(\MC{P})$, where $t_5$ is a t-distribution with $5$ degrees of freedom.
	\item Bernoulli rewards: $R_t | X_t, A_t \sim \TN{Bernoulli}( expit(\nu_t) )$ for $\nu_t = \tilde{X}_t^\top \theta_0^*(\MC{P}) + A_t \tilde{X}_t^\top \theta_1^*(\MC{P})$ and $expit(x) = \frac{1}{1 + \exp(-x)}$.
	\item Poisson rewards: $R_t | X_t, A_t \sim \TN{Poisson}( \exp(\nu_t) )$ for $\nu_t = \tilde{X}_t^\top \theta_0^*(\MC{P}) + A_t \tilde{X}_t^\top \theta_1^*(\MC{P})$.
\end{itemize}

\paragraph{Algorithm}
\begin{itemize}
	\item Thompson Sampling with $\N(0, I_d)$ priors on each arm.
	\item $0.05$ clipping
	\item Pre-processing rewards before received by algorithm:
	\begin{itemize}
		\item Bernoulli: $2 R_t - 1$
		\item Poisson: $0.6 R_t$
	\end{itemize}
\end{itemize}

\paragraph{Compute Time and Resources}
All simulations run within a few hours on a MacBook Pro.

\subsection{Details on Constructing of Confidence Regions}
\label{app:simCR}
For notational convenience, we define $Z_t = [\tilde{X}_t, A_t \tilde{X}_t]$.

\subsubsection{Least Squares Estimators}
\begin{itemize}
	\item $\hat{\theta}_T = \left( \sum_{t=1}^T W_t Z_t Z_t^\top \right)^{-1} \sum_{t=1}^T W_t Z_t R_t$
	\begin{itemize}
		\item For unweighted least squares, $W_t = 1$ and we call the estimator $\hat{\theta}_T^{\TN{OLS}}$.
		\item For adaptively weighted least squares, $W_t = \frac{1}{\sqrt{ \pi_t(A_t, X_t, \HH_{t-1}) } }$; this is equivalent to using square-root importance weights with a uniform stabilizing policy. We call the estimator $\hat{\theta}_T^{\TN{AW-LS}}$.
	\end{itemize}
	\item We assume homoskedastic errors and estimate the noise variance $\sigma^2$ as follows:
		\begin{equation*}
			\hat{\sigma}_T^2 = \frac{1}{T} \sum_{t=1}^T (R_t -Z_t^\top \hat{\theta}_T )^2.
		\end{equation*}
	\item We use a Hotelling t-squared test statistic to construct confidence regions for $\theta^*(\MC{P})$:
	\begin{multline}
		\label{eqn:LShotellingsT}
			C_T(\alpha) = \bigg\{ \theta \in \real^d : \left[ \hat{\Sigma}_T^{-1/2} \left( \frac{1}{T} \sum_{t=1}^T W_t Z_t Z_t^\top \right) 
			\sqrt{T} ( \hat{\theta}_T - \theta ) \right]^{\otimes 2} \\ 
			\leq \frac{d(T-1)}{T-d} F_{d,T-d}(1-\alpha) \bigg\}.
		\end{multline}
		\begin{itemize}
			\item For the unweighted least-squares estimator we use the following variance estimator: $\hat{\Sigma}_T = \hat{\sigma}_T^2 \frac{1}{T} \sum_{t=1}^T Z_t Z_t^\top$.
			\item For the AW-Least Squares estimator we use the following variance estimator: $\hat{\Sigma}_T = \hat{\sigma}_T^2 \frac{1}{T} \sum_{t=1}^T \frac{1}{\pi_t(A_t, X_t, \HH_{t-1})}^{A_t} \frac{1}{1-\pi_t(A_t, X_t, \HH_{t-1})}^{1-A_t} Z_t Z_t^\top$.
		\end{itemize}
	\item To construct (non-projected) confidence regions for $\theta_1^*( \MC{P} ) \in \real^{d_1}$ we treat the unweighted least squares / AW-LS estimators, $\hat{\theta}_{T, 1}$, as $\N \left( \theta_1^* (\MC{P}), \frac{1}{T} \left( \frac{1}{T} \sum_{t=1}^T W_t Z_t Z_t^\top \right)^{-1} \hat{\Sigma}_T \left( \frac{1}{T} \sum_{t=1}^T W_t Z_t Z_t^\top \right)^{-1} \right)$. We use a Hotelling t-squared test statistic to construct confidence regions for $\theta_1^*(\MC{P})$:
	\begin{equation*}
			C_T(\alpha) = \left\{ \theta_1 \in \real^{d_1} : \left[ V_{1,T}^{-1/2} \sqrt{T} ( \hat{\theta}_{T,1} - \theta_1 ) \right]^{\otimes 2} \leq \frac{d_1(T-1)}{T-d_1} F_{d_1,T-d_1}(1-\alpha) \right\},
		\end{equation*}
		where $V_{1,T}$ is the lower right $d_1 \times d_1$ block of matrix $\left( \frac{1}{T} \sum_{t=1}^T W_t Z_t Z_t^\top \right)^{-1} \hat{\Sigma}_T \left( \frac{1}{T} \sum_{t=1}^T W_t Z_t Z_t^\top \right)^{-1}$. Recall that for the unweighted least squares estimator $W_t = 1$ and for AW-LS $W_t = \frac{1}{\sqrt{ \pi_t(A_t, X_t, \HH_{t-1}) } }$.
	\item For the AW-least squares estimator, we also construct projected confidence regions for $\theta_1^*(\MC{P})$ using the confidence region defined in equation \eqref{eqn:LShotellingsT}. See Section \ref{app:projCR} below for more details on constructing projected confidence regions.
\end{itemize}

\subsubsection{MLE Estimators}

\begin{center}
\begin{tabular}{ |c|c|c|c|c|c|c| } 
 \hline
 \textbf{Distribution} & $\bs{\nu}$ & $\bs{b(\nu)}$ & $\bs{b'(\nu)}$ & $\bs{b''(\nu)}$  & $\bs{b'''(\nu)}$ \\
 \hline
 $\N(\mu, 1)$ & $\mu$ 										 	& $\frac{1}{2} \nu^2$			& $\nu = \mu$ 										& $1$ & $0$\\ 
  \hline
 Poisson($\lambda$)    & $\log \lambda$ 									& $\exp(\nu)$ 					& $\exp(\nu) = \lambda$ 						& $\exp(\nu) = \lambda$ & $\exp(\nu) = \lambda$\\ 
  \hline
 Bernoulli($p$)          & $\log \big( \frac{p}{1-p} \big)$ 			& $\log(1 + e^\nu )$ 	& $\frac{e^\nu }{1+e^\nu } = p$	& $\frac{e^\nu }{(1+e^\nu )^2} = p(1-p)$ & $p(1-p)(1-2p)$ \\ 
 \hline
\end{tabular}
\end{center}

\begin{itemize}
	\item $\thetahat_T$ is the root of the score function:
		\begin{equation*}
			0 = \sum_{t=1}^T W_t \left( R_t - b'(\thetahat_T^\top Z_t) \right) Z_t.
		\end{equation*}
		We use Newton Raphson optimization to solve for $\hat{\theta}_T$.
		\begin{itemize}
			\item For unweighted MLE, $W_t = 1$.
			\item For AW-MLE, $W_t = \frac{1}{\sqrt{ \pi_t(A_t, X_t, \HH_{t-1}) } }$; this is equivalent to using square-root importance weights with a uniform stabilizing policy.
		\end{itemize}
	\item Second derivative of score function: $- \sum_{t=1}^T b''(\thetahat_T^\top Z_t) Z_t Z_t^\top$.
	\item We use a Hotelling t-squared test statistic to construct confidence regions for $\theta^*(\MC{P})$:
	\begin{multline}
		\label{eqn:GLMhotellingsT}
			C_T(\alpha) = \bigg\{ \theta \in \real^d : \left[ \hat{\Sigma}_T^{-1/2} \left( \frac{1}{T} \sum_{t=1}^T W_t b''(\thetahat_T^\top Z_t) Z_t Z_t^\top \right) \sqrt{T} ( \hat{\theta}_T - \theta ) \right]^{\otimes 2} \\
			\leq \frac{d(T-1)}{T-d} F_{d,T-d}(1-\alpha) \bigg\}.
		\end{multline}
		\begin{itemize}
			\item For the MLE variance estimator, we use $\hat{\Sigma}_T = \frac{1}{T} \sum_{t=1}^T b''(\thetahat_T^\top Z_t) Z_t Z_t^\top$.
			\item For the AW-MLE variance estimator, we use $\hat{\Sigma}_T = \frac{1}{T} \sum_{t=1}^T \frac{1}{\pi_t(A_t, X_t, \HH_{t-1})}^{A_t} \frac{1}{1-\pi_t(A_t, X_t, \HH_{t-1})}^{1-A_t} b''(\thetahat_T^\top Z_t) Z_t Z_t^\top$.
		\end{itemize}
	\item To construct (non-projected) confidence regions for $\theta_1^*(\MC{P}) \in \real^{d_1}$ we treat the MLE / AW-MLE estimators, $\hat{\theta}_{T, 1}$, as $\N \left( \theta_1^*(\MC{P}), \frac{1}{T} \left( \frac{1}{T} \sum_{t=1}^T W_t b''(\thetahat_T^\top Z_t) Z_t Z_t^\top \right) \hat{\Sigma}_T^{-1} \left( \frac{1}{T} \sum_{t=1}^T W_t b''(\thetahat_T^\top Z_t) Z_t Z_t^\top \right) \right)$. We use a Hotelling t-squared test statistic to construct confidence regions for $\theta_1^*(\MC{P})$:
	\begin{equation*}
			C_T(\alpha) = \left\{ \theta_1 \in \real^{d_1} :  \left[ V_{1,T}^{-1/2} \sqrt{T} ( \hat{\theta}_{T,1} - \theta_1 ) \right]^{\otimes 2 } 
			\leq \frac{d_1(T-1)}{T-d_1} F_{d_1,T-d_1}(1-\alpha) \right\},
		\end{equation*}
		where $V_{1,T}$ is the lower right $d_1 \times d_1$ block of matrix $\left( \frac{1}{T} \sum_{t=1}^T W_t b''(\thetahat_T^\top Z_t) Z_t Z_t^\top \right) \hat{\Sigma}_T^{-1} \left( \frac{1}{T} \sum_{t=1}^T W_t b''(\thetahat_T^\top Z_t) Z_t Z_t^\top \right)$.
	\item For the AW-MLE estimator, we also construct projected confidence regions for $\theta_1^*(\MC{P})$ using the confidence region defined in equation \eqref{eqn:GLMhotellingsT}. See Section \ref{app:projCR} below for more details on constructing projected confidence regions.
\end{itemize}


\subsubsection{W-Decorrelated}
The following is based on Algorithm 1 of \citet{deshpande}.
\begin{itemize}
	\item The W-decorrelated estimator for $\theta^*(\MC{P})$ is constructed as follows with adaptive weights for $W_t \in \real^d$:
	\begin{equation*}
		\hat{\theta}_T^{\TN{WD}} = \hat{\theta}_T^{\TN{OLS}} + \sum_{t=1}^T W_t (R_t - \tilde{X}_t^\top \hat{\theta}_T^{\TN{OLS}}).
	\end{equation*}
	\item The weights are set as follows: \\
	$W_1 = 0 \in \real^d$ and $W_t = ( I_d - \sum_{s=1}^t \sum_{u=1}^t W_s Z_u^\top ) Z_t \frac{1}{ \lambda_T + \| Z_t \|_2^2 }$ for $t > 1$.
	\item We choose $\lambda_T = \TN{mineig}_{0.01}(Z_t Z_t^\top) / \log T$ and $\TN{mineig}_{\alpha}(Z_t Z_t^\top)$ represents the $\alpha$ quantile of the minimum eigenvalue of $Z_t Z_t^\top$. This is similar to the procedure used in the simulations of \citet{deshpande} and is guided by Proposition 5 in their paper.
	\item We assume homoskedastic errors and estimate the noise variance $\sigma^2$ as follows:
	\begin{equation*}
			\hat{\sigma}_T^2 = \frac{1}{T} \sum_{t=1}^T (R_t -Z_t^\top \hat{\theta}_T^{\OLS} )^2.
		\end{equation*}
	\item To construct confidence ellipsoids for $\theta^*(\MC{P})$ are constructed using a Hotelling t-squared statistic:
	\begin{equation*}
		C_T(\alpha) = \left\{ \theta \in \real^d : (\hat{\theta}_T^{\TN{WD}} - \theta)^\top V_T^{-1} (\hat{\theta}_T^{\TN{WD}} - \theta) \leq \frac{d(T-1)}{T-d} F_{d,T-d}(1-\alpha) \right\}
	\end{equation*}
	where $V_T = \hat{\sigma}_T^2 \sum_{t=1}^T W_t W_t^\top$.
	\item To construct confidence ellipsoids for $\theta_1^*(\MC{P}) \in \real^{d_1}$ with the following confidence ellipsoid where $V_{T,1}$ is the lower right $d_1 \times d_1$ block of matrix $V_T$:
	\begin{equation*}
		C_T(\alpha) = \left\{ \theta_1 \in \real^{d_1} : (\hat{\theta}_{T,1}^{\TN{WD}} - \theta_1)^\top V_{T,1}^{-1} (\hat{\theta}_{T,1}^{\TN{WD}} - \theta_1 ) \leq \frac{d_1(T-1)}{T-d_1} F_{d_1,T-d_1}(1-\alpha) \right\}.
	\end{equation*}
\end{itemize}

\subsubsection{Self-Normalized Martingale Bound}

We construct $1-\alpha$ confidence region using the following equation taken from Theorem 2 of \cite{abbasi2011improved}:
\begin{equation*}
	C_T(\alpha) = \left\{ \theta \in \Theta : ( \thetahat_T - \theta )^\top V_T ( \thetahat_T - \theta ) 
	\leq \sigma \sqrt{2 \log \left( \frac{\det(V_T)^{1/2} \det(\lambda I_d)^{-1/2} }{\alpha} \right) } + \lambda^{1/2} S \right\}.
\end{equation*}

\begin{itemize}
	\item $\hat{\theta}_T = \left( \lambda I_d + \sum_{t=1}^T Z_t Z_t^\top \right)^{-1} \sum_{t=1}^T Z_t R_t$.
	\item $V_T = I_d \lambda + \sum_{t=1}^T Z_t Z_t^\top$.
	\item $\lambda = 1$ (ridge regression regularization parameter).
	\item $\sigma = 1$ (assumes rewards are $\sigma$-subgaussian).
	\item $S = 6$, where it is assumed that $\| \theta^*(\MC{P}) \| \leq S$ (recall that in our simulations $\theta^*(\MC{P}) \in \real^6$).
	\item $\Theta = \{ \theta \in \real^6 : \| \theta \|_2 \leq 6 \}$.
	\item For constructing confidence regions for $\theta^*(\MC{P})$, we use projected confidence regions. \end{itemize}

\subsubsection{Construction of Projected Confidence Regions}
\label{app:projCR}
We are interested in getting the confidence ellipsoid of the projection of a $d$-dimensional ellipsoid onto $p$-dimensional space, for $p < d$.

\begin{itemize}
	\item Defining the original $d$-dimensional ellipsoid, for $\boldx \in \real^d$ and $\boldB \in \real^{d \by d}$:
		\begin{equation*}
			\boldx^\top \boldB \boldx = 1
		\end{equation*}
	\item Partitioning the matrix $\boldB$ and vector $\boldx$: \\
		For $y \in \real^{d-p}$ and $z \in \real^p$.
		\begin{equation*}
			\boldx = \begin{bmatrix} \boldy \\ \boldz \end{bmatrix}
		\end{equation*}
		For $\boldC \in \real^{d-p \by d-p}$, $\boldE \in \real^{p \by p}$, and $\boldD \in \real^{d-p \by p}$.
		\begin{equation*}
			\boldB = \begin{bmatrix}
				\boldC & \boldD \\
				\boldD^\top & \boldE
			\end{bmatrix}
		\end{equation*}
	\item Gradient of $\boldx^\top \boldB \boldx$ with respect to $\boldx$:
		\begin{equation*}
			(\boldB + \boldB^\top) \boldx
			= 2 \boldB \boldx
			= \begin{bmatrix}
				\boldC & \boldD \\
				\boldD^\top & \boldE
			\end{bmatrix} 
			\begin{bmatrix} \boldy \\ \boldz \end{bmatrix}.
		\end{equation*}
		Since we are projecting onto the p-dimensional space, our projection is such that the gradient of $\boldx^\top \boldB \boldx$ with respect to $\boldy$ is zero, which means
		\begin{equation*}
			\boldC \boldy + \boldD \boldz = 0.
		\end{equation*}
		This means in the projection that $\boldy = -\boldC^{-1} \boldD \boldz$.
	\item Returning to our definition of the ellipsoid, plugging in $\boldz$, we have that
	\begin{equation*}
		1 = \boldx^\top \boldB \boldx
		= \begin{bmatrix} \boldy^\top & \boldz^\top \end{bmatrix}
		\begin{bmatrix}
				\boldC & \boldD \\
				\boldD^\top & \boldE
			\end{bmatrix} 
		\begin{bmatrix} \boldy \\ \boldz \end{bmatrix}
		= \boldy^\top \boldC \boldy
		+ 2 \boldz^\top \boldD^\top \boldy
		+ \boldz^\top \boldE \boldz
	\end{equation*}
	\begin{equation*}
		= (\boldC^{-1} \boldD \boldz)^\top \boldC (\boldC^{-1} \boldD \boldz)
		- 2 \boldz^\top \boldD^\top ( \boldC^{-1} \boldD \boldz )
		+ \boldz^\top \boldE \boldz
	\end{equation*}
	\begin{equation*}
		= \boldz^\top \boldD^\top \boldC^{-1} \boldD \boldz
		- 2 \boldz^\top \boldD^\top \boldC^{-1} \boldD \boldz
		+ \boldz^\top \boldE \boldz
	\end{equation*}
	\begin{equation*}
		= \boldz^\top (\boldE - \boldD^\top \boldC^{-1} \boldD ) \boldz.
	\end{equation*}
	Thus the equation for the final projected ellipsoid is
	\begin{equation*}
		\boldz^\top (\boldE - \boldD^\top \boldC^{-1} \boldD ) \boldz = 1.
	\end{equation*}
\end{itemize}

\subsection{Additional Simulation Results}
\label{app:simulations}

In addition to the continuous reward and a binary reward settings, here we also consider a discrete count reward setting. 
In this discrete reward setting, the reward $R_t$ is generated from a Poisson distribution with expectation $\E_{\MC{P}}[ R_t | X_t, A_t ] = \exp ( \tilde{X}_t^\top \theta_0^*(\MC{P}) - A_t \tilde{X}_t^\top \theta_1^*(\MC{P}) )$. All other data generation methods are equivalent to those used for the other simulation settings. Additionally we will consider the setting in which $\theta^*(\MC{P}) = [0.1, 0.1, 0.1, 0.2, 0.1, 0]$ for the continuous reward, binary reward, and discrete count settings.

To analyze the data, in the discrete count reward setting, we assume a correctly specified model for the expected reward.
We use both unweighted and adaptively weighted maximum likelihood estimators (MLEs), which correspond to an M-estimators with $m_\theta(R_t, X_t, A_t)$ set to the negative log-likelihood of $R_t$ given $X_t, A_t$. 
We solve for these estimators using Newton--Raphson optimization and do not put explicit bounds on the parameter space $\Theta$.

\begin{figure}[h]
	\centerline{ 
	    \includegraphics[width=0.42\linewidth]{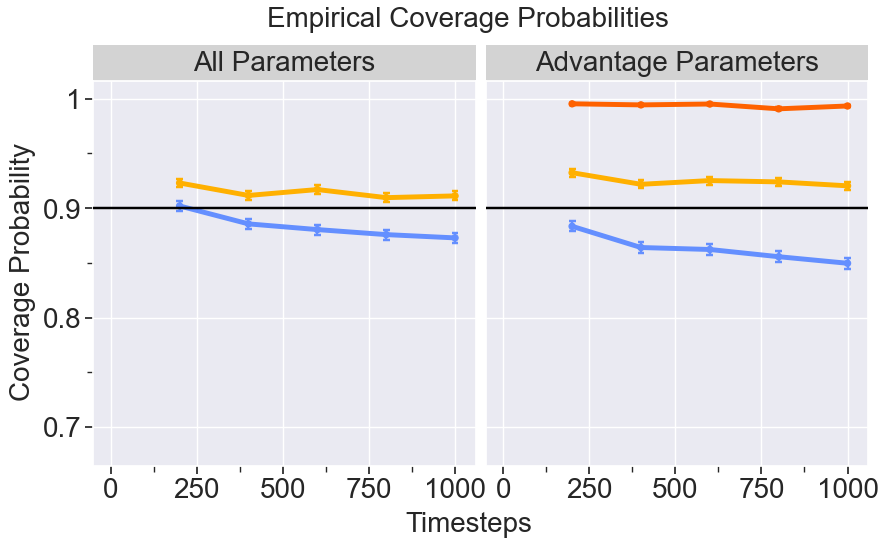}
	    ~~~
	    \includegraphics[width=0.58\linewidth]{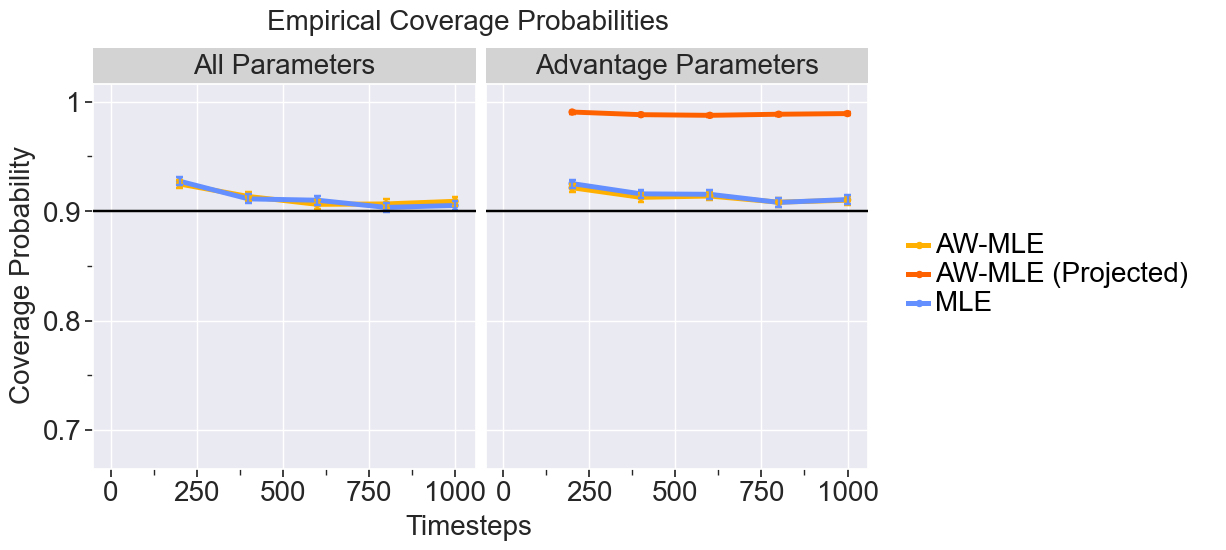}
	}
	\centerline{ 
	    \includegraphics[width=0.42\linewidth]{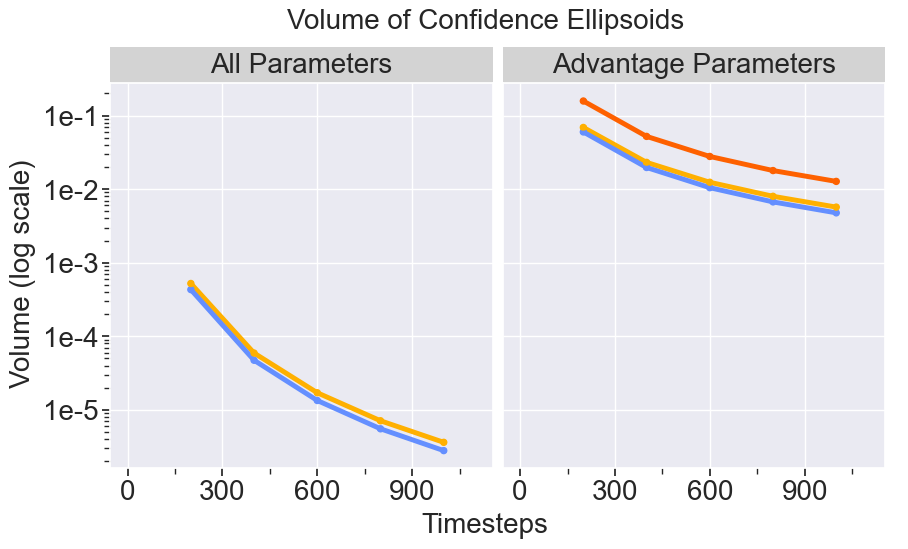}
	    ~~~
	    \includegraphics[width=0.58\linewidth]{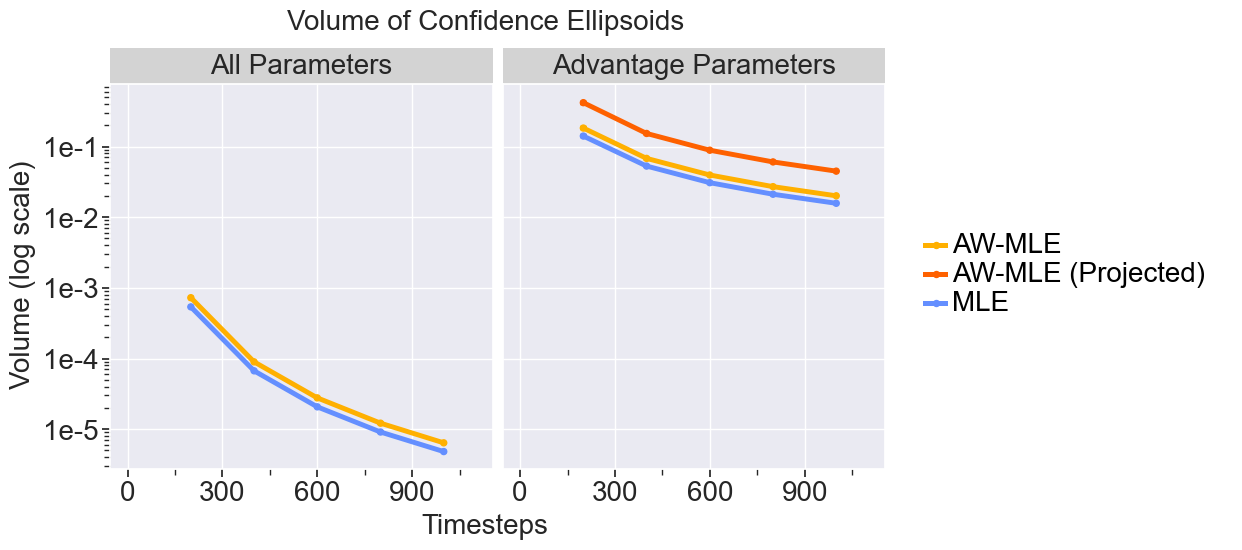}
	}
	\caption{\bo{Poisson Rewards:} Empirical coverage probabilities for 90\% confidence ellipsoids for parameters $\theta^*(\MC{P})$ and parameters $\theta_1^*(\MC{P})$ (top row). We also plot the volumes of these 90\% confidence ellipsoids for $\theta^*(\MC{P})$ and parameters $\theta_1^*(\MC{P})$ (bottom row).
	We set the true parameters to $\theta^*(\MC{P}) = [0.1, 0.1, 0.1, 0, 0, 0]$ (left) and to $\theta^*(\MC{P}) = [0.1, 0.1, 0.1, 0.2, 0.1, 0]$ (right). }
  \label{fig:poisson_CR}
\end{figure}

\clearpage

\begin{figure}[t]
	\centerline{ 
	\includegraphics[width=\linewidth]{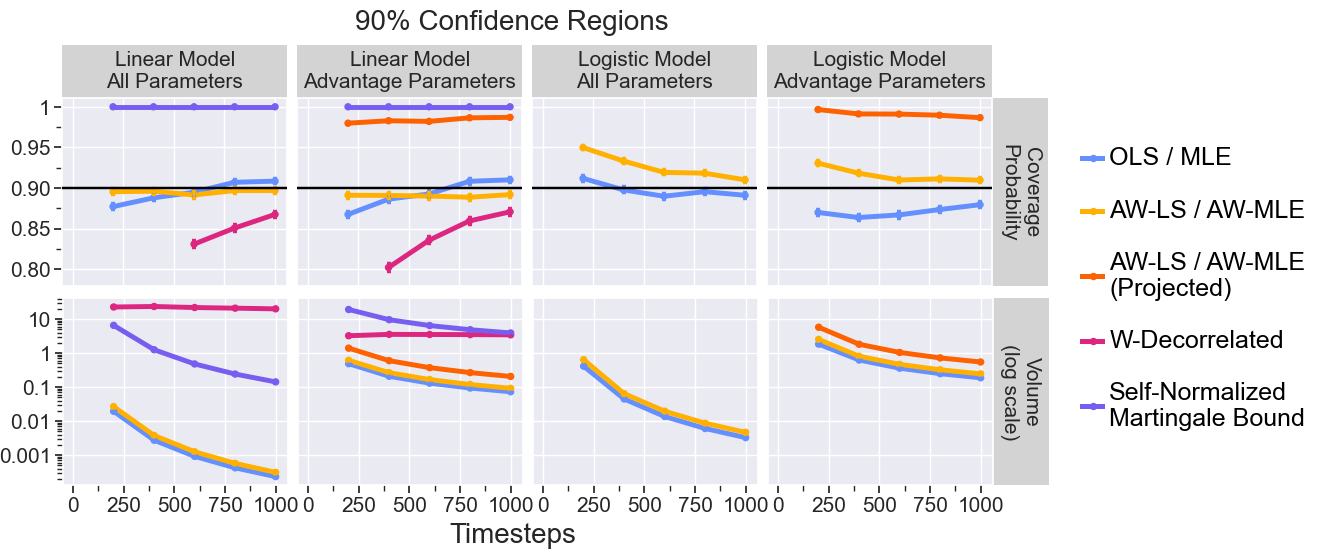}
	}
	\medskip
  \caption{Empirical coverage probabilities (upper row) and volume (lower row) of 90\% confidence ellipsoids. In these simulations, $\theta^*(\MC{P}) = [0.1, 0.1, 0.1, 0.2, 0.1, 0]$. The left two columns are for the linear reward model setting (t-distributed rewards) and the right two columns are for the logistic regression model setting (Bernoulli rewards). We consider confidence ellipsoids for all parameters $\theta^*(\MC{P})$ and for advantage parameters $\theta_1^*(\MC{P})$ for both settings.}
  \label{fig:simulations}
\end{figure}

\begin{figure}[H]
	\centerline{ 
	    \includegraphics[width=0.43\linewidth]{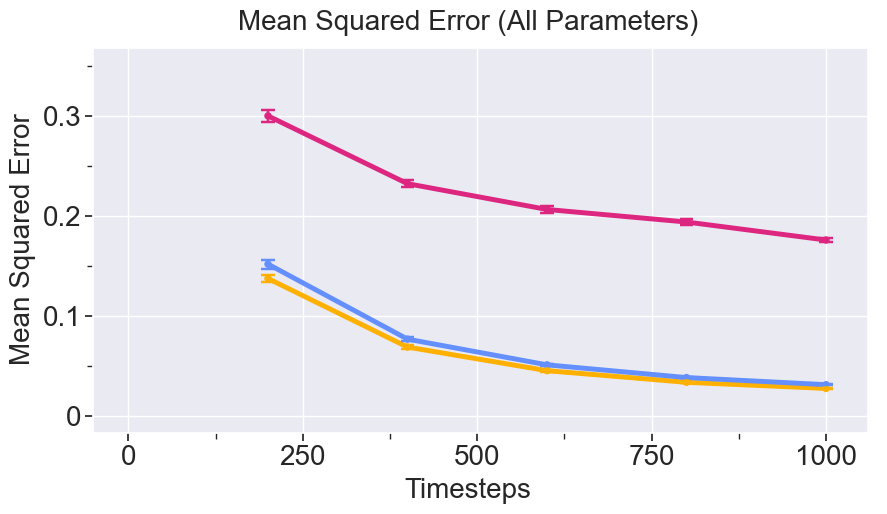}
	    ~~~
	    \includegraphics[width=0.57\linewidth]{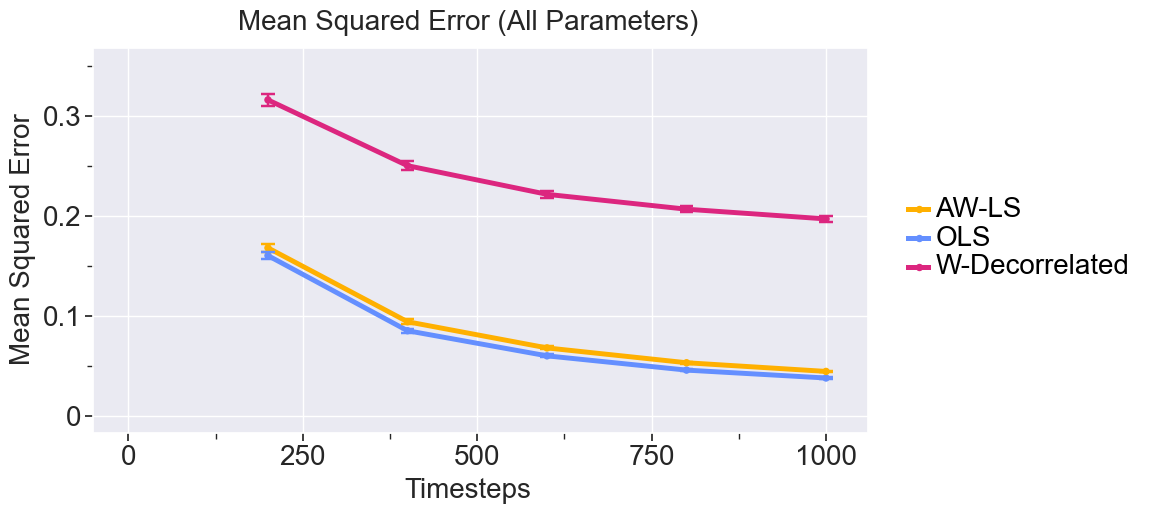}
	}
	\centerline{ 
	    \includegraphics[width=0.45\linewidth]{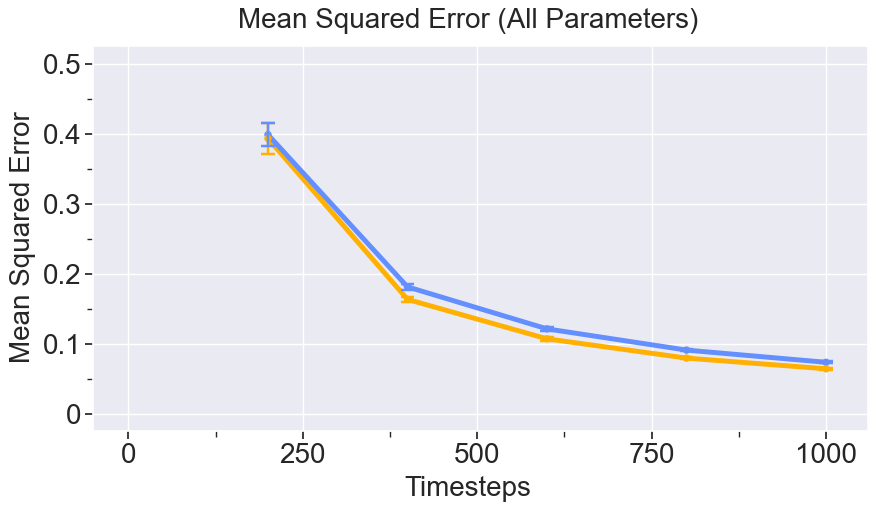}
	    ~~~
	    \includegraphics[width=0.55\linewidth]{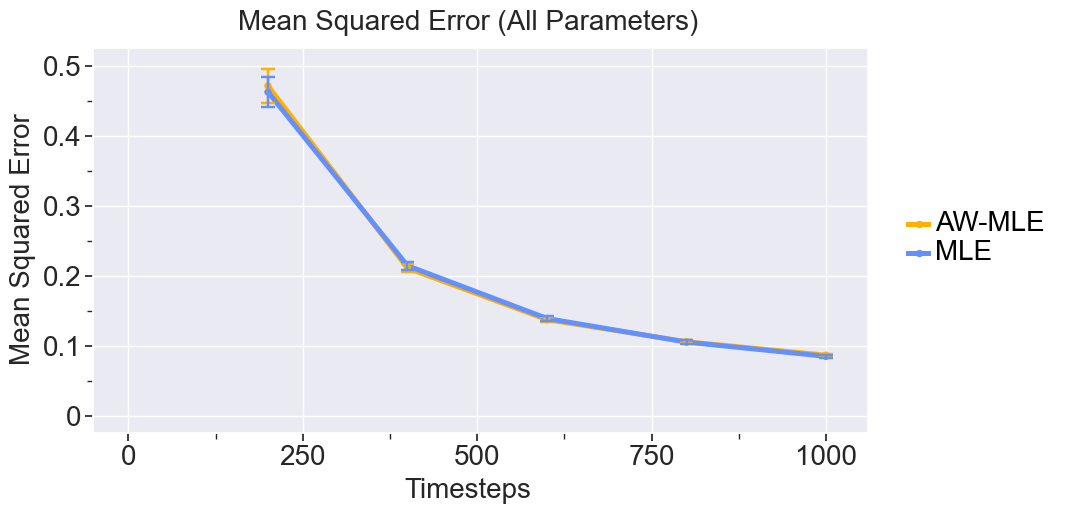}
	}
	\centerline{ 
	    \includegraphics[width=0.45\linewidth]{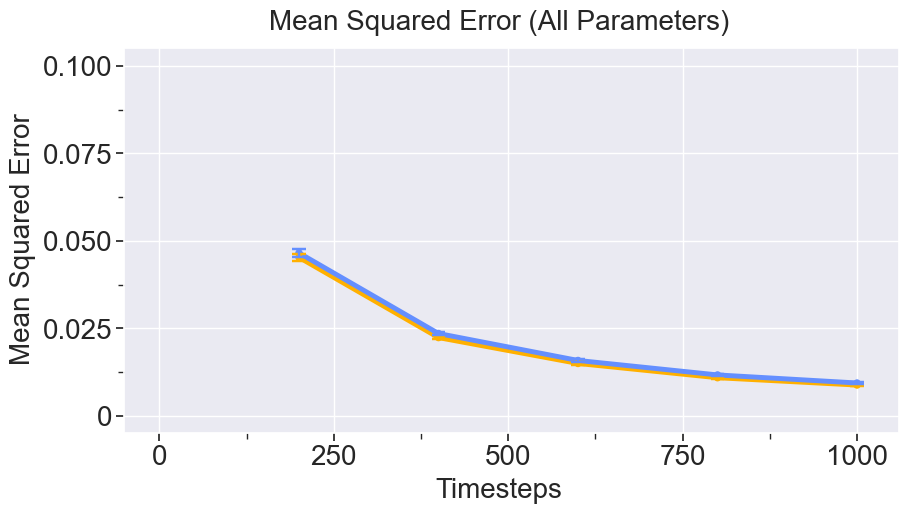}
	    ~~~
	    \includegraphics[width=0.55\linewidth]{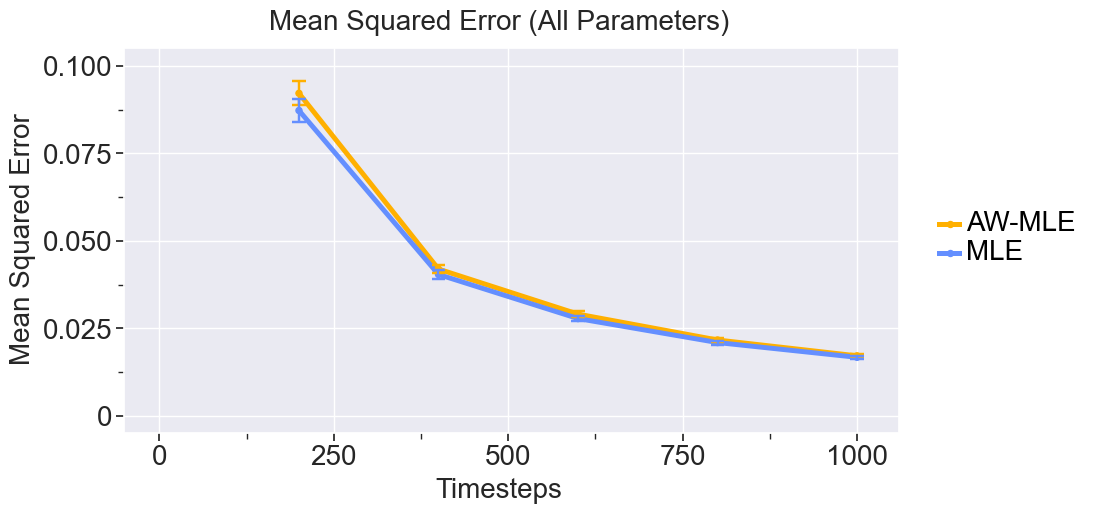}
	}
  \caption{Mean squared error estimators of $\theta^*(\MC{P})$ for linear model (top), logistic regression model (middle), and generalized linear model for Poisson rewards (bottom). We consider simulations with $\theta^*(\MC{P}) = [0.1, 0.1, 0.1, 0, 0, 0]$ (left) and simulations with $\theta^*(\MC{P}) = [0.1, 0.1, 0.1, 0.2, 0.1, 0]$ (right).}
  \label{fig:GLM_rewards_MSE}
\end{figure}

In Figure \ref{fig:GLM_rewards_MSE}, we plot the mean squared errors of all estimators for all three simulation settings (same simulation hyperparameters as described previously for the respective simulation settings).

%% file: appendix/m_estimator.tex
\section{Asymptotic Results}
\label{app:asymptoticresults}

Throughout, $\| \cdot \|$ refers to the $L_2$ norm. 

\subsection{Definitions}

Here we define convergence in probability and distribution that is uniform over the true parameter. We follow the definitions are based on those in \citet{kasy2019uniformity} and \citet[Chapter 1.12]{van1996weak}.

\begin{definition}[Uniform Convergence in Probability]
	\label{def:unifprob}
	Let $\{ Z_T(\MC{P}) \}_{T \geq 1}$ be a sequence of random variables whose distributions are defined by some $\MC{P} \in \boldP$ and some nuisance component $\eta$.
	We say that $Z_T(\MC{P}) \Pto c$ uniformly over $\MC{P} \in \boldP$ as $T \to \infty$ if for any $\epsilon > 0$,
	\begin{equation}
		\label{eqn:uniformPdef}
		\sup_{ \MC{P} \in \boldP } \PP_{\MC{P}, \eta} \left( \| Z_T(\MC{P}) - c \| > \epsilon \right) \to 0.
	\end{equation}
	For simplicity of notation, throughout we denote $Z_T(\MC{P}) - c = o_{\MC{P} \in \boldP }(1)$ to mean $Z_T(\MC{P}) \Pto c$ uniformly over $\MC{P} \in \boldP$ as $T \to \infty$.
\end{definition}

\begin{definition}[Uniformly Stochastically Bounded]
	\label{def:}
	Let $\{ Z_T(\MC{P}) \}_{T \geq 1}$ be a sequence of random variables whose distributions are defined by some $\MC{P} \in \boldP$ and some nuisance component $\eta$.
	We say that $Z_T(\MC{P})$ is uniformly stochastically bounded over $\MC{P} \in \boldP$ as $T \to \infty$ if for any $\epsilon > 0$ there exists some $k < \infty$ such that
	\begin{equation*}
		\label{eqn:uniformBigO}
		\limsup_{T \to \infty} \sup_{\MC{P} \in \boldP} \PP_{\MC{P}, \eta} \left( \| Z_T(\MC{P}) \| > k \right) < \epsilon.
	\end{equation*}
	Similarly we denote $Z_T(P) = O_{\MC{P} \in \boldP}(1)$ to mean $Z_T(\MC{P})$ is stochastically bounded uniformly over $\MC{P} \in \boldP$ as $T \to \infty$.
\end{definition}

\begin{definition}[Uniform Convergence in Distribution]
	\label{def:unifdist}
	Let $Z(\MC{P}) \in \real^{d_Z}$ and $\{ Z_T(\MC{P}) \}_{T \geq 1}  \in \real^{d_Z}$ be a sequence of random variables whose distributions are defined by some $\MC{P} \in \boldP$ and some nuisance component $\eta$.
	We say that $Z_T(\MC{P}) \Dto Z(\MC{P})$ uniformly over $\MC{P} \in \boldP$  as $T \to \infty$ if 
	\begin{equation}
		\label{eqn:uniformDdef}
		\sup_{\MC{P} \in \boldP} \sup_{f \in BL_1} \bigg| \E_{\MC{P}, \eta} \left[ f \left( Z_T(\MC{P}) \right) \right] - \E_{\MC{P}, \eta} \left[ f \left( Z(\MC{P}) \right) \right] \bigg| \to 0,
	\end{equation}
	where $BL_1$ is the set of functions $f : \real^{d_z} \to \real$ with $\| f(z) \|_\infty \leq 1$ and $| f(z) - f(z') | \leq \| z - z' \|$ for all $z, z' \in \real^{d_Z}$.
\end{definition}

As discussed in \citet{kasy2019uniformity},  Equation~\eqref{eqn:uniformPdef} holds if and only if for any $\epsilon > 0$ and any sequence $\{ \MC{P}_T \}_{T \geq 1}$ such that $\MC{P}_T \in \boldP$ for all $T \geq 1$, $\PP_{\MC{P}_T, \eta} \left( \| Z_T(\MC{P}_T) - c \| > \epsilon \right) \to 0$. 

Similarly,  Equation~\eqref{eqn:uniformDdef} holds if and only if for any sequence $\{ \MC{P}_T \}_{T \geq 1}$ such that $\MC{P}_T \in \boldP$ for all $T \geq 1$, $\sup_{f \in BL_1} \bigg| \E_{\MC{P}_T, \eta} \left[ f \left( Z_T(\MC{P}_T) \right) \right] - \E_{\MC{P}_T, \eta} \left[ f \left( Z(\MC{P}_T) \right) \right] \bigg| \to 0$.

\subsection{Consistency}

We prove the first part of Theorem \ref{thm:normality}, i.e., that $\thetahat_T \Pto \theta^*(\MC{P})$ uniformly over $\MC{P} \in \boldP$.
We abbreviate $m_{\theta}(Y_t, X_t, A_t)$ with $m_{\theta, t}$. 
By definition of $\thetahat_T$,
\begin{equation*}
	\sum_{t=1}^T W_t m_{\thetahat_T, t}
	= \sup_{\theta \in \Theta} \sum_{t=1}^T W_t m_{\theta, t} 
	\geq \sum_{t=1}^T W_t m_{\theta^*(\MC{P}), t}.
\end{equation*}
Note that $\| \thetahat_T - \theta^*(\MC{P}) \| > \epsilon > 0$ implies that
\begin{equation*}
	\sup_{\theta \in \Theta : \|\theta-\theta^*(\MC{P})\| > \epsilon} \sum_{t=1}^T W_t m_{\theta, t}
	= \sup_{\theta \in \Theta} \sum_{t=1}^T W_t m_{\theta, t}.
\end{equation*}
Thus, the above two results imply the following inequality:
\begin{equation*}
	\sup_{\MC{P} \in \boldP} \PP_{\MC{P}, \pi} \left( \|\thetahat_T-\theta^*(\MC{P})\| > \epsilon \right)
	\leq \sup_{\MC{P} \in \boldP} \PP_{\MC{P}, \pi} \left( \sup_{\theta \in \Theta:\|\theta-\theta^*(\MC{P})\| > \epsilon} \sum_{t=1}^T W_t m_{\theta,t}
	\geq \sum_{t=1}^T W_t m_{\theta^*(\MC{P}),t} \right) 
\end{equation*}
\begin{equation*}
	= \sup_{\MC{P} \in \boldP} \PP_{\MC{P}, \pi} \left( \sup_{\theta \in \Theta:\|\theta-\theta^*(\MC{P})\| > \epsilon} \left\{ \frac{1}{T} \sum_{t=1}^T W_t m_{\theta,t} \right\}
	- \frac{1}{T} \sum_{t=1}^T W_t m_{\theta^*(\MC{P}),t} \geq 0 \right)
\end{equation*}
\begin{multline*}
	= \sup_{\MC{P} \in \boldP} \PP_{\MC{P}, \pi} \bigg( \sup_{\theta \in \Theta:\|\theta-\theta^*(\MC{P})\| > \epsilon} \bigg\{ \frac{1}{T} \sum_{t=1}^T W_t m_{\theta,t} 
	- \E_{\MC{P}, \pi} [ W_t m_{\theta,t} | \HH_{t-1} ] + \E_{\MC{P}, \pi} [ W_t m_{\theta,t} | \HH_{t-1} ] \bigg\} \\
	- \frac{1}{T} \sum_{t=1}^T \bigg\{ W_t m_{\theta^*(\MC{P}),t} 
	- \E_{\MC{P}, \pi} [ W_t m_{\theta^*(\MC{P}),t} | \HH_{t-1} ] 
	+ \E_{\MC{P}, \pi} [ W_t m_{\theta^*(\MC{P}),t} | \HH_{t-1} ] \bigg\}
	\geq 0 \bigg).
\end{multline*}

By triangle inequality,
\begin{multline}
	\label{eqn:abcprob}
	\leq \sup_{\MC{P} \in \boldP} \PP_{\MC{P}, \pi} \bigg( \underbrace{ \sup_{\theta \in \Theta:\|\theta-\theta^*(\MC{P})\| > \epsilon} \left\{ \frac{1}{T} \sum_{t=1}^T \left( W_t m_{\theta,t} 
	- \E_{\MC{P}, \pi} [ W_t m_{\theta,t} | \HH_{t-1} ] \right) \right\} }_{(a)} \\
	+ \underbrace{ \sup_{\theta \in \Theta:\|\theta-\theta^*(\MC{P})\| > \epsilon} \left\{ \frac{1}{T} \sum_{t=1}^T \E_{\MC{P}, \pi} \left[ W_t ( m_{\theta,t} - m_{\theta^*(\MC{P}),t} ) \big| \HH_{t-1} \right] \right\} }_{(b)} \\
	- \underbrace{ \frac{1}{T} \sum_{t=1}^T \bigg\{ W_t m_{\theta^*(\MC{P}),t} - \E_{\MC{P}, \pi} [ W_t m_{\theta^*(\MC{P}),t} | \HH_{t-1} ] \bigg\} }_{(c)}
	\geq 0 \bigg) \to 0.
\end{multline}

We now show that the limit in Equation~\eqref{eqn:abcprob} above holds.
\begin{itemize}
	\item Regarding term (c), by moment bounds of Condition \ref{cond:moments} and Lemma \ref{lem:wllnW}, \\
	$\frac{1}{T} \sum_{t=1}^T \big\{ W_t m_{\theta^*(\MC{P}),t} - \E_{\MC{P}, \pi} [ W_t m_{\theta^*(\MC{P}),t} | \HH_{t-1} ] \big\} = o_{\MC{P} \in \boldP}(1)$. 
	\item Regarding term (a), by Lemma \ref{lem:fumWLLN}, \\
	$\sup_{\theta \in \Theta:\|\theta-\theta^*(\MC{P})\| > \epsilon} \left\{ \frac{1}{T} \sum_{t=1}^T \left( W_t m_{\theta,t} - \E_{\MC{P}, \pi} [ W_t m_{\theta,t} | \HH_{t-1} ] \right) \right\} = o_{\MC{P} \in \boldP}(1)$.
\end{itemize}

Thus it is sufficient to show that term (b) is such that for some $\delta' > 0$,
\begin{equation}
    \label{eqn:termblowerbound}
	\sup_{\theta \in \Theta:\|\theta-\theta^*(\MC{P})\| > \epsilon} \left\{ \frac{1}{T} \sum_{t=1}^T \E_{\MC{P}, \pi} [ W_t ( m_{\theta,t} - m_{\theta^*(\MC{P}),t}  ) | \HH_{t-1} ] \right\} \leq -\delta' \TN{ w.p. } 1.
\end{equation}
By law of iterated expectations,
\begin{equation*}
	\sup_{\theta \in \Theta :\|\theta-\theta^*(\MC{P})\| > \epsilon} \left\{ \frac{1}{T} \sum_{t=1}^T \E_{\MC{P}, \pi} [ W_t ( m_{\theta,t} - m_{\theta^*(\MC{P}),t} ) | \HH_{t-1} ] \right\}
\end{equation*}
\begin{equation*}
	= \sup_{\theta \in \Theta :\|\theta-\theta^*(\MC{P})\| > \epsilon} \left\{ \frac{1}{T} \sum_{t=1}^T \E_{\MC{P}} \left[ \int_{a \in \MC{A}} \pi_t(a, X_t, \HH_{t-1} ) \E_{\MC{P}} [ W_t ( m_{\theta,t} - m_{\theta^*(\MC{P}),t} ) | \HH_{t-1}, X_t, A_t = a ] da \bigg| \HH_{t-1} \right] \right\}.
\end{equation*}
Since $W_t \in \sigma( \HH_{t-1}, X_t, A_t )$, we have that $\E_{\MC{P}} [ W_t ( m_{\theta,t} - m_{\theta^*(\MC{P}),t} ) | \HH_{t-1}, X_t, A_t = a ] = W_t \E_{\MC{P}} [ m_{\theta,t} - m_{\theta^*(\MC{P}),t} | \HH_{t-1}, X_t, A_t = a ]$. By Condition \ref{cond:iidpo}, we have that $W_t \E_{\MC{P}} [ m_{\theta,t} - m_{\theta^*(\MC{P}),t} | \HH_{t-1}, X_t, A_t = a ] = W_t \E_{\MC{P}} [ m_{\theta,t} - m_{\theta^*(\MC{P}),t} | X_t, A_t = a ]$. Thus we have,
\begin{equation*}
	= \sup_{\theta \in \Theta :\|\theta-\theta^*(\MC{P})\| > \epsilon} \left\{ \frac{1}{T} \sum_{t=1}^T \E_{\MC{P}} \left[ \int_{a \in \MC{A}} \pi_t(a, X_t, \HH_{t-1}) W_t \E_{\MC{P}} [ m_{\theta,t} - m_{\theta^*(\MC{P}),t} | X_t, A_t = a ] da \bigg| \HH_{t-1} \right] \right\}.
\end{equation*}
Since for all $\theta \in \Theta$, $\E_{\MC{P}} [ m_{\theta,t} - m_{\theta^*(\MC{P}),t} | X_t, A_t ] \leq 0$ with probability $1$ by Condition \ref{cond:optimalsolution} and since $0 < \frac{ W_t }{ \sqrt{\rho_{\max}} } \leq 1$ with probability $1$ by Condition \ref{cond:sqrtweights},
\begin{equation*}
	\leq \sup_{\theta \in \Theta:\|\theta-\theta^*(\MC{P})\| > \epsilon} \left\{ \frac{1}{T \sqrt{\rho_{\max}} } \sum_{t=1}^T \E_{\MC{P}} \left[ \int_{a \in \MC{A}} \pi_t(a, X_t, \HH_{t-1} ) W_t^2 \E_{\MC{P}} [ m_{\theta,t} - m_{\theta^*(\MC{P}),t} | X_t, A_t = a ] da \bigg| \HH_{t-1} \right] \right\}.
\end{equation*}
Since $W_t^2 = \frac{ \pi_t^{\TN{sta}} (A_t, X_t) }{ \pi_t (A_t, X_t, \HH_{t-1}) }$,
\begin{equation*}
	= \sup_{\theta \in \Theta:\|\theta-\theta^*(\MC{P})\| > \epsilon} \left\{ \frac{1}{T \sqrt{\rho_{\max}} } \sum_{t=1}^T \E_{\MC{P}} \left[ \int_{a \in \MC{A}} \pi_t^{\TN{sta}}(a, X_t ) \E_{\MC{P}} [ m_{\theta,t} - m_{\theta^*(\MC{P}),t} | X_t, A_t = a ] da \bigg| \HH_{t-1} \right] \right\}.
\end{equation*}
By Condition \ref{cond:iidpo} and since $\pi_t^{\TN{sta}}$ is pre-specified, we can drop the conditioning on $\HH_{t-1}$, i.e.,
\begin{equation*}
	= \sup_{\theta \in \Theta:\|\theta-\theta^*(\MC{P})\| > \epsilon} \left\{ \frac{1}{T \sqrt{\rho_{\max}} } \sum_{t=1}^T \E_{\MC{P}} \left[ \int_{a \in \MC{A}} \pi_t^{\TN{sta}}(a, X_t ) \E_{\MC{P}} [ m_{\theta,t} - m_{\theta^*(\MC{P}),t} | X_t, A_t = a ] da \right] \right\}.
\end{equation*}
By law of iterated expectations,
\begin{equation*}
	= \sup_{\theta \in \Theta:\|\theta-\theta^*(\MC{P})\| > \epsilon} \left\{ \frac{1}{T \sqrt{\rho_{\max}} } \sum_{t=1}^T \E_{\MC{P}, \pi_t^{\TN{sta}}} \left[ m_{\theta,t} - m_{\theta^*(\MC{P}),t} \right] \right\}
	\leq - \frac{1}{ \sqrt{\rho_{\max}} } \delta.
\end{equation*}
The last inequality above holds for some $\delta > 0$ for all sufficiently large $T$ by Condition \ref{cond:wellseparated}. Thus Equation~\eqref{eqn:termblowerbound} holds for $\delta' = \frac{1}{ \sqrt{\rho_{\max}} } \delta$.

\subsection{Asymptotic Normality}

We prove the second part of Theorem \ref{thm:normality}, i.e., that
\begin{equation}
	\label{eqn:Mestgoalapp}
	\Sigma_T(\MC{P})^{-1/2} \ddot{M}_T( \thetahat_T ) \sqrt{T} ( \thetahat_T - \theta^*(\MC{P}) )
	\Dto \N \left( 0, I_d \right)  \TN{ uniformly over } \MC{P} \in \boldP.
\end{equation}

\subsubsection{Main Argument}

The three results we show to ensure Equation~\eqref{eqn:Mestgoalapp} holds are as follows:
\begin{equation}
	\label{result:normality}
	\Sigma_T(\MC{P})^{-1/2} \sqrt{T} \dot{M}_T( \theta^*(\MC{P}) ) \Dto \N \left( 0, I_d \right)  \TN{ uniformly over } \MC{P} \in \boldP.
\end{equation}
For $\dddot{\epsilon}_{\ddot{m}} > 0$ as defined in Condition \ref{cond:thirdderivdomination},
\begin{equation}
	\label{result:dddotM}
	\sup_{\theta \in \Theta : \| \theta - \theta^*(\MC{P}) \| \leq \epsilon_{\dddot{m}}}
	\left\| \dddot{M}_T(\theta) \right\|_1 = O_{\MC{P} \in \boldP}(1).
\end{equation}
For matrix $H$ positive definite,
\begin{equation}
	\label{result:ddotM}
	- \ddot{M}_T(\theta^*(\MC{P}))
	\succeq H + o_{\MC{P} \in \boldP}(1).
\end{equation}
For a reminder on the notation of $o_{\MC{P} \in \boldP}(1)$ and $O_{\MC{P} \in \boldP}(1)$ see definitions \ref{eqn:uniformPdef} and \ref{eqn:uniformBigO}. 
For now, we assume that  Equations~\eqref{result:normality}, \eqref{result:dddotM}, and \eqref{result:ddotM} hold; we will show they hold in Sections \ref{sec:asyptoticnormality}, \ref{sec:dddotM}, and \ref{sec:ddotM} respectively. Our argument is based on \citet[Theorem of 5.41]{van2000asymptotic}.

By differentiability Condition \ref{cond:differentiable}, since $\hat{\theta}_T$ is the maximizer of criterion $M_T( \theta )$,
\begin{equation*}
	0 = \dot{M}_T( \thetahat_T ).
\end{equation*}
By differentiability Condition \ref{cond:differentiable} again and Taylor's theorem we have that for some random $\tilde{\theta}_T$ on the line segment between $\theta^*(\MC{P})$ and $\thetahat_T$,
\begin{equation*}
	0 = \dot{M}_T( \thetahat_T )
	= \dot{M}_T( \theta^*(\MC{P}) ) + \ddot{M}_T( \theta^*(\MC{P}) ) ( \thetahat_T - \theta^*(\MC{P}) ) + \frac{1}{2} ( \thetahat_T - \theta^*(\MC{P}) )^\top \dddot{M}_T( \tilde{\theta}_T ) ( \thetahat_T - \theta^*(\MC{P}) ).
\end{equation*}

By rearranging terms and multiplying by $\sqrt{T}$,
\begin{equation*}
	- \sqrt{T} \dot{M}_T( \theta^*(\MC{P}) ) 
	= \ddot{M}_T( \theta^*(\MC{P}) ) \sqrt{T} ( \thetahat_T - \theta^*(\MC{P}) ) 
	+ \frac{1}{2} ( \thetahat_T - \theta^*(\MC{P}) )^\top \dddot{M}_T( \tilde{\theta}_T ) \sqrt{T} ( \thetahat_T - \theta^*(\MC{P}) )
\end{equation*}
\begin{equation*}
	= \left[ \ddot{M}_T( \theta^*(\MC{P}) ) + \frac{1}{2} ( \thetahat_T - \theta^*(\MC{P}) )^\top \dddot{M}_T( \tilde{\theta}_T ) \right] \sqrt{T} ( \thetahat_T - \theta^*(\MC{P}) ).
\end{equation*}

Note that by the above equation and Equation~\eqref{result:normality}, we have that
\begin{multline}
	\label{eqn:normality1}
	\Sigma_T(\MC{P})^{-1/2} \left[ \ddot{M}_T( \theta^*(\MC{P}) ) + \frac{1}{2} ( \thetahat_T - \theta^*(\MC{P}) )^\top \dddot{M}_T( \tilde{\theta}_T ) \right] \sqrt{T} ( \thetahat_T - \theta^*(\MC{P}) ) \\
	\Dto \N \left( 0, I_d \right)  \TN{ uniformly over } \MC{P} \in \boldP.
\end{multline}

By Equation~\eqref{result:ddotM}, the probability that $\ddot{M}_T(\theta^*(\MC{P}))$ is invertible goes to $1$ uniformly over $\MC{P} \in \boldP$. Thus by Equation~\eqref{eqn:normality1}, we have that
\begin{equation*}
    \Sigma_T(\MC{P})^{-1/2} \left[ I_d + \frac{1}{2} ( \thetahat_T - \theta^*(\MC{P}) )^\top \dddot{M}_T( \tilde{\theta}_T ) \ddot{M}_T( \theta^*(\MC{P}) )^{-1} \right] 
	\ddot{M}_T( \theta^*(\MC{P}) )  
	\sqrt{T} ( \thetahat_T - \theta^*(\MC{P}) )
\end{equation*}
\begin{multline}
	\label{eqn:normality2}
	= \left[ I_d + \frac{1}{2} \Sigma_T(\MC{P})^{-1/2} ( \thetahat_T - \theta^*(\MC{P}) )^\top \dddot{M}_T( \tilde{\theta}_T ) \ddot{M}_T( \theta^*(\MC{P}) )^{-1} \Sigma_T(\MC{P})^{1/2} \right] \\
	\Sigma_T(\MC{P})^{-1/2} \ddot{M}_T( \theta^*(\MC{P}) )  
	\sqrt{T} ( \thetahat_T - \theta^*(\MC{P}) ) 
	\Dto \N \left( 0, I_d \right)  \TN{ uniformly over } \MC{P} \in \boldP.
\end{multline}

We now show that $\frac{1}{2} \Sigma_T(\MC{P})^{-1/2} ( \thetahat_T - \theta^*(\MC{P}) )^\top \dddot{M}_T( \tilde{\theta}_T ) \ddot{M}_T( \theta^*(\MC{P}) )^{-1} \Sigma_T(\MC{P})^{1/2} = o_{\MC{P} \in \boldP}(1)$. 
It is sufficient to show that $\| \Sigma_T(\MC{P})^{-1/2} \| \| \thetahat_T - \theta^*(\MC{P}) \| \| \dddot{M}_T( \tilde{\theta}_T ) \|_1 \| \ddot{M}_T(\theta^*(\MC{P}))^{-1} \| \| \Sigma_T(\MC{P})^{1/2} \| = o_{\MC{P} \in \boldP}(1)$.

\begin{itemize}
    \item By Condition \ref{cond:moments}, the minimum eigenvalue of $\Sigma_T(\MC{P})$ is bounded uniformly above some constant greater than zero, so $\sup_{\MC{P} \in \boldP} \| \Sigma_T(\MC{P})^{-1/2} \| = O(1)$.
	\item By uniform consistency of $\thetahat_T$, $\| \thetahat_T - \theta^*(\MC{P}) \| = o_{\MC{P} \in \boldP}(1)$.
	\item By uniform consistency of $\thetahat_T$, $\1_{ \| \tilde{\theta}_T - \theta^*(\MC{P}) \| \leq \epsilon_{\dddot{m}} } = o_{\MC{P} \in \boldP}(1)$. Thus by Equation~\eqref{result:dddotM}, $\dddot{M}_T( \tilde{\theta}_T ) = O_{\MC{P} \in \boldP}(1)$.
	\item By Equation~\eqref{result:ddotM}, the minimum eigenvalue of $-\ddot{M}_T(\theta^*(\MC{P}))^{-1}$ is bounded above that of positive definite matrix $H$. Thus $\| \ddot{M}_T(\theta^*(\MC{P}))^{-1} \| = O_{\MC{P} \in \boldP}(1)$.
	\item By Condition \ref{cond:moments}, $\sup_{\MC{P} \in \boldP} \| \Sigma_T(\MC{P})^{1/2} \| = O(1)$.
\end{itemize}

Thus, by Slutsky's Theorem and Equation~\eqref{eqn:normality2}, we have that
\begin{equation}
	\label{eqn:normality3}
	\Sigma_T(\MC{P})^{-1/2} \ddot{M}_T( \theta^*(\MC{P}) )
	\sqrt{T} ( \thetahat_T - \theta^*(\MC{P}) ) 
	\Dto \N \left( 0, I_d \right)  \TN{ uniformly over } \MC{P} \in \boldP.
\end{equation}

Lastly, to show our desired result, that $\Sigma_T(\MC{P})^{-1/2} \ddot{M}_T( \thetahat_T ) \sqrt{T} ( \thetahat_T - \theta^*(\MC{P}) ) \Dto \N \left( 0, I_d \right)  \TN{ uniformly over } \MC{P} \in \boldP$, by Equation~\eqref{eqn:normality3} and Slutsky's Theorem it is sufficient to show that $\Sigma_T(\MC{P})^{-1/2} \ddot{M}_T( \thetahat_T ) \ddot{M}_T( \theta^*(\MC{P}) )^{-1} \Sigma_T(\MC{P})^{1/2} \Pto I_d$ uniformly over $\MC{P} \in \boldP$. 
Note if we can show that $\ddot{M}_T( \thetahat_T ) \ddot{M}_T( \theta^*(\MC{P}) )^{-1} \Pto I_d$ uniformly over $\MC{P} \in \boldP$, then $\Sigma_T(\MC{P})^{-1/2} \ddot{M}_T( \thetahat_T ) \ddot{M}_T( \theta^*(\MC{P}) )^{-1} \Sigma_T(\MC{P})^{1/2} 
= \Sigma_T(\MC{P})^{-1/2} \left[ I_d + o_{\MC{P} \in \boldP}(1) \right] \Sigma_T(\MC{P})^{1/2}
= I_d + \Sigma_T(\MC{P})^{-1/2} o_{\MC{P} \in \boldP}(1) \Sigma_T(\MC{P})^{1/2} = I_d + o_{\MC{P} \in \boldP}(1)$. The last limit holds since $\| \Sigma_T(\MC{P})^{-1/2} \| = O_{\MC{P} \in \boldP}(1)$ and $\| \Sigma_T(\MC{P})^{1/2} \| = O_{\MC{P} \in \boldP}(1)$ by Condition \ref{cond:moments} (use the same argument as that used in the bullet points below Equation \eqref{eqn:normality2}).

Thus it is sufficient to show that $\ddot{M}_T( \thetahat_T ) \ddot{M}_T( \theta^*(\MC{P}) )^{-1} \Pto I_d$ uniformly over $\MC{P} \in \boldP$.
By Taylor's Theorem, for some random $\bar{\theta}_T$ on the line segment between $\thetahat_T$ and $\theta^*(\MC{P})$,
\begin{equation*}
	\ddot{M}_T( \thetahat_T )
	= \ddot{M}_T( \theta^*(\MC{P}) ) 
	+ \dddot{M}_T( \bar{\theta}_T )  ( \thetahat_T - \theta^*(\MC{P}) ).
\end{equation*}
Recall that the probability the inverse of $\ddot{M}_T( \theta^*(\MC{P}) )$ exists goes to $1$ by Equation~\eqref{result:ddotM} (use the same argument as that used in the bullet points below Equation~\eqref{eqn:normality2}). Thus we have that $\ddot{M}_T( \thetahat_T ) \ddot{M}_T( \theta^*(\MC{P}) )^{-1}$ equals the following:
\begin{equation*}
	\left[ \ddot{M}_T( \theta^*(\MC{P}) ) 
	+ \dddot{M}_T( \bar{\theta}_T )  ( \thetahat_T - \theta^*(\MC{P}) ) \right] \ddot{M}_T( \theta^*(\MC{P}) )^{-1}
\end{equation*}
\begin{equation*}
	= I_d + \dddot{M}_T( \bar{\theta}_T )  ( \thetahat_T - \theta^*(\MC{P}) ) \ddot{M}_T( \theta^*(\MC{P}) )^{-1}
\end{equation*}
Note that $\dddot{M}_T( \bar{\theta}_T )  ( \thetahat_T - \theta^*(\MC{P}) ) \ddot{M}_T( \theta^*(\MC{P}) )^{-1} = o_{\MC{P} \in \boldP}(1)$ because 
\begin{itemize}
    \item By uniform consistency of $\thetahat_T$, $\1_{ \| \tilde{\theta}_T - \theta^*(\MC{P}) \| \leq \epsilon_{\dddot{m}} } = o_{\MC{P} \in \boldP}(1)$. Thus by Equation~\eqref{result:dddotM}, $\dddot{M}_T( \tilde{\theta}_T ) = O_{\MC{P} \in \boldP}(1)$.
	\item By uniform consistency of $\thetahat_T$, $\| \thetahat_T - \theta^*(\MC{P}) \| = o_{\MC{P} \in \boldP}(1)$.
	\item By Equation~\eqref{result:ddotM}, $\| \ddot{M}_T(\theta^*(\MC{P}))^{-1} \| = O_{\MC{P} \in \boldP}(1)$.
\end{itemize}

\subsubsection{Asymptotic Normality of $\Sigma_T(\MC{P})^{-1/2} \sqrt{T} \dot{M}_T( \theta^*(\MC{P}) )$}
\label{sec:asyptoticnormality}

We will show that Equation~\eqref{result:normality} holds by applying a martingale central limit theorem. For notational convenience, we let $\dot{m}_{\theta, t} := \dot{m}_\theta( Y_t, X_t, A_t )$. Note that  by definition $\Sigma_T(\MC{P})^{-1/2} \sqrt{T} \dot{M}_T( \theta^*(\MC{P}) ) = \Sigma_T(\MC{P})^{-1/2} \frac{1}{ \sqrt{T} } \sum_{t=1}^T W_t \dot{m}_{\theta^*(\MC{P}), t}$. We first show that $\left\{ \Sigma_T(\MC{P})^{-1/2} \frac{1}{ \sqrt{T} } W_t \dot{m}_{\theta^*(\MC{P}), t} \right\}_{t=1}^T$ is a martingale difference sequence with respect to $\{ \HH_t \}_{t=0}^T$. For any $t \in [1 \colon T]$,
\begin{equation*}
	\E_{\MC{P}, \pi} \left[ \frac{1}{ \sqrt{T} } \Sigma_T(\MC{P})^{-1/2} W_t \boldc^\top \dot{m}_{\theta^*(\MC{P}), t} \bigg| \HH_{t-1} \right]
\end{equation*}
\begin{equation*}
	\underset{(a)}{=} \frac{1}{ \sqrt{T} } \E_{\MC{P}, \pi} \left[ \E_{\MC{P}} \left[ \Sigma_T(\MC{P})^{-1/2} W_t \boldc^\top \dot{m}_{\theta^*(\MC{P}), t} \big| \HH_{t-1}, X_t, A_t \right] \bigg| \HH_{t-1} \right]
\end{equation*}
\begin{equation*}
	\underset{(b)}{=} \frac{1}{ \sqrt{T} } \Sigma_T(\MC{P})^{-1/2} \E_{\MC{P}, \pi} \left[ W_t \boldc^\top \E_{\MC{P}} \left[ \dot{m}_{\theta^*(\MC{P}), t} \big| \HH_{t-1}, X_t, A_t \right] \bigg| \HH_{t-1} \right] \underset{(c)}{=} 0
\end{equation*}

\begin{itemize}
	\item Above, (a) holds by law of iterated expectations.
	\item (b) holds since $W_t \in \sigma(\HH_{t-1}, X_t, A_t)$ and since $\Sigma_T(\MC{P})$ are a function of stabilizing policies $\{ \pi_t^{\TN{sta}} \}_{t \geq 1}$, which are pre-specified.
	\item By Condition \ref{cond:iidpo}, $\E_{\MC{P}} \left[ \dot{m}_{\theta^*(\MC{P}), t} \big| \HH_{t-1}, X_t, A_t \right] = \E_{\MC{P}} \left[ \dot{m}_{\theta^*(\MC{P}), t} \big| X_t, A_t \right]$. Equality (c) holds because $\E_{\MC{P}} \left[ \dot{m}_{\theta^*(\MC{P}), t} \big| X_t, A_t \right] = 0$ with probability $1$ by Condition \ref{cond:optimalsolution}; note that $\theta^*(\MC{P})$ is a critical point of $\E_{\MC{P}}[ m_{\theta, t} | X_t, A_t ]$.
\end{itemize}

By Cramer-Wold device, to show that Equation~\eqref{result:normality} holds, it is sufficient to show that for any fixed $\boldc \in \real^d$ with $\| \boldc \|_2 = 1$, that $\boldc^\top \Sigma_T(\MC{P})^{-1/2} \frac{1}{ \sqrt{T} } \sum_{t=1}^T W_t \dot{m}_{\theta^*(\MC{P}), t} 
	\Dto \N \left( 0, \boldc^\top I_d \boldc \right)  \TN{ uniformly over } \MC{P} \in \boldP$.
We now apply Theorem \ref{thm:umCLT}, a uniform version of the martingale central limit theorem of \citet{dvoretzky1972asymptotic}; while the original theorem holds for any fixed $\MC{P}$, we can show uniform convergence in distribution by ensuring that the conditions of the theorem hold uniformly over $\MC{P} \in \boldP$ (see Definition \ref{def:unifdist}). By Theorem \ref{thm:umCLT}, it is sufficient to show that the following two conditions hold:

1.  \bo{Conditional Variance:} $\frac{1}{T} \sum_{t=1}^T \E_{\MC{P}, \pi} \left[ \left\{ \boldc^\top \Sigma_T(\MC{P})^{-1/2} W_t \dot{m}_{\theta^*(\MC{P}), t} \right\}^2 \bigg| \HH_{t-1} \right] \Pto \sigma^2$ uniformly over $\MC{P} \in \boldP$.

2. \bo{Conditional Lindeberg:} For any $\delta > 0$, \\
	$\frac{1}{T} \sum_{t=1}^T \E_{\MC{P}, \pi} \left[ \left\{ \boldc^\top \Sigma_T(\MC{P})^{-1/2} W_t \dot{m}_{\theta^*(\MC{P}), t} \right\}^2 \1_{| \boldc^\top \Sigma_T(\MC{P})^{-1/2} W_t \dot{m}_{\theta^*(\MC{P}), t}  | > \delta \sqrt{T} } \bigg| \HH_{t-1} \right] \Pto 0$ uniformly over $\MC{P} \in \boldP$.

\paragraph{1. Conditional Variance}
\begin{equation*}
	\frac{1}{T} \sum_{t=1}^T \E_{\MC{P}, \pi} \left[ \left( \boldc^\top W_t \Sigma_T(\MC{P})^{-1/2} \dot{m}_{\theta^*(\MC{P}), t} \right)^2 \bigg| \HH_{t-1} \right]
\end{equation*}
\begin{equation*}
	= \frac{1}{T} \sum_{t=1}^T \E_{\MC{P}, \pi} \left[ W_t^2 \boldc^\top \Sigma_T(\MC{P})^{-1/2} \dot{m}_{\theta^*(\MC{P}), t}^{\otimes 2} \Sigma_T(\MC{P})^{-1/2} \boldc \bigg| \HH_{t-1} \right]
\end{equation*}
\begin{equation*}
	\underset{(a)}{=} \boldc^\top \Sigma_T(\MC{P})^{-1/2} \left\{ \frac{1}{T}  \sum_{t=1}^T \E_{\MC{P}, \pi} \left[ W_t^2 \dot{m}_{\theta^*(\MC{P}), t}^{\otimes 2} \bigg| \HH_{t-1} \right] \right\} \Sigma_T(\MC{P})^{-1/2} \boldc
\end{equation*}
\begin{equation*}
	\underset{(b)}{=} \boldc^\top \Sigma_T(\MC{P})^{-1/2} \left\{ \frac{1}{T} \sum_{t=1}^T \E_{\MC{P}} \left[ \int_{a \in \MC{A}} \pi_t(a, X_t, \HH_{t-1}) \E_{\MC{P}} \left[ W_t^2 \dot{m}_{\theta^*(\MC{P}), t}^{\otimes 2} \big| \HH_{t-1}, X_t, A_t = a \right] da \bigg| \HH_{t-1} \right] \right\} \Sigma_T(\MC{P})^{-1/2} \boldc
\end{equation*}
\begin{equation*}
	\underset{(c)}{=} \boldc^\top \Sigma_T(\MC{P})^{-1/2} \left\{ \frac{1}{T} \sum_{t=1}^T \E_{\MC{P}} \left[ \int_{a \in \MC{A}} \pi_t^{\TN{sta}}(a, X_t) \E_{\MC{P}} \left[ \dot{m}_{\theta^*(\MC{P}), t}^{\otimes 2} \big| \HH_{t-1}, X_t, A_t = a \right] da \bigg| \HH_{t-1} \right] \right\} \Sigma_T(\MC{P})^{-1/2} \boldc
\end{equation*}
\begin{equation*}
	\underset{(d)}{=} \boldc^\top \Sigma_T(\MC{P})^{-1/2} \left\{ \frac{1}{T} \sum_{t=1}^T \E_{\MC{P}} \left[ \E_{\MC{P}, \pi_t^{\TN{sta}}} \left[ \dot{m}_{\theta^*(\MC{P}), t}^{\otimes 2} \big| X_t \right] \bigg| \HH_{t-1} \right] \right\} \Sigma_T(\MC{P})^{-1/2} \boldc
\end{equation*}
\begin{equation*}
	\underset{(e)}{=} \boldc^\top \Sigma_T(\MC{P})^{-1/2} \left\{ \frac{1}{T} \sum_{t=1}^T \E_{\MC{P}, \pi_t^{\TN{sta}}} \left[ \dot{m}_{\theta^*(\MC{P}), t}^{\otimes 2} \right] \right\} \Sigma_T(\MC{P})^{-1/2} \boldc
\end{equation*}
\begin{equation*}
	\underset{(f)}{=} \boldc^\top \Sigma_T(\MC{P})^{-1/2} \Sigma_T(P) \Sigma_T(\MC{P})^{-1/2} \boldc
	= \boldc^\top I_d \boldc
\end{equation*}

\begin{itemize}
	\item Above, (a) holds since $\Sigma_T(\MC{P})$ are a function of stabilizing policies $\{ \pi_t^{\TN{sta}} \}_{t \geq 1}$, which are pre-specified.
	\item Equality (b) holds by law of iterated expectations.
	\item Equality (c) holds since $W_t = \sqrt{ \frac{ \pi_t^{\TN{sta}} (A_t, X_t) }{ \pi_t (A_t, X_t, \HH_{t-1}) } } \in \sigma( \HH_{t-1}, X_t, A_t )$.
	\item Equality (d) holds because by Condition \ref{cond:iidpo}, $\E_{\MC{P}} [ \dot{m}_{\theta^*(\MC{P}), t}^{\otimes 2} | \HH_{t-1}, X_t, A_t = a ] 
	= \E_{\MC{P}} [ \dot{m}_{\theta^*(\MC{P}), t}^{\otimes 2} | X_t, A_t = a ]$ and by law of iterated expectations.
	\item Equality (e) holds because by Condition \ref{cond:iidpo}, the distribution of $X_t$ does not depend on $\HH_{t-1}$, so $\E_{\MC{P}} \left[ \E_{\MC{P}, \pi_t^{\TN{sta}}} \left[ \dot{m}_{\theta^*(\MC{P}), t}^{\otimes 2} \big| X_t \right] \bigg| \HH_{t-1} \right] 
	= \E_{\MC{P}} \left[ \E_{\MC{P}, \pi_t^{\TN{sta}}} \left[ \dot{m}_{\theta^*(\MC{P}), t}^{\otimes 2} \big| X_t \right] \right] 
	= \E_{\MC{P}, \pi_t^{\TN{sta}}} \left[ \dot{m}_{\theta^*(\MC{P}), t}^{\otimes 2} \right]$; the last equality holds by law of iterated expectations.
	\item Equality (f) holds by definition.
\end{itemize}

\paragraph{2. Conditional Lindeberg}
\begin{equation*}
	\frac{1}{T} \sum_{t=1}^T \E_{\MC{P}, \pi} \left[ \left( \boldc^\top W_t \Sigma_T(\MC{P})^{-1/2} \dot{m}_{\theta^*(\MC{P}), t} \right)^2 \1_{ \left| \boldc^\top W_t \Sigma_T(\MC{P})^{-1/2} \dot{m}_{\theta^*(\MC{P}), t} \right| > \delta \sqrt{T} } \bigg| \HH_{t-1} \right]
\end{equation*}
\begin{equation*}
	= \frac{1}{T} \sum_{t=1}^T \E_{\MC{P}, \pi} \left[ W_t^2 \boldc^\top \Sigma_T(\MC{P})^{-1/2} \dot{m}_{\theta^*(\MC{P}), t}^{\otimes 2} \Sigma_T(\MC{P})^{-1/2} \boldc 
	\1_{ \left| \boldc^\top W_t \Sigma_T(\MC{P})^{-1/2} \dot{m}_{\theta^*(\MC{P}), t} \right| > \delta \sqrt{T} } \bigg| \HH_{t-1} \right]
\end{equation*}
\begin{equation*}
	\underset{(a)}{\leq} \frac{1}{T^2 \delta^2} \sum_{t=1}^T \E_{\MC{P}, \pi} \left[ W_t^4 \left( \boldc^\top \Sigma_T(\MC{P})^{-1/2} \dot{m}_{\theta^*(\MC{P}), t}^{\otimes 2} \Sigma_T(\MC{P})^{-1/2} \boldc \right)^2 \bigg| \HH_{t-1} \right]
\end{equation*}
\begin{equation*}
	\underset{(b)}{\leq} \frac{\rho_{\max}}{T^2 \delta^2} \sum_{t=1}^T \E_{\MC{P}, \pi} \left[ W_t^2 \left( \boldc^\top \Sigma_T(\MC{P})^{-1/2} \dot{m}_{\theta^*(\MC{P}), t}^{\otimes 2} \Sigma_T(\MC{P})^{-1/2} \boldc \right)^2 \bigg| \HH_{t-1} \right]
\end{equation*}
\begin{equation*}
	\underset{(c)}{=} \frac{\rho_{\max}}{T^2 \delta^2} \sum_{t=1}^T \E_{\MC{P}} \left[ \int_{a \in \MC{A}} \pi_t(a, X_t, \HH_{t-1}) \E_{\MC{P}} \left[ W_t^2 \left( \boldc^\top \Sigma_T(\MC{P})^{-1/2} \dot{m}_{\theta^*(\MC{P}), t}^{\otimes 2} \Sigma_T(\MC{P})^{-1/2} \boldc \right)^2 \bigg| \HH_{t-1}, X_t, A_t = a \right] da \bigg| \HH_{t-1} \right]
\end{equation*}
\begin{equation*}
	\underset{(d)}{=} \frac{\rho_{\max}}{T^2 \delta^2} \sum_{t=1}^T \E_{\MC{P}} \left[ \int_{a \in \MC{A}} \pi_t^{\TN{sta}}(a, X_t) \E_{\MC{P}} \left[ \left( \boldc^\top \Sigma_T(\MC{P})^{-1/2} \dot{m}_{\theta^*(\MC{P}), t}^{\otimes 2} \Sigma_T(\MC{P})^{-1/2} \boldc \right)^2 \bigg| \HH_{t-1}, X_t, A_t = a \right] da \bigg| \HH_{t-1} \right]
\end{equation*}
\begin{equation*}
	\underset{(e)}{=} \frac{\rho_{\max}}{T^2 \delta^2} \sum_{t=1}^T \E_{\MC{P}} \left[  \E_{\MC{P}} \left[ \left( \boldc^\top \Sigma_T(\MC{P})^{-1/2} \dot{m}_{\theta^*(\MC{P}), t}^{\otimes 2} \Sigma_T(\MC{P})^{-1/2} \boldc \right)^2 \bigg| X_t \right] \bigg| \HH_{t-1} \right]
\end{equation*}
\begin{equation*}
	\underset{(f)}{=} \frac{\rho_{\max}}{T^2 \delta^2} \sum_{t=1}^T \E_{\MC{P}, \pi_t^{\TN{sta}}} \left[ \left( \boldc^\top \Sigma_T(\MC{P})^{-1/2} \dot{m}_{\theta^*(\MC{P}), t}^{\otimes 2} \Sigma_T(\MC{P})^{-1/2} \boldc \right)^2 \right]
	\underset{(g)}{\to} 0
\end{equation*}

\begin{itemize}
	\item Above, inequality (a) holds because $\1_{ \left| W_t \boldc^\top \Sigma_T(\MC{P})^{-1/2} \dot{m}_{\theta^*(\MC{P}), t} \right| > \sqrt{T} \delta}=1$
		if and only if \\
		$W_t^2 \frac{1}{T \delta^2} \boldc^\top \Sigma_T(\MC{P})^{-1/2} \dot{m}_{\theta^*(\MC{P}), t}^{\otimes 2} \Sigma_T(\MC{P})^{-1/2} \boldc > 1$.
	\item Inequality (b) holds because by Condition \ref{cond:sqrtweights}, $W_t^2 \leq \rho_{\max}$ with probability $1$.
	\item Equality (c) holds by the law of iterated expectations.	
	\item Equality (d) holds since $W_t = \sqrt{ \frac{ \pi_t^{\TN{sta}} (A_t, X_t) }{ \pi_t (A_t, X_t, \HH_{t-1}) } } \in \sigma( \HH_{t-1}, X_t, A_t )$.
	\item Equality (e) holds because by Condition \ref{cond:iidpo}, \\
	$\E_{\MC{P}} \left[ ( \boldc^\top \Sigma_T(\MC{P})^{-1/2} \dot{m}_{\theta^*(\MC{P}), t}^{\otimes 2} \Sigma_T(\MC{P})^{-1/2} \boldc )^2 \big| \HH_{t-1}, X_t, A_t=a \right]
	= \E_{\MC{P}} \left[ ( \boldc^\top \Sigma_T(\MC{P})^{-1/2} \dot{m}_{\theta^*(\MC{P}), t}^{\otimes 2} \Sigma_T(\MC{P})^{-1/2} \boldc )^2 \big| X_t \right]$ and by law of iterated expectations.
	\item Equality (f) holds since the distribution of $X_t$ does not depend on $\HH_{t-1}$ by Condition \ref{cond:iidpo} and by law of iterated expectations.
	\item Regarding limit (g), it is sufficient to show that $\frac{1}{T} \sum_{t=1}^T \E_{\MC{P}, \pi_t^{\TN{sta}}} \left[ \left( \boldc^\top \Sigma_T(\MC{P})^{-1/2} \dot{m}_{\theta^*(\MC{P}), t}^{\otimes 2} \Sigma_T(\MC{P})^{-1/2} \boldc \right)^2 \right]$ is uniformly bounded over $\MC{P} \in \boldP$ for all sufficiently large $T$.
	By Condition \ref{cond:moments}, the minimum eigenvalue of $\Sigma_T(P)$ is bounded above zero uniformly over $\MC{P} \in \boldP$ for all sufficiently large $T$; this bounds the maximum eigenvalue of $\Sigma_T(P)^{-1}$. Also by Condition \ref{cond:moments} the fourth moment of $\dot{m}_{\theta^*(\MC{P}), t}$ with respect to $\MC{P}$ and policy $\pi_t^{\TN{sta}}$ is uniformly bounded over $\MC{P} \in \boldP$ and $t \geq 1$. With these two properties we have that $\frac{1}{T} \sum_{t=1}^T \E_{\MC{P}, \pi_t^{\TN{sta}}} \left[ \left( \boldc^\top \Sigma_T(\MC{P})^{-1/2} \dot{m}_{\theta^*(\MC{P}), t}^{\otimes 2} \Sigma_T(\MC{P})^{-1/2} \boldc \right)^2 \right]$ is uniformly bounded over $\MC{P} \in \boldP$ for all sufficiently large $T$.
\end{itemize}

\subsubsection{Showing that $\sup_{\theta \in \Theta : \| \theta - \theta^*(\MC{P}) \| \leq \epsilon_{\dddot{m}}} \left\| \dddot{M}_T(\theta) \right\|_1$ is bounded in probability}
\label{sec:dddotM}

Recall that for any $B \in \real^{d \by d \by d}$, we denote $\| B \|_1 = \sum_{i=1}^d \sum_{j=1}^d \sum_{k=1}^d |B_{i,j,k}|$. We abbreviate $\dddot{m}_{\theta} (Y_t, X_t, A_t)$ with $\dddot{m}_{\theta, t}$.

By triangle inequality, $\left\| \dddot{M}_T( \theta ) \right\|_1 = \left\| \frac{1}{T} \sum_{t=1}^T W_t \dddot{m}_{\theta, t} \right\|_1 \leq \frac{1}{T} \sum_{t=1}^T W_t \left\| \dddot{m}_{\theta, t} \right\|_1$. Thus we have that
\begin{equation*}
	\sup_{\theta \in \Theta : \| \theta - \theta^*(\MC{P}) \| \leq \epsilon_{\dddot{m}} }  \left\| \dddot{M}_T( \theta ) \right\|_1
	\leq \sup_{\theta \in \Theta : \| \theta - \theta^*(\MC{P}) \| \leq \epsilon_{\dddot{m}} } 
	\frac{1}{T} \sum_{t=1}^T W_t \left\| \dddot{m}_{\theta, t} \right\|_1.
\end{equation*}
By Condition \ref{cond:thirdderivdomination} (ii), there exists a function $\dddot{m}$ (note it is not indexed by $\theta$) such that for all $\MC{P} \in \boldP$, we have that $\sup_{ \theta \in \Theta : \| \theta - \theta^*(\MC{P}) \| \leq \epsilon_{\dddot{m}}}\left\| \dddot{m}_{\theta, t} \right\|_1 \leq \left\| \dddot{m} (Y_t, X_t, A_t) \right\|_1$.
\begin{equation*}
	\leq \frac{1}{T} \sum_{t=1}^T W_t \left\| \dddot{m} (Y_t, X_t, A_t) \right\|_1.
\end{equation*}
Adding and subtracting $\frac{1}{T} \sum_{t=1}^T \E_{\MC{P}, \pi} \left[ W_t \left\| \dddot{m} (Y_t, X_t, A_t) \right\|_1 | \HH_{t-1} \right]$,
\begin{equation*}
	= \frac{1}{T} \sum_{t=1}^T W_t \left\| \dddot{m} (Y_t, X_t, A_t) \right\|_1
	- \E_{\MC{P}, \pi} \left[ W_t \left\| \dddot{m} (Y_t, X_t, A_t) \right\|_1 | \HH_{t-1} \right]
	+ \E_{\MC{P}, \pi} \left[ W_t \left\| \dddot{m} (Y_t, X_t, A_t) \right\|_1 | \HH_{t-1} \right].
\end{equation*}
By second moment bounds on $\left\| \dddot{m} (Y_t, X_t, A_t) \right\|_1$ from Condition \ref{cond:thirdderivdomination} (i), by Lemma \ref{lem:wllnW}, we have that $\frac{1}{T} \sum_{t=1}^T W_t \left\| \dddot{m} (Y_t, X_t, A_t) \right\|_1 - \E_{\MC{P}, \pi} \left[ W_t \left\| \dddot{m} (Y_t, X_t, A_t) \right\|_1 | \HH_{t-1} \right] = o_{\MC{P} \in \boldP}(1)$.
\begin{equation*}
	= o_{\MC{P} \in \boldP}(1) + \frac{1}{T} \sum_{t=1}^T \E_{\MC{P}, \pi} \left[ W_t \left\| \dddot{m} (Y_t, X_t, A_t) \right\|_1 | \HH_{t-1} \right]
\end{equation*}

Since by Condition \ref{cond:sqrtweights}, $\frac{W_t}{ \sqrt{\rho_{\min}} } \geq 1$ with probability $1$, 
\begin{equation*}
	\leq o_{\MC{P} \in \boldP}(1) + \frac{1}{ T \sqrt{\rho_{\min}} } \sum_{t=1}^T \E_{\MC{P}, \pi} \left[ W_t^2 \left\| \dddot{m} (Y_t, X_t, A_t) \right\|_1 | \HH_{t-1} \right]
\end{equation*}
Since $W_t^2 = \frac{ \pi_t^{\TN{sta}} (A_t, X_t) }{ \pi_t(A_t, X_t, \HH_{t-1}) }$ and by Condition \ref{cond:iidpo},
\begin{equation*}
	= o_{\MC{P} \in \boldP}(1) + \frac{1}{ T \sqrt{\rho_{\min}} } \sum_{t=1}^T \E_{\MC{P}, \pi_t^{\TN{sta}}} \left[ \left\| \dddot{m} (Y_t, X_t, A_t) \right\|_1\right] = O_{\MC{P} \in \boldP}(1).
\end{equation*}
Note that by Jensen's inequality, $\E_{\MC{P}, \pi_t^{\TN{sta}}} \left[ \left\| \dddot{m} (Y_t, X_t, A_t) \right\|_1 \right] \leq \sqrt{ \E_{\MC{P}, \pi_t^{\TN{sta}}} \left[ \left\| \dddot{m} (Y_t, X_t, A_t) \right\|_1^2 \right] }$. By Condition \ref{cond:thirdderivdomination} (i), $\sup_{\MC{P} \in \boldP, t \geq 1} \E_{\MC{P}, \pi_t^{\TN{sta}}} \left[ \left\| \dddot{m} (Y_t, X_t, A_t) \right\|_1^2 \right]$ is bounded, which implies the final limit above.

\subsubsection{Lower bounding $- \ddot{M}_T(\theta^*(\MC{P}))$}
\label{sec:ddotM}

We now show that $- \ddot{M}_T(\theta^*(\MC{P})) \succeq H + o_{\MC{P} \in \boldP}(1)$, for positive definite matrix $H$ introduced in Condition \ref{cond:optimalsolution} (ii).

By Condition \ref{cond:moments} and Lemma \ref{lem:wllnW}, $\frac{1}{T} \sum_{t=1}^T W_t \ddot{m}_{\theta^*(\MC{P}), t} - \E_{\MC{P}, \pi} \left[ W_t \ddot{m}_{\theta^*(\MC{P}), t} | \HH_{t-1} \right] = o_{\MC{P} \in \boldP}(1)$, so
\begin{equation*}
	-\ddot{M}_T(\theta^*(\MC{P})) = -\frac{1}{T} \sum_{t=1}^T W_t \ddot{m}_{\theta^*(\MC{P}), t} = o_{\MC{P} \in \boldP}(1) - \frac{1}{T} \sum_{t=1}^T \E_{\MC{P}, \pi} \left[ W_t \ddot{m}_{\theta^*(\MC{P}), t} | \HH_{t-1} \right]
\end{equation*}
By law of iterated expectations,
\begin{equation*}
	= o_{\MC{P} \in \boldP}(1) - \frac{1}{T} \sum_{t=1}^T \E_{\MC{P}, \pi} \left[ W_t \E_{\MC{P}} \left[ \ddot{m}_{\theta^*(\MC{P}), t}  | \HH_{t-1}, X_t, A_t \right] | \HH_{t-1} \right]
\end{equation*}
By Condition \ref{cond:iidpo},
\begin{equation*}
	= o_{\MC{P} \in \boldP}(1) - \frac{1}{T} \sum_{t=1}^T \E_{\MC{P}, \pi} \left[ W_t \E_{\MC{P}} \left[ \ddot{m}_{\theta^*(\MC{P}), t}  | X_t, A_t \right] | \HH_{t-1} \right]
\end{equation*}

By Condition \ref{cond:optimalsolution}, we have that $\E_{\MC{P}} \left[ \ddot{m}_{\theta^*(\MC{P}), t}  | X_t, A_t \right] \preceq 0$; recall that $\theta^*(\MC{P})$ is a maximizing value of $\E_{\MC{P}, \pi} \left[ m_{\theta, t}  | X_t, A_t \right]$. 
Also since $\frac{W_t}{ \sqrt{\rho_{\max}} } \leq 1$ with probability $1$ by Condition \ref{cond:sqrtweights},
\begin{equation*}
	\succeq o_{\MC{P} \in \boldP}(1) - \frac{1}{T \sqrt{\rho_{\max}} } \sum_{t=1}^T \E_{\MC{P}, \pi} \left[ W_t^2 \E_{\MC{P}, \pi} \left[ \ddot{m}_{\theta^*(\MC{P}), t}  | X_t, A_t \right] | \HH_{t-1} \right]
\end{equation*}
Since $W_t^2 = \frac{ \pi_t^{\TN{sta}} (A_t, X_t) }{ \pi_t(A_t, X_t, \HH_{t-1}) }$,
\begin{equation*}
	= o_{\MC{P} \in \boldP}(1) - \frac{1}{T \sqrt{\rho_{\max}} } \sum_{t=1}^T \E_{\MC{P}, \pi_t^{\TN{sta}}} \left[ \ddot{m}_{\theta^*(\MC{P}), t} | \HH_{t-1} \right]
\end{equation*}
Note that for any $t \geq 1$, $\E_{\MC{P}, \pi_t^{\TN{sta}}} \left[ \ddot{m}_{\theta^*(\MC{P}), t} | \HH_{t-1} \right] = \E_{\MC{P}, \pi_t^{\TN{sta}}} \left[ \ddot{m}_{\theta^*(\MC{P}), t} \right]$ because $\{ \pi_t^{\TN{sta}} \}_{t \geq 1}$ are pre-specified. Recall that by Condition \ref{cond:optimalsolution} for all sufficiently large $T$, $-\frac{1}{T} \sum_{t=1}^T \E_{\MC{P}, \pi_t^{\TN{sta}}} \left[ \ddot{m}_{\theta^*(\MC{P}), t} \right] \succeq H$ for all $\MC{P} \in \boldP$. Thus our final result is that
\begin{equation}
	\label{eqn:boundingddotm}
	- \ddot{M}_T(\theta^*(\MC{P}))
	\succeq H + o_{\MC{P} \in \boldP}(1).
\end{equation}


\subsection{Lemmas and Other Helpful Results}
\label{app:normalityLemmas}

\begin{theorem}[Uniform Martingale Central Limit Theorem]
	\label{thm:umCLT}
	Let $\{ Z_T(\MC{P}) \}_{T \geq 1}$ be a sequence of random variables whose distributions are defined by some $\MC{P} \in \boldP$ and some nuisance component $\eta$. Moreover, let $\{ Z_T(\MC{P}) \}_{T \geq 1}$ be a martingale difference sequence with respect to $\F_t$, meaning $\E_{\MC{P}, \eta}[ Z_t(\MC{P}) | \F_{t-1} ] = 0$ for all $t \geq 1$ and $\MC{P} \in \boldP$. 
	\begin{enumerate}[label=(\alph*)]
		\item $\frac{1}{T} \sum_{t=1}^T \E_{\MC{P}, \eta}[ Z_t(\MC{P})^2 | \F_{t-1} ] \Pto \sigma^2$ uniformly over $\MC{P} \in \boldP$, where $\sigma^2$ is a constant $0 < \sigma^2 < \infty$.
		\item For any $\epsilon > 0$, $\frac{1}{T} \sum_{t=1}^T \E_{\MC{P}, \eta}[ Z_t(\MC{P})^2 \1_{|Z_t(\MC{P})| > \epsilon} | \F_{t-1} ] \Pto 0$ uniformly over $\MC{P} \in \boldP$.
	\end{enumerate}
	Under the above conditions,
	\begin{equation*}
		\frac{1}{\sqrt{T}} \sum_{t=1}^T Z_t(\MC{P}) \Dto \N(0, \sigma^2) \TN{ uniformly over } \MC{P} \in \boldP.
	\end{equation*}
\end{theorem}

\paragraph{Proof:}
By by \citet[Lemma 1]{kasy2019uniformity}, it is sufficient to show that for any sequence $\{ \MC{P}_T \}_{T=1}^\infty$ with $\MC{P}_T \in \boldP$ for all $T \geq 1$, $\frac{1}{\sqrt{T}} \sum_{t=1}^T Z_t(\MC{P}_T) \Dto \N(0, \sigma^2)$. In this setting, since $\MC{P}_T$ depends on $T$, we consider triangular array asymptotics and additionally index by $T$, e.g., $\F_{T,t}$.

Note that $\frac{1}{T} \sum_{t=1}^T \E_{\MC{P}_T, \eta} [ Z_t(\MC{P}_T)^2 | \F_{T,t-1} ] \Pto \sigma^2$, by \citet[Lemma 1]{kasy2019uniformity} and condition (a) above.

Also, for any $\epsilon > 0$, $\frac{1}{T} \sum_{t=1}^T \E_{\MC{P}_T, \eta} \left[ Z_t(\MC{P}_T)^2 \1_{| Z_t(\MC{P}_T) | > \epsilon} \big| \F_{T,t-1} \right] \Pto 0$, by \citet[Lemma 1]{kasy2019uniformity} and condition (b) above.

Thus by the martingale central limit theorem of \citet{dvoretzky1972asymptotic}, we have that for the sequence $\{ \MC{P}_T \}_{T=1}^\infty$,
\begin{equation*}
	\frac{1}{\sqrt{T}} \sum_{t=1}^T Z_t(\MC{P}_T) \Dto \N(0, 1).
\end{equation*}
Since the sequence $\{ \MC{P}_T \}_{T=1}^\infty$ were chosen arbitrarily from $\boldP$, the desired result is implied again by \citet[Lemma 1]{kasy2019uniformity}.


\begin{lemma}
	\label{lem:wllnW}
	Let $f(Y_t, X_t, A_t) \in \real^{d_f}$ be a function such that \\
	$\sup_{\MC{P} \in \boldP, t \geq 1} \E_{\MC{P}, \pi_t^{\TN{sta}}} \left[ \left\| f( Y_t, X_t, A_t ) \right\|^2 \right] < m$ for some $m < \infty$. Under Conditions \ref{cond:iidpo} and \ref{cond:sqrtweights},
	\begin{equation}
	    \label{eqn:wllnW}
	    \frac{1}{\sqrt{T}} \sum_{t=1}^T \bigg\{ W_t f(Y_t, X_t, A_t) - \E_{\MC{P}, \pi}[ W_t f(Y_t, X_t, A_t) | \HH_{t-1} ] \bigg\} = O_{\MC{P} \in \boldP}(1).
    \end{equation}
    Note that the above equation implies that
    \begin{equation*}
	    \frac{1}{T} \sum_{t=1}^T \bigg\{ W_t f(Y_t, X_t, A_t) - \E_{\MC{P}, \pi}[ W_t f(Y_t, X_t, A_t) | \HH_{t-1} ] \bigg\} = o_{\MC{P} \in \boldP}(1).
    \end{equation*}
\end{lemma}

Lemma \ref{lem:wllnW} is a type of martingale weak law of large number result and the proof is similar to the weak law of large numbers proofs for i.i.d. random variables.

\paragraph{Proof:}
We denote the $k^{ \TN{th} } \in [1 \colon d_f]$ dimension of vector $f(Y_t, X_t, A_t)$ as $f^k(Y_t, X_t, A_t)$. It is sufficient to show the result for any dimension of vector $f(Y_t, X_t, A_t)$. For notational convenience, let $f_t := f^k(Y_t, X_t, A_t)$. Let $\epsilon > 0$.

\begin{equation*}
	\sup_{\MC{P} \in \boldP} \PP_{\MC{P}, \pi} \left( \left| \frac{1}{\sqrt{T}} \sum_{t=1}^T \bigg\{ W_t f_t - \E_{\MC{P}, \pi} [ W_t f_t | \HH_{t-1} ] \bigg\} \right| > \epsilon \right)
\end{equation*}
\begin{equation*}
	\underset{(a)}{\leq} \frac{1}{ T \epsilon^2 } \sup_{\MC{P} \in \boldP}  \E_{\MC{P}, \pi} \left[ \left( \sum_{t=1}^T \bigg\{ W_t f_t - \E_{\MC{P}, \pi} [ W_t f_t | \HH_{t-1} ] \bigg\} \right)^2 \right]
\end{equation*}
\begin{equation*}
	\underset{(b)}{=} \frac{1}{T \epsilon^2 } \sup_{\MC{P} \in \boldP}  \sum_{t=1}^T \E_{\MC{P}, \pi} \left[ \bigg\{ W_t f_t - \E_{\MC{P}, \pi} [ W_t f_t | \HH_{t-1} ] \bigg\}^2 \right]
\end{equation*}
\begin{equation*}
	\underset{(c)}{\leq} \frac{1}{T \epsilon^2} \sup_{\MC{P} \in \boldP} \sum_{t=1}^T \E_{\MC{P}, \pi} \left[ W_t^2 f_t^2 \right] 
\end{equation*}
\begin{equation*}
		\underset{(d)}{=} \frac{1}{T \epsilon^2} \sup_{\MC{P} \in \boldP} \sum_{t=1}^T \E_{\MC{P}} \left[ \int_{a \in \MC{A}} W_t^2 \pi_t(a, X_t, \HH_{t-1}) \E_{\MC{P}} [ f_t^2 | \HH_{t-1}, X_t, A_t = a] da \right] 
\end{equation*}
\begin{equation*}
	\underset{(e)}{=} \frac{1}{T \epsilon^2 } \sup_{\MC{P} \in \boldP} \sum_{t=1}^T \E_{\MC{P}} \left[ \int_{a \in \MC{A}} \pi_t^{\TN{sta}}(a, X_t) \E_{\MC{P}} [ f_t^2 | \HH_{t-1}, X_t, A_t = a] da \right] 
\end{equation*}
\begin{equation*}
	\underset{(f)}{=} \frac{1}{T \epsilon^2} \sup_{\MC{P} \in \boldP} \sum_{t=1}^T \E_{\MC{P}, \pi_t^{\TN{sta}}} \left[ f_t^2 \right] 
	\underset{(g)}{\leq} \frac{4 m}{\epsilon^2}
\end{equation*}

\begin{itemize}
	\item Above (a) holds by Chebyshev's inequality.
	\item (b) holds because the above terms form a martingale difference sequence with respect to $\HH_{t-1}$, i.e., $\E_{\MC{P}, \pi} \big[ W_t f_t - \E_{\MC{P}, \pi}[ W_t f_t | \HH_{t-1} ] \big| \HH_{t-1} \big] = 0$; this implies that cross terms disappear, i.e., for $t > s$, 
	\begin{equation*}
		\E_{\MC{P}, \pi} \bigg[ \bigg( W_t f_t - \E_{\MC{P}, \pi} [ W_t f_t | \HH_{t-1} ] \bigg) \bigg( W_s f_s - \E_{\MC{P}, \pi} [ W_s f_s | \HH_{s-1} ] \bigg) \bigg]
	\end{equation*}
	\begin{equation*}
		= \E_{\MC{P}, \pi} \bigg[ \E_{\MC{P}, \pi} \bigg[ \bigg( W_t f_t - \E_{\MC{P}, \pi} [ W_t f_t | \HH_{t-1} ] \bigg) \bigg( W_s f_s - \E_{\MC{P}, \pi} [ W_s f_s | \HH_{s-1} ] \bigg) \bigg| \HH_{t-1} \bigg] \bigg]
	\end{equation*}
	Since $s > t$,
	\begin{equation*}
		= \E_{\MC{P}, \pi} \bigg[ \bigg( W_s f_s - \E_{\MC{P}, \pi} [ W_s f_s | \HH_{s-1} ] \bigg) \E_{\MC{P}, \pi} \bigg[ W_t f_t - \E_{\MC{P}, \pi} [ W_t f_t | \HH_{t-1} ] \bigg| \HH_{t-1} \bigg] \bigg] = 0.
	\end{equation*}
	\item (c) holds because $\E_{\MC{P}, \pi} \left[ \left\{ W_t f_t - \E_{\MC{P}, \pi} [ W_t f_t | \HH_{t-1} ] \right\}^2 \right] = \E_{\MC{P}, \pi} \left[ W_t^2 f_t^2 \right] - \E_{\MC{P}, \pi} \left[ \E_{\MC{P}, \pi}[ W_t f_t | \HH_{t-1} ]^2 \right] \leq \E_{\MC{P}, \pi} \left[ W_t^2 f_t^2 \right]$.
	\item (d) holds by law of iterated expectations.
	\item (e) holds because $W_t = \sqrt{ \frac{ \pi_t^{\TN{sta}}(A_t, X_t) }{\pi_t(A_t, X_t, \HH_{t-1}) } }$.
	\item (f) holds since by Condition \ref{cond:iidpo}, $\E_{\MC{P}} [ f_t^2 | \HH_{t-1}, X_t, A_t ]  = \E_{\MC{P}} [ f_t^2 | X_t, A_t]$ and by law of iterated expectations $\E_{\MC{P}, \pi_t^{\TN{sta}}} \left[ f_t^2 \right] = \E_{\MC{P}} \left[ \int_{a \in \MC{A}} \pi_t^{\TN{sta}}(a, X_t) \E_{\MC{P}} [ f_t^2 | X_t, A_t = a] da \right]$.
	\item (g) holds since $\sup_{\MC{P} \in \boldP, t \geq 1} \E_{\MC{P}, \pi_t^{\TN{sta}}} \left[ f_t^2 \right] < m < \infty$.
\end{itemize}

\begin{lemma}
\label{lem:fumWLLN}
Let $m_{\theta,t} := m_{\theta} (Y_t, X_t, A_t)$. Under Conditions \ref{cond:iidpo}, \ref{cond:compact}, \ref{cond:lipschitz}, \ref{cond:moments}, \ref{cond:optimalsolution}, and \ref{cond:sqrtweights},
\begin{equation}
	\label{eqn:fumWLLN}
	\sup_{\theta \in \Theta} \left\{ \frac{1}{T} \sum_{t=1}^T W_t m_{\theta,t} - \E_{\MC{P}, \pi}[ W_t m_{\theta,t} | \HH_{t-1} ] \right\} = O_{\MC{P} \in \boldP}(1).
\end{equation}
\end{lemma}

Lemma \ref{lem:wllnW} is a type of martingale functionally uniform law of large number result and the proof is similar to the functionally uniform law of large numbers proofs for i.i.d. random variables \citet[Theorem 2.4.1]{van1996weak}.

\paragraph{Proof:} 
~

\bo{Finite Bracketing Number:} 
Let $\delta > 0$. We construct a set $B_\delta$ which is made up of pairs of functions $(l, u)$. We show that we can find $B_\delta$ that satisfies the following:
\begin{enumerate}[label=(\alph*)]
	\item \label{prop:FBN1} For any $\theta \in \Theta$, we can find $(l, u) \in B_{\delta}$ such that \\
	(i) $l(y, x, a) \leq m_\theta(y, x, a) \leq u(y, x, a)$ for all $(x, y)$ in the joint support of $\{ \MC{P} \in \boldP \}$ and all $a \in \MC{A}$. \\
	(ii) $\sup_{\MC{P} \in \boldP, t \geq 1} \E_{\MC{P}, \pi_t^{\TN{sta}}} \left[ \left| u(Y_t, X_t, A_t) - l(Y_t, X_t, A_t) \right| \right] \leq \delta$. 
	\item \label{prop:FBN2} The number of pairs in this set is finite, i.e., $| B_{\delta} | < \infty$.
	\item \label{prop:FBN3} For any $(l, u) \in B_{\delta}$, for some $m < \infty$ which does no depend on $\delta$, \\
	$\sup_{\MC{P} \in \boldP, t \geq 1} \E_{\MC{P},\pi_t^{\TN{sta}}} \left[ u (Y_t, X_t, A_t)^2 \right] \leq m$ and $\sup_{\MC{P} \in \boldP, t \geq 1} \E_{\MC{P},\pi_t^{\TN{sta}}} \left[ l (Y_t, X_t, A_t)^2 \right] \leq m$.
\end{enumerate}
Showing that we can find $B_\delta$ that satisfy \ref{prop:FBN1}, means that $| B_{\delta} |$ is an upper bound on the bracketing number of $\{ m_\theta : \theta \in \Theta \}$. For more information on bracketing functions, see \citet{van1996weak} and \citet{van2000asymptotic}.

To construct $B_\delta$, we follow a similar argument to Example 19.7 of \citet{van2000asymptotic} (page 271). Make a grid over $\Theta$ with meshwidth $\lambda/2 > 0$ and let the points in this grid be the set $G_{\lambda/2} \subseteq \Theta$; we will specify $\lambda$ later. Note that by construction, for any $\theta \in \Theta$ we can find a $\theta \in G_{\lambda/2}$ such that $\| \theta' - \theta \| \leq \lambda$. 

By our Lipschitz Condition \ref{cond:lipschitz}, we have that for any $\theta, \theta' \in \Theta$, $| m_{\theta} (Y_t, X_t, A_t) - m_{\theta'} (Y_t, X_t, A_t) | \leq g( Y_t, X_t, A_t) \| \theta - \theta' \|$ for function $g$ such that for some $m_g < \infty$,
\begin{equation}
	\label{eqn:mg}
	\sup_{\MC{P} \in \boldP, t \geq 1} \E_{\MC{P}, \pi_t^{\TN{sta}}}[ g (Y_t, X_t, A_t)^2 ] \leq m_g.
\end{equation} 
We now show that we can choose $B_\delta = \left\{ \left( m_\theta - g( Y_t, X_t, A_t), m_\theta + g( Y_t, X_t, A_t) \right) : \theta \in G_{\lambda/2} \right\}$. Note that by compactness of $\Theta$, Condition \ref{cond:compact}, the number of points in $G_{\lambda/2}$ is finite, so \ref{prop:FBN2} above holds.

To show that \ref{prop:FBN1} holds for our choice of $B_\delta$, recall that for any $\theta \in \Theta$ we can find a $\theta' \in G_{\lambda/2}$ such that $\| \theta' - \theta \| \leq \lambda$. Also, by the Lipschitz Condition \ref{cond:lipschitz}, $| m_{\theta} (Y_t, X_t, A_t) - m_{\theta'} (Y_t, X_t, A_t) | \leq g( Y_t, X_t, A_t) \| \theta - \theta' \| \leq g( Y_t, X_t, A_t) \lambda$. Thus we have that
\begin{equation*}
    m_{\theta'} (Y_t, X_t, A_t) - g( Y_t, X_t, A_t) \lambda
    \leq m_{\theta} (Y_t, X_t, A_t)
    \leq m_{\theta'} (Y_t, X_t, A_t) + g( Y_t, X_t, A_t) \lambda.
\end{equation*}
Note that
\begin{equation*}
    \sup_{\MC{P} \in \boldP, t \geq 1} \E_{\MC{P}, \pi_t^{\TN{sta}}} \left[ m_{\theta'} (Y_t, X_t, A_t) + g( Y_t, X_t, A_t) \lambda - \left\{ m_{\theta'} (Y_t, X_t, A_t) - g( Y_t, X_t, A_t) \lambda \right\} \right]
\end{equation*}
\begin{equation*}
    = 2 \lambda \sup_{\MC{P} \in \boldP, t \geq 1} \E_{\MC{P}, \pi_t^{\TN{sta}}} \left[ g( Y_t, X_t, A_t) \right]
    \leq 2 \lambda \sqrt{m_g} < \infty.
\end{equation*}
The inequalities above hold by Equation~\eqref{eqn:mg} and since $\E_{\MC{P}, \pi_t^{\TN{sta}}} \left[ g( Y_t, X_t, A_t) \right] \leq \sqrt{ \E_{\MC{P}, \pi_t^{\TN{sta}}} \left[ g( Y_t, X_t, A_t)^2 \right] }$ by Jensen's inequality. \ref{prop:FBN1} above holds for our choice of $B_\delta$ by letting meshwidth $\lambda = \delta / (2 \sqrt{m_g})$. 

We now show that \ref{prop:FBN3} above holds. Note that
\begin{equation*}
	\sup_{\MC{P} \in \boldP, t \geq 1} \E_{\MC{P}, \pi_t^{\TN{sta}}} \left[ \left\{ m_\theta(Y_t, X_t, A_t) + g (Y_t, X_t, A_t) \right\}^2 \right] 
\end{equation*}
\begin{equation}
	\label{eqn:bracketingUpperbound}
	\leq 3 \sup_{\MC{P} \in \boldP, t \geq 1} \E_{\MC{P}, \pi_t^{\TN{sta}}} \left[ m_\theta(Y_t, X_t, A_t)^2 \right] 
	+ 3 \sup_{\MC{P} \in \boldP, t \geq 1} \E_{\MC{P}, \pi_t^{\TN{sta}}} \left[ g (Y_t, X_t, A_t)^2 \right].
\end{equation}
Note that the above upper bound, Equation~\eqref{eqn:bracketingUpperbound}, also holds for \\
$\sup_{\MC{P} \in \boldP, t \geq 1} \E_{\MC{P}, \pi_t^{\TN{sta}}} \left[ \left\{ m_\theta(Y_t, X_t, A_t) - g (Y_t, X_t, A_t) \right\}^2 \right]$.

Since, $m_\theta(Y_t, X_t, A_t) = m_\theta(Y_t, X_t, A_t) - m_{\theta^*(\MC{P})}(Y_t, X_t, A_t) + m_{\theta^*(\MC{P})}(Y_t, X_t, A_t)$,
\begin{multline*}
	\leq 9 \sup_{\MC{P} \in \boldP, t \geq 1} \E_{\MC{P}, \pi_t^{\TN{sta}}} \left[ \left\{ m_\theta(Y_t, X_t, A_t) - m_{\theta^*(\MC{P})}(Y_t, X_t, A_t) \right\}^2 \right] \\
	+ 9 \sup_{\MC{P} \in \boldP, t \geq 1} \E_{\MC{P}, \pi_t^{\TN{sta}}} \left[ m_{\theta^*(\MC{P})}(Y_t, X_t, A_t)^2 \right] \\
	+ 3 \sup_{\MC{P} \in \boldP, t \geq 1} \E_{\MC{P}, \pi_t^{\TN{sta}}} \left[ g (Y_t, X_t, A_t)^2 \right].
\end{multline*}
Note that $\sup_{\MC{P} \in \boldP, t \geq 1} \E_{\MC{P}, \pi_t^{\TN{sta}}} \left[ m_{\theta^*(\MC{P})}(Y_t, X_t, A_t)^2 \right]$ is bounded by our moment Condition \ref{cond:moments} and that $\sup_{\MC{P} \in \boldP, t \geq 1} \E_{\MC{P}, \pi_t^{\TN{sta}}} \left[ g (Y_t, X_t, A_t)^2 \right]$ is bounded by Equation~\eqref{eqn:mg}.

By our Lipschitz Condition \ref{cond:lipschitz}, for any $\theta \in \Theta$, $| m_{\theta} (Y_t, X_t, A_t) - m_{\theta^*(\MC{P})} (Y_t, X_t, A_t) | \leq g( Y_t, X_t, A_t) \| \theta - \theta^*(\MC{P}) \|$. Thus,
\begin{multline*}
	\sup_{\MC{P} \in \boldP, t \geq 1} \E_{\MC{P}, \pi_t^{\TN{sta}}} \left[ \left\{ m_\theta(Y_t, X_t, A_t) - m_{\theta^*(\MC{P})}(Y_t, X_t, A_t) \right\}^2 \right] \\
	\leq \sup_{\MC{P} \in \boldP, t \geq 1} \E_{\MC{P}, \pi_t^{\TN{sta}}} \left[ g( Y_t, X_t, A_t)^2 \right] \| \theta - \theta^*(\MC{P}) \|^2.
\end{multline*}
The above is bounded by Equation~\eqref{eqn:mg} and by compactness of $\Theta$, Condition \ref{cond:compact}. Thus \ref{prop:FBN3} above holds for our choice of $B_\delta$.

\bo{Main Argument:}
We now show that for any $\epsilon > 0$,
\begin{equation}
	\label{eqn:UsquaredDesired}
	\sup_{\MC{P} \in \boldP} \PP_{\MC{P}, \pi} \left( \sup_{\theta \in \Theta} \left\{ \frac{1}{T} \sum_{t=1}^T  W_t m_{\theta,t} - \E_{\MC{P}, \pi} [ W_t m_{\theta,t} | \HH_{t-1} ] \right\} > \epsilon \right) \to 0.
\end{equation}
An analogous argument can be made to show that \\
$\sup_{\MC{P} \in \boldP} \PP_{\MC{P}, \pi} \left( \sup_{\theta \in \Theta} \left\{ - \frac{1}{T} \sum_{t=1}^T W_t m_{\theta,t} - \E_{\MC{P}, \pi} [ W_t m_{\theta,t} | \HH_{t-1} ] \right\} > \epsilon \right) \to 0.$

Let $\delta > 0$; we will choose $\delta$ later. Let $B_\delta$ be the set of pairs of functions as constructed earlier.
\begin{equation*}
	\sup_{\theta \in \Theta} \left\{ \frac{1}{T} \sum_{t=1}^T W_t m_{\theta,t} - \E_{\MC{P}, \pi} [ W_t m_{\theta,t} | \HH_{t-1} ] \right\}
\end{equation*}
Note that by \ref{prop:FBN1}, we get the following upper bound:
\begin{equation*}
	\leq \max_{ (l,u) \in B_\delta } \left\{ \frac{1}{T} \sum_{t=1}^T W_t u(Y_t, X_t, A_t) - \E_{\MC{P}, \pi} [ W_t l(Y_t, X_t, A_t) | \HH_{t-1} ] \right\}.
\end{equation*}
By adding and subtracting $\E_{\MC{P}, \pi} \left[ W_t u(Y_t, X_t, A_t) \big| \HH_{t-1} \right]$ and triangle inequality,
\begin{multline*}
	\leq \max_{ (l,u) \in B_\delta } \left\{ \frac{1}{T} \sum_{t=1}^T \E_{\MC{P}, \pi} \left[ W_t \left\{ u(Y_t, X_t, A_t) - l(Y_t, X_t, A_t) \right\} \big| \HH_{t-1} \right] \right\} \\
	+ \max_{ (l,u) \in B_\delta } \left\{ \frac{1}{T} \sum_{t=1}^T W_t u(Y_t, X_t, A_t) - \E_{\MC{P}, \pi} \left[ W_t u(Y_t, X_t, A_t) \big| \HH_{t-1} \right] \right\}.
\end{multline*}

Note that by Condition \ref{cond:sqrtweights}, $W_t = \sqrt{ \frac{ \pi_t^{\TN{sta}} (A_t, X_t) }{ \pi_t(A_t, X_t, \HH_{t-1}) } } \leq \sqrt{ \rho_{\max}}$ with probability $1$, so \\
$\E_{\MC{P}, \pi} \left[ W_t \left\{ u(Y_t, X_t, A_t) - l(Y_t, X_t, A_t) \right\} \big| \HH_{t-1} \right] 
\leq \frac{1}{ \sqrt{\rho_{\max}} } \E_{\MC{P}, \pi} \left[ W_t^2 \left\{ u(Y_t, X_t, A_t) - l(Y_t, X_t, A_t) \right\} \big| \HH_{t-1} \right]$ \\
$= \frac{1}{ \sqrt{\rho_{\max}} } \E_{\MC{P}, \pi_t^{\TN{sta}}} \left[ u(Y_t, X_t, A_t) - l(Y_t, X_t, A_t) \right] \leq \frac{1}{ \sqrt{\rho_{\max}} } \delta$; the last equality holds by Condition \ref{cond:iidpo} and the last inequality holds by \ref{prop:FBN1}. And since $\max_{i \in [1 \colon n]} \{ a_i \} \leq \sum_{i=1}^n |a_i|$,
\begin{equation*}
	\leq \frac{1}{ \sqrt{\rho_{\max}} } \delta
	+ \sum_{ (l, u) \in B_\delta } \left| \frac{1}{T} \sum_{t=1}^T W_t u(Y_t, X_t, A_t) - \E_{\MC{P}, \pi} \left[ W_t u(Y_t, X_t, A_t) | \HH_{t-1} \right] \right|
\end{equation*}
By Lemma \ref{lem:wllnW} and \ref{prop:FBN3}, for any $(l,u) \in B_\delta$, $\frac{1}{T} \sum_{t=1}^T W_t u(Y_t, X_t, A_t) - \E_{\MC{P}, \pi} \left[ W_t u(Y_t, X_t, A_t) \big| \HH_{t-1} \right] = o_{\MC{P} \in \boldP}(1)$ . Since $| B_{\delta} | < \infty$ by \ref{prop:FBN2}, the convergence holds for all $(l,u) \in B_\delta$ simultaneously, so
\begin{equation*}
	= \frac{1}{ \sqrt{\rho_{\max}} } \delta + o_{\MC{P} \in \boldP}(1).
\end{equation*}
Equation~\eqref{eqn:UsquaredDesired} holds by choosing $\delta = \sqrt{\rho_{\max}} \epsilon / 2$.

\subsection{Least-Squares Estimator}
\label{app:AWLS}

We use $\phi(X_t, A_t)$ to denote a feature vector that constructed using context $X_t$ and action $A_t$.

\begin{condition}[Linear Expected Outcome]
	\label{cond:linearoutcome}
	For all $\MC{P} \in \boldP$, the following holds w.p. $1$,
	\begin{equation*}
		\E_{\MC{P}} \left[ Y_t | X_t, A_t \right] = \phi(X_t, A_t)^\top \theta^*(\MC{P}).
	\end{equation*}
\end{condition}

\begin{condition}[Moment Conditions for Least Squares]
	\label{cond:momentsLS}
	The fourth moments of $\phi(X_t, A_t) \left( Y_t - \phi(X_t, A_t)^\top \theta^*(\MC{P}) \right)$ and $\phi(X_t, A_t)$ with respect to $\MC{P}$ and policy $\pi_t^{\TN{sta}}$ are respectively bounded uniformly over $\MC{P} \in \boldP$ and $t \geq 1$. 
	
	Also the minimum eigenvalue of $\Sigma_T(\MC{P}) = \frac{1}{T} \sum_{t=1}^T \E_{ \MC{P}, \pi_t^{\TN{sta}} } \left[ \phi (Y_t, X_t, A_t)^{\otimes 2} \left( Y_t - \phi (Y_t, X_t, A_t)^\top \theta^*(\MC{P}) \right)^2 \right]$ and $\frac{1}{T} \sum_{t=1}^T \E_{\MC{P}, \pi_t^{\TN{sta}}} \left[ \phi(X_t, A_t)^{\otimes 2} \right]$ respectively are both bounded above constant some constant greater than zero for all $\MC{P} \in \boldP$.
\end{condition}

\begin{condition}[Importance Ratios for Least Squares]
	\label{cond:growingratios}
	Let $\rho_{\min} > 0$ and $\rho_{\max, T} > 0$ be a non-random sequence such that $\frac{\rho_{\max, T}}{T} \to 0$. $\{ \pi_t^{\TN{sta}} \}_{t=1}^T$ are pre-specified and do not depend on data $\{ Y_t, X_t, A_t \}_{t=1}^T$.
	For all $\MC{P} \in \boldP$, the following holds w.p. $1$,
	\begin{equation*}
		\rho_{\min} \leq \frac{ \pi^{\TN{sta}}_t(A_t, X_t) }{ \pi_t(A_t, X_t, \HH_{t-1}) } \leq \rho_{\max,T}.
	\end{equation*}
\end{condition}
Note that Condition \ref{cond:growingratios} allows $\pi_t(A_t, X_t, \HH_{t-1})$ to go to zero at some rate for stabilizing policies $\{ \pi_t^{\TN{sta}} \}_{t \geq 1}$ that are strictly bounded away from $0$ and $1$. 

We now define the AW-LS estimator for $\theta^*(\MC{P}) \in \real^d$:
\begin{equation}
	\label{eqn:AWLScriterion}
	\thetahat_T^{\TN{AW-LS}} := \argmax_{\theta \in \real^d} \left\{ - \sum_{t=1}^T W_t \left( Y_t - \phi(X_t, A_t)^\top \theta \right)^2 \right\}.
\end{equation}

\begin{theorem}[Consistency and Asymptotic Normality of Adaptively-Weighted Least Squares Estimator]
	Under Conditions \ref{cond:iidpo}, \ref{cond:linearoutcome}, \ref{cond:momentsLS}, and \ref{cond:growingratios},
	\begin{equation*}
		\Sigma_T(\MC{P})^{-1/2} \left( \frac{1}{ \sqrt{T} } \sum_{t=1}^T W_t \phi(X_t, A_t)^{\otimes 2} \right) 
		\left( \thetahat_T^{\TN{AW-LS}} - \theta^*(\MC{P}) \right)
		\Dto \N( 0, I_d) 
		\TN{ uniformly over } \MC{P} \in \boldP,
	\end{equation*}
	where $\Sigma_T(\MC{P}) := \frac{1}{T} \sum_{t=1}^T \phi(X_t, A_t)^{\otimes 2} \left( Y_t - \phi(X_t, A_t)^\top \theta^*(\MC{P}) \right)^2$.
\end{theorem}

\paragraph{Proof:}

By taking the derivative of Equation~\eqref{eqn:AWLScriterion} with respect to the parameters, we have that
\begin{equation*}
	0 = \sum_{t=1}^T W_t \phi(X_t, A_t) \left( Y_t - \phi(X_t, A_t)^\top \thetahat_T^{\TN{AW-LS}} \right).
\end{equation*}
By rearranging terms, we have that
\begin{multline}
	\label{eqn:taylorAWLS}
	- \frac{1}{ \sqrt{T} } \sum_{t=1}^T W_t \phi(X_t, A_t) \left( Y_t - \phi(X_t, A_t)^\top \theta^*(\MC{P}) \right) \\
	= \frac{1}{ \sqrt{T} } \sum_{t=1}^T W_t \phi(X_t, A_t)^{\otimes 2} \left( \thetahat_T^{\TN{AW-LS}} - \theta^*(\MC{P}) \right).
\end{multline}

We first show that the following holds:
\begin{equation}
	\label{eqn:normalityAWLS}
	\Sigma_T(\MC{P})^{-1/2} \frac{1}{ \sqrt{T} } \sum_{t=1}^T W_t \phi(X_t, A_t) \left( Y_t - \phi(X_t, A_t)^\top \theta^*(\MC{P}) \right)
	\Dto \N( 0, I_d) \TN{ uniformly over } \MC{P} \in \boldP.
\end{equation}
Equation~\eqref{eqn:normalityAWLS} holds by a similar argument as that used in Section \ref{sec:asyptoticnormality}, for $\dot{m}_\theta(Y_t, X_t, A_t) = \phi(X_t, A_t) \left( Y_t - \phi(X_t, A_t)^\top \theta^*(\MC{P}) \right)$ by showing that the conditions of Theorem \ref{thm:umCLT} hold.
It can be checked that all the arguments hold even when we allow $\rho_{\max,T}$ to grow at a rate such that $\frac{ \rho_{\max,T} }{T} \to 0$.

By Equations~\eqref{eqn:taylorAWLS} and \eqref{eqn:normalityAWLS},
\begin{equation}
	\label{eqn:normalityAWLS2}
	\Sigma_T(\MC{P})^{-1/2} \frac{1}{ \sqrt{T} } \sum_{t=1}^T W_t \phi(X_t, A_t)^{\otimes 2} \left( \thetahat_T^{\TN{AW-LS}} - \theta^*(\MC{P}) \right)
	\Dto \N( 0, I_d) \TN{ uniformly over } \MC{P} \in \boldP.
\end{equation}
By Equation~\eqref{eqn:normalityAWLS2}, to ensure that $\thetahat_T^{\TN{AW-LS}} \Pto \theta^*(\MC{P})$ uniformly over $\MC{P} \in \boldP$, it is sufficient to show that the minimum eigenvalue of $\Sigma_T(\MC{P})^{-1/2} \frac{1}{ \sqrt{T} } \sum_{t=1}^T W_t \phi(X_t, A_t)^{\otimes 2}$ goes to infinity uniformly over $\MC{P} \in \boldP$ as $T \to \infty$. 

By Condition \ref{cond:momentsLS}, the maximum eigenvalue of $\Sigma_T(\MC{P})$ is bounded uniformly over $\MC{P} \in \boldP$, so the minimum eigenvalue of $\Sigma_T(\MC{P})^{-1/2}$ is bounded uniformly above $0$. Thus it is sufficient to show that the minimum eigenvalue of $\frac{1}{ \sqrt{T} } \sum_{t=1}^T W_t \phi(X_t, A_t)^{\otimes 2}$ goes to infinity uniformly over $\MC{P} \in \boldP$ as $T \to \infty$. 

Note that by Lemma \ref{lem:wllnW} and Condition \ref{cond:momentsLS},
\begin{equation}
	\label{eqn:awlsWLLN}
	\frac{1}{ \sqrt{T} } \sum_{t=1}^T W_t \phi(X_t, A_t)^{\otimes 2} - \E_{\MC{P}, \pi} \left[ W_t \phi(X_t, A_t)^{\otimes 2} \big| \HH_{t-1} \right] = O_{\MC{P} \in \boldP}(1).
\end{equation}
Note that by law of iterated expectations,
\begin{equation*}
	\E_{\MC{P}, \pi} \left[ W_t \phi(X_t, A_t)^{\otimes 2} \big| \HH_{t-1} \right]
\end{equation*}
\begin{equation*}
	= \E_{\MC{P}} \left[ \int_{a \in \MC{A}} \pi_t(a, X_t, \HH_{t-1}) \E_{\MC{P}} \left[ W_t \phi(X_t, A_t)^{\otimes 2} | \HH_{t,1}, X_t, a \right] da \bigg| \HH_{t-1} \right].
\end{equation*}
By Condition \ref{cond:iidpo} and since $W_t = \sqrt{ \frac{ \pi^{\TN{sta}}_t(A_t, X_t) }{ \pi_t(A_t, X_t, \HH_{t-1}) } }$,
\begin{equation*}
	= \E_{\MC{P}} \left[ \int_{a \in \MC{A}} \sqrt{ \frac{ \pi_t(a, X_t, \HH_{t-1}) }{ \pi^{\TN{sta}}_t(a, X_t) } } 
	\pi^{\TN{sta}}_t(a, X_t) \E_{\MC{P}} \left[ \phi(X_t, A_t)^{\otimes 2} | X_t, a \right] da \bigg| \HH_{t-1} \right]
\end{equation*}
Since by Condition \ref{cond:growingratios}, $\frac{ \pi_t(a, X_t, \HH_{t-1}) }{ \pi^{\TN{sta}}_t(a, X_t) } \geq \frac{1}{ \sqrt{ \rho_{\max, T} } }$ and $\phi(X_t, A_t)^{\otimes 2} \succeq 0$,
\begin{equation*}
	\succeq \frac{1}{ \sqrt{ \rho_{\max, T} } } 
	\E_{\MC{P}} \left[ \int_{a \in \MC{A}} 
	\pi^{\TN{sta}}_t(a, X_t) \E_{\MC{P}} \left[ \phi(X_t, A_t)^{\otimes 2} | X_t, a \right] da \bigg| \HH_{t-1} \right].
\end{equation*}
Since $\pi_t^{\TN{sta}}$ are pre-specified and since by our i.i.d. potential outcomes assumption (Condition \ref{cond:iidpo}) $X_t$ do not depend on $\HH_{t-1}$, 
\begin{equation*}
	= \frac{1}{ \sqrt{ \rho_{\max, T} } } 
	\E_{\MC{P}} \left[ \int_{a \in \MC{A}} 
	\pi^{\TN{sta}}_t(a, X_t) \E_{\MC{P}} \left[ \phi(X_t, A_t)^{\otimes 2} | X_t, a \right] da \right].
\end{equation*}
By law of iterated expectations,
\begin{equation*}
	= \frac{1}{ \sqrt{ \rho_{\max, T} } } 
	\E_{\MC{P}, \pi_t^{\TN{sta}}} \left[ \phi(X_t, A_t)^{\otimes 2} \right].
\end{equation*}
The above result and Equation~\eqref{eqn:awlsWLLN} implies that
\begin{equation}
	\label{eqn:awlsLowerbound}
	\frac{1}{ \sqrt{T} } \sum_{t=1}^T W_t \phi(X_t, A_t)^{\otimes 2} 
	\succeq O_{\MC{P} \in \boldP}(1) + \sqrt{ \frac{T}{ \rho_{\max, T} } } \frac{1}{T} \sum_{t=1}^T \E_{\MC{P}, \pi_t^{\TN{sta}}} \left[ \phi(X_t, A_t)^{\otimes 2} \right].
\end{equation}
By Condition \ref{cond:momentsLS}, the minimum eigenvalue of $\frac{1}{T} \sum_{t=1}^T \E_{\MC{P}, \pi_t^{\TN{sta}}} \left[ \phi(X_t, A_t)^{\otimes 2} \right]$ is bounded above some constant greater than zero for all $\MC{P} \in \boldP$. By Condition \ref{cond:growingratios}, $\sqrt{ \frac{T}{ \rho_{\max, T} } } \to \infty$. Thus by Equation~\eqref{eqn:normalityAWLS2} and Equation~\eqref{eqn:awlsLowerbound}, we have that $\thetahat_T^{\TN{AW-LS}} \Pto \theta^*(\MC{P})$ uniformly over $\MC{P} \in \boldP$.

%% file: appendix/stabilizing_policy.tex
\section{Choice of Stabilizing Policy}
\label{app:stabilizingpolicy}

\subsection{Optimal Stabilizing Policy in Multi-Arm Bandit Setting}
Here we consider the multi-armed bandit setting where $\E_{\MC{P}}[ Y_t(a) ] = \theta_a^*(\MC{P})$ and $\Var_{\MC{P}} ( Y_t(a) ) = \sigma^2$. We consider the adaptively-weighted least-squares estimator where $m_\theta( Y_t, A_t ) = - \1_{A_t=a} (Y_t - \theta_a^*(\MC{P}))^2$. By Theorem \ref{thm:normality}, we have that 
\begin{equation*}
    \left( \frac{1}{T} \sum_{t=1}^T \E_{\MC{P}, \pi_t^{\TN{sta}}} \left[ \1_{A_t=a} (Y_t - \theta_a^*(\MC{P}))^2 \right] \right)^{-1/2}
    \left( \frac{1}{T} \sum_{t=1}^T W_t \1_{A_t=a} \right) \sqrt{T} ( \hat{\theta}_{T,a}^{\TN{AW-LS}} - \theta_a^*(\MC{P})) \Dto \N \left( 0, 1 \right).
\end{equation*}
While the asymptotic variance of $\sqrt{T} ( \hat{\theta}_{T,a}^{\TN{AW-LS}} - \theta_a^*(\MC{P}))$ does not necessarily concentrate we can examine the following:
\begin{equation*}
    \left( \frac{1}{T} \sum_{t=1}^T W_t \1_{A_t=a} \right)^{-1}
    \left( \frac{1}{T} \sum_{t=1}^T \E_{\MC{P}, \pi_t^{\TN{sta}}} \left[ \1_{A_t=a} (Y_t - \theta_a^*(\MC{P}))^2 \right] \right)
    \left( \frac{1}{T} \sum_{t=1}^T W_t \1_{A_t=a} \right)^{-1}
\end{equation*}

By Lemma \ref{lem:wllnW}, we have that $\frac{1}{T} \sum_{t=1}^T W_t \1_{A_t=a} - \sqrt{ \pi_t^{\TN{sta}}(a) \pi_t(A_t,  \HH_{t-1}) } \Pto 0$. Thus we have
\begin{equation*}
    = \left( \frac{1}{T} \sum_{t=1}^T \pi_t^{\TN{sta}}(a) \sigma^2 \right)
    \left( o_p(1) + \frac{1}{T} \sum_{t=1}^T \sqrt{ \pi_t^{\TN{sta}}(a) \pi_t(A_t,  \HH_{t-1}) } \right)^{-2}.
\end{equation*}

As long as $\pi_t^{\TN{sta}}(a), \pi_t(A_t,  \HH_{t-1})$ are bounded away from zero w.p. $1$, the $o_p(1)$ term is asymptotically negligible and we can just consider $\left( \frac{1}{T} \sum_{t=1}^T \pi_t^{\TN{sta}}(a) \sigma^2 \right)
    \left( \frac{1}{T} \sum_{t=1}^T \sqrt{ \pi_t^{\TN{sta}}(a) \pi_t(A_t,  \HH_{t-1}) } \right)^{-2}$.
    
 By Cauchy-Schwartz inequality, \\
$\left( \frac{1}{T} \sum_{t=1}^T \sqrt{ \pi_t^{\TN{sta}}(a) \pi_t(a,  \HH_{t-1}) } \right)^2 \leq \left( \frac{1}{T} \sum_{t=1}^T \pi_t^{\TN{sta}}(a) \right) \left( \frac{1}{T} \sum_{t=1}^T \pi_t(a,  \HH_{t-1}) \right)$. 

Thus, $\frac{1}{ \frac{1}{T} \sum_{t=1}^T \pi_t(a,  \HH_{t-1}) } \leq \frac{ \frac{1}{T} \sum_{t=1}^T \pi_t^{\TN{sta}}(a) }{ \left( \frac{1}{T} \sum_{t=1}^T \sqrt{ \pi_t^{\TN{sta}}(a) \pi_t(a,  \HH_{t-1}) } \right)^2 }$, so
\begin{equation*}
	\frac{ \frac{1}{T} \sum_{t=1}^T \pi_t^{\TN{sta}}(a) }{ \left( \frac{1}{T} \sum_{t=1}^T \sqrt{ \pi_t(a,  \HH_{t-1}) \pi_t^{\TN{sta}}(a) } \right)^2 }
	\geq \frac{1}{ \frac{1}{T} \sum_{t=1}^T \pi_t(a,  \HH_{t-1}) }.
\end{equation*}

Note that this lower bound is achieved when $\pi_t^{\TN{sta}}(a) = \pi_t(a)$. However, since $\pi_t$ is a function of $\HH_{t-1}$ and stabilizing policies$\{ \pi_{t}^{\TN{sta}} \}_{t=1}^T$ are pre-specified, setting $\pi_t^{\TN{sta}}(A_t) = \pi_{t,a}$ is generally an unfeasible choice. Thus we want to choose $\pi_t^{\TN{sta}}$ to be as close to $\pi_t$ as possible, subject to the constraint that the stabilizing policies are pre-specified, i.e., not a function of the data $\{ Y_t, X_t, A_t \}_{t \geq 1}$.

\subsection{Approximating the Optimal Stabilizing Policy}
One way to approximately choose the optimal evaluation policy is to select $\pi_t^{\TN{sta}}(a, x) = \E_{\MC{P}, \pi}[ \pi_t(a, x, \HH_{t-1}) ]$. Note that $\E_{\MC{P}, \pi}[ \pi_t(a, x, \HH_{t-1}) ]$ depends on the $\MC{P}$, which is unknown. Thus it is natural to choose $\pi_t^{\TN{sta}}(a, x)$ to be $\E_{\MC{P}, \pi}[ \pi_t(a, x, \HH_{t-1}) ]$ weighted by a prior on $\MC{P}$. Note that as long as the evaluation policy ensures that weights $W_t$ are bounded, the choice of evaluation policy does not affect the asymptotic validity of the estimator. 

\begin{figure}[H]
   \centerline{ \includegraphics[width=0.5\linewidth]{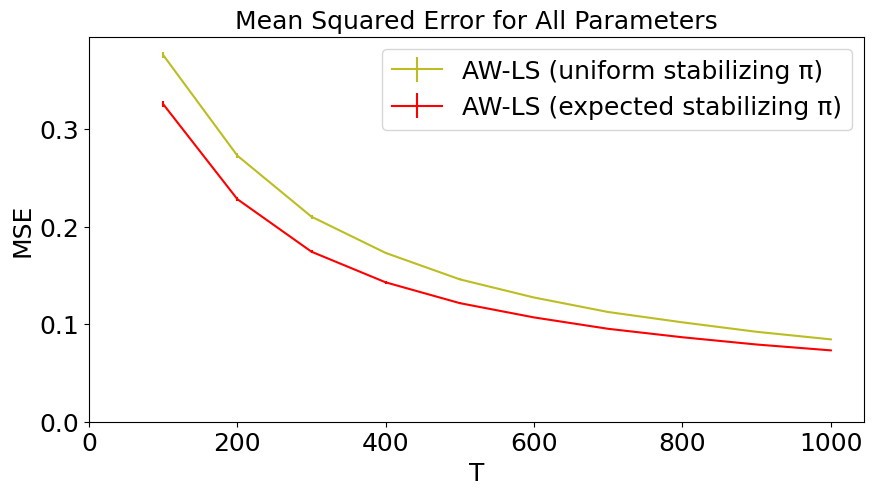}
	}
  \caption{Above we plot the mean squared errors for the adaptively-weighted least squares estimator with evaluation policies: (1) uniform evaluation policy which selects actions uniformly from $\MC{A}$ and (2) expected $\pi_t(a, \HH_{t-1})$ evaluation policy for which $\pi_t^{\TN{sta}}(a) = \E_{\MC{P}, \pi} \left[ \pi_t(a) \right]$ (oracle quantity). In a two arm bandit setting we perform Thompson Sampling with standard normal priors, $0.01$ clipping, $\theta^*(\MC{P}) = [ \theta_0^*(\MC{P}), \theta_1^*(\MC{P}) ] = [0,1]$, standard normal errors, and $T=1000$. Error bars denote standard errors computed over 5,000 Monte Carlo simulations.
  }
  \label{fig:pieval}
\end{figure}

In Figure \ref{fig:pieval}, we display the difference in mean squared error for the AW-LS estimator in a two-armed bandit setting for two different choices of evaluation policy: (1) the uniform evaluation policy which selects actions uniformly from $\MC{A}$ and (2) the expected $\pi_t(a, \HH_{t-1})$ evaluation policy for which $\pi_t^{\TN{sta}}(a) = \E_{\MC{P}, \pi}[\pi_t(a, \HH_{t-1})]$. We can see in this setting that by setting $\pi_t^{\TN{sta}}(a) = \E_{\MC{P}, \pi}[\pi_t(a, \HH_{t-1})]$ we are able to decrease the mean squared error of the AW-LS estimator compared AW-LS with the uniform evaluation policy. Note though that in some cases setting $\pi_t^{\TN{sta}}(a) = \E_{\MC{P}, \pi}[\pi_t(a, \HH_{t-1})]$ is equivalent to choosing the uniform evaluation policy. For example, a two-armed bandit with identical arms so under common bandit algorithms $\E_{\MC{P}, \pi}[\pi_t(a, \HH_{t-1})] = 0.5$ for all $t \in [1 \colon T]$, which will make the evaluation policy $\pi_t^{\TN{sta}}(a) = \E_{\MC{P}, \pi}[\pi_t(a, \HH_{t-1})]$ equivalent to the uniform policy. 

%% file: appendix/uniformity_stuff.tex
\section{Need for Uniformly Valid Inference on Data Collected with Bandit Algorithms}
\label{app:uniformity}

Here we consider the two-armed bandit setting where $\E_{\MC{P}}[ R_t(a) ] = \theta_{0,a}(\MC{P})$, $\Var_{\MC{P}} ( R_t(a) ) = \sigma^2$, and $\E_{\MC{P}}[ R_t(a)^4 ] < c < \infty$ for $a \in \{ 0, 1 \}$. The unweighted least squares estimator is asymptotically normal on adaptively collected data under the following condition of \citet{lai1982least}, there exists a non-random sequence $\{ b_t \}_{t \geq 1}$ such that
\begin{equation}
	\label{eqn:laiweistability}
	b_T \cdot \sum_{t=1}^T A_t \Pto 1.
\end{equation}
Specifically, by Theorem 3 of \citet{lai1982least}, under \eqref{eqn:laiweistability}, 
\begin{equation*}
	\sqrt{ \sum_{t=1}^T A_t } ( \thetahat_{T,1}^{\TN{OLS}} - \theta_1^*(\MC{P}) ) 
	= \frac{ \sum_{t=1}^T A_t ( R_t - \theta_1^*(\MC{P}) ) }{\sqrt{ \sum_{t=1}^T A_t }} \Dto \N (0, \sigma^2).
\end{equation*}

\begin{figure}[H]
	\centerline{
	\includegraphics[width=0.35\linewidth]{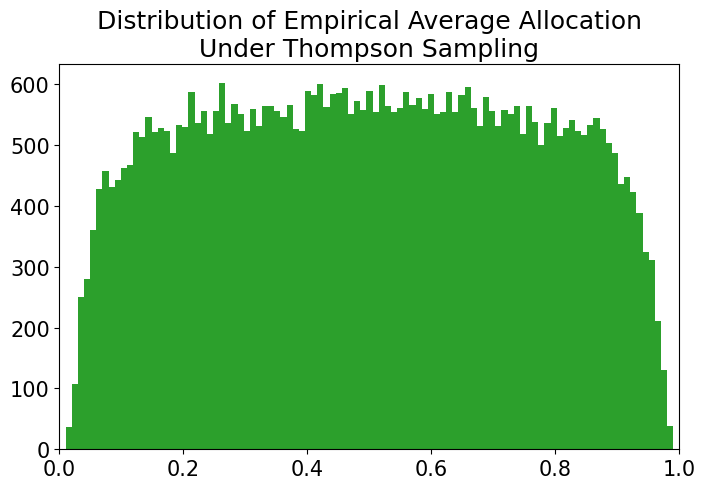}
	~~~~~~~
	\includegraphics[width=0.35\linewidth]{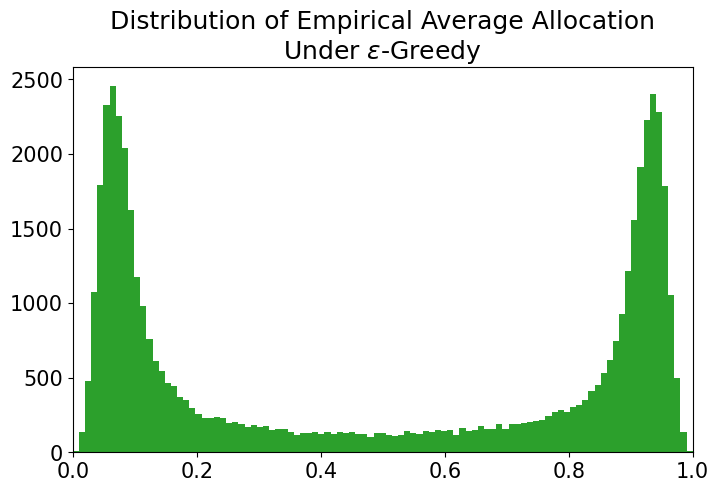}
	}
  \caption{Above we plot empirical allocations, $\frac{1}{T} \sum_{t=1}^T A_t$, under both Thompson Sampling (standard normal priors, $0.01$ clipping) and $\epsilon$-greedy ($\epsilon=0.1$) under zero margin $\theta_0^*(\MC{P}) = \theta_1^*(\MC{P}) = 0$. For our simulations $T=100$, errors are standard normal, and we use $50k$ Monte Carlo repetitions.}
  \label{fig:nonconcentration}
\end{figure}

However, as discussed in \citet{deshpande} and \citet{zhang2020inference}, \eqref{eqn:laiweistability} can fail to to hold for common bandit algorithms when there is no unique optimal policy, i.e., when $\theta_0^*(\MC{P}) - \theta_1^*(\MC{P}) = 0$. For example, in Figure \ref{fig:nonconcentration} we plot $\frac{1}{T} \sum_{t=1}^T A_t$ for Thompson Sampling and $\epsilon$-greedy for a bandit with two identical arms.

In order to construct reliable confidence intervals using asymptotic approximations, it is crucial that that estimators converge uniformly in distribution. To illustrate the importance of uniformity, consider the following example. We can modify Thompson Sampling to ensure that $\frac{1}{T} \sum_{t=1}^T A_t \Pto 0.5$ when $\theta_1^*(\MC{P}) - \theta_0^*(\MC{P}) = 0$. For example, we could do this by using an algorithm we call Thompson Sampling Hodges (inspired by the Hodges estimator; see \citet[Page 109]{van2000asymptotic}), defined below:
\begin{equation*}
	\pi_t(1, \HH_{t-1}) = \PP ( \tilde{\theta}_1 > \tilde{\theta}_0 | \HH_{t-1} ) \1_{ | \mu_{1,t} - \mu_{0,t}  | > t^{-4} }
	+ 0.5 \1_{ | \mu_{1,t} - \mu_{0,t} | \leq t^{-4} }
\end{equation*}
Under standard Thompson Sampling arm one is chosen according to the posterior probability that is optimal, so $\pi_t(1, \HH_{t-1}) = \PP ( \tilde{\theta}_1 > \tilde{\theta}_0 | \HH_{t-1} )$. Above, $\mu_{a,t}$ denotes the posterior mean for the mean reward for arm $a$ at time $t$. Under TS-Hodges, if difference between the posterior means, $|\mu_{1,t}-\mu_{0,t}|$, is less than $t^{-4}$, $\pi_t$ is set to $0.5$. Additionally, we clip the action selection probabilities to bound them strictly away from $0$ and $1$ for some constant $\pi_{\min}$ in the following sense $\TN{clip}(\pi_t) = (1-\pi_{\min}) \wedge ( \pi_t \vee \pi_{\min} )$. Under TS-Hodges with clipping, we can show that 
\begin{equation}
	\label{eqn:tshodgesallocation}
	\frac{1}{T} \sum_{t=1}^T A_t \Pto \begin{cases}
		1-\pi_{\min} & \TN{ if } \theta_1^*(\MC{P}) - \theta_0^*(\MC{P}) > 0 \\
		\pi_{\min} & \TN{ if } \theta_1^*(\MC{P}) - \theta_0^*(\MC{P}) < 0 \\
		0.5 & \TN{ if } \theta_1^*(\MC{P}) - \theta_0^*(\MC{P}) = 0 \\
	\end{cases}
\end{equation}

\begin{figure}[H]
	\centerline{
	\includegraphics[width=0.42\linewidth]{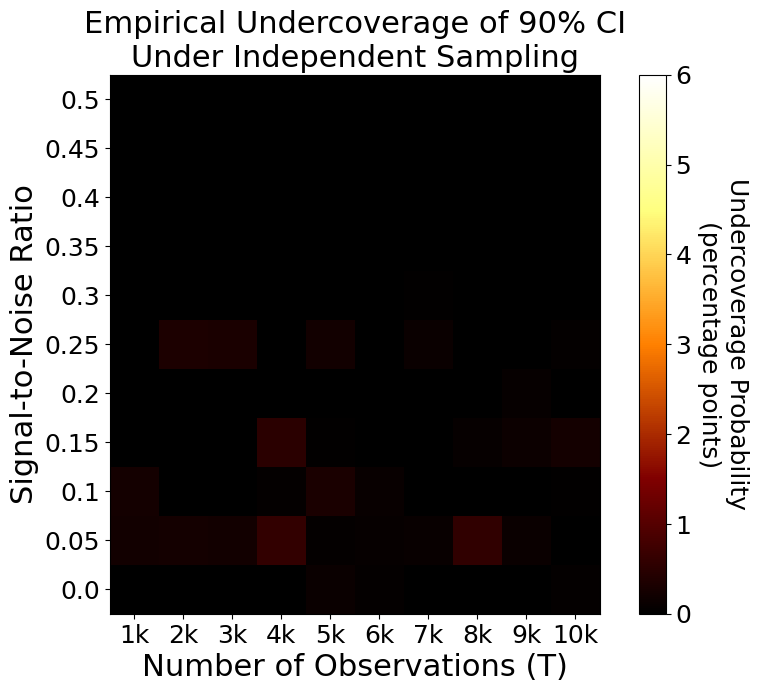}
	~~~~~~~
	\includegraphics[width=0.42\linewidth]{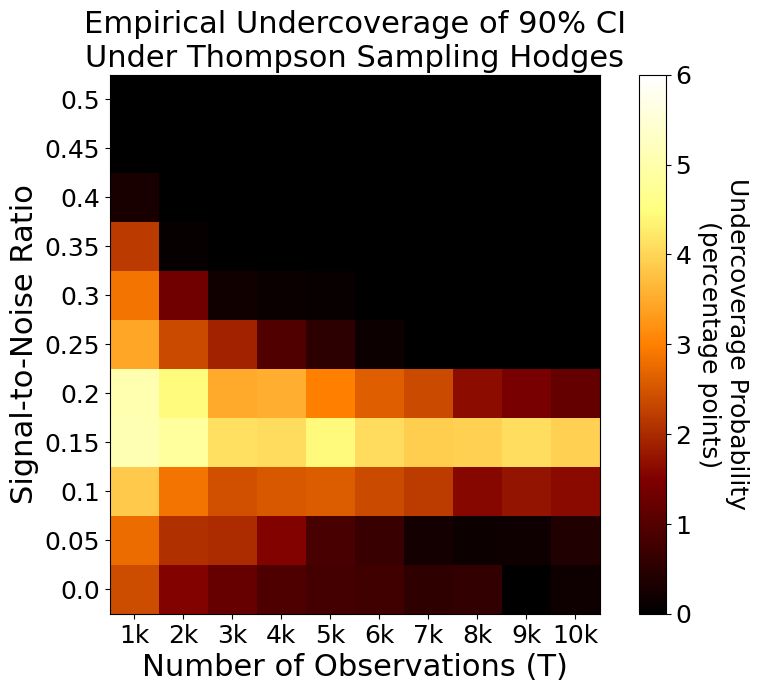}
	}
  \caption{Above we construct confidence intervals for $\theta_1^*(\MC{P}) - \theta_0^*(\MC{P})$ using a normal approximation for the OLS estimator. We compare independent sampling ($\pi_t=0.5$) and TS Hodges, both with standard normal priors, $0.01$ clipping, standard normal errors, and $T=10,000$. We vary the value of $\theta_1^*(\MC{P}) - \theta_0^*(\MC{P})$ in the simulations to demonstrate the non-uniformity of the confidence intervals.}
  \label{fig:nonunformity}
\end{figure}

By equation \eqref{eqn:tshodgesallocation}, we satisfy \eqref{eqn:laiweistability} pointwise for every fixed $\MC{P}$ and we have that the OLS estimator is asymptotically normal \textit{pointwise} \citep{lai1982least}. However, equation \eqref{eqn:tshodgesallocation} fails to hold uniformly over $\MC{P} \in \boldP$. Specifically, it fails to hold for any sequence of $\{ \MC{P}_t \}_{t=1}^\infty$ such that $\theta_1^*(\MC{P}_t) - \theta_0^*(\MC{P}_t) = t^{-4}$. In Figure \ref{fig:nonunformity}, we show that confidence intervals constructed using normal approximations fail to provide reliable confidence intervals, even for very large sample sizes for the worst case values of $\theta_1^*(\MC{P}) - \theta_0^*(\MC{P})$. 

%% file: appendix/chen_paper.tex
\section{Discussion of \citet{chen2020statistical}}
\label{app:chen_paper}


Here we show formally that Theorem 3.1 in \citet{chen2020statistical}, which proves that the OLS estimator is asymptotically normal on data collected with an $\epsilon$-greedy algorithm, does not cover the case in which there is no unique optimal policy.

They assume that for rewards $R_t$, context vectors $X_t$, and binary actions $A_t \in \{ 0, 1 \}$,
\begin{equation*}
	\E [R_t| X_t, A_t ] = A_t \boldX_t^\top \bs{\beta}_1 + (1-A_t) \boldX_t^\top \bs{\beta}_0.
\end{equation*} 
They define $\bs{\beta} := \bs{\beta}_1 - \bs{\beta}_0$.

Specifically at part 1(b) of their proof on page 4 of the supplementary material, they claim that $g( \hat{\bs{\beta}}_t, \epsilon ) \Pto g( \bs{\beta}, \epsilon)$, where $\hat{\bs{\beta}}_t$ is the OLS estimator for $\bs{\beta} := \bs{\beta}_1 - \bs{\beta}_0$ and $g$ is defined as follows:
\begin{equation*}
	g(\bs{\beta}_0, \bs{\beta}_1, \epsilon) = \frac{\epsilon}{2} \int \boldv^\top \boldx \boldx^\top \boldv d \MC{P}_x
	+ (1-\epsilon) \int \1_{ \bs{\beta}^\top \boldx \geq 0} \boldv^\top \boldx \boldx^\top \boldv d \MC{P}_x
\end{equation*}
Above $v \in \real^d$ is arbitrary fixed vector and $x \in \real^d$ are the context vectors. $\MC{P}_x$ is the distribution of the context vectors $\bs{X}_t$.

Specifically, they claim that $g( \hat{\bs{\beta}}_t, \epsilon ) \Pto g( \bs{\beta}, \epsilon)$ because $\hat{\bs{\beta}}_t \Pto \bs{\beta}$ (Corollary 3.1) and by continuous mapping theorem.

Recall the continuous mapping theorem for convergence in probability \citep[Theorem 2.3]{van2000asymptotic}:

\begin{theorem}[Continuous Mapping Theorem]
	Let $g: \real^k \to \real^m$ be continuous at every point of a set $C$ such that $\PP( X \in C ) = 1$. If $X_n \Pto X$, then $g(X_n) \Pto g(X)$.
\end{theorem}

Note that $g$ is not continuous in $\bs{\beta}$ at the value $\bs{\beta} = \bs{0} \in \real^d$; this is due to the indicator term $\1_{\bs{\beta}^\top \boldx \geq 0}$. Thus, the standard continuous mapping theorem can not be applied in this setting. Note that the case that $0 = \bs{\beta} =  \bs{\beta}_1 - \bs{\beta}_0$, is exactly when there is no unique optimal policy. This means that Theorem 3.1 in \citet{chen2020statistical} does not cover the setting in which there is no unique optimal policy.